\documentclass[12pt]{amsart} 

\usepackage{graphicx,amssymb,colordvi,textcomp,latexsym} 

\usepackage{color}
\usepackage{soul}

\newcommand{\examend}{\hfill \mbox{\textreferencemark}}
\newcommand{\rend}{\hfill $\maltese$}

\newcommand{\grad}[1]{{\text{\rm grad}({#1})}}
\newcommand{\is}[1]{{\mathbf{#1}}}
\newcommand{\propG}{{\text{Property }(\mathcal{G})}}

\newcommand{\isZ}[1]{{\mathbf{Z}({#1})}}
\newcommand{\dd}{\mbox{$\;|\;$}}
\newcommand{\trans}{\pitchfork}

\newcommand{\cx}{\mathbb{C}}

\newcommand{\Refb}[1]{(\ref{#1})}

\newcommand{\real}{{\mathbb{R}}}

\newcommand{\intg}{{\mathbb{Z}}}
\newcommand{\xx}{{\mathbf{x}}}
\newcommand{\zz}{{\mathbf{z}}}

\newcommand{\cal}[1]{{\mathcal{#1}}}

\newcommand{\On}[1]{{\text{\rm O}({#1})}}
\newcommand{\defoo}{\stackrel{\mathrm{def}}{=}}

\newcommand{\yy}{{\is{y}}}
\newcommand{\vv}{{\is{v}}}

\newcommand{\VV}{{\is{V}}}
\newcommand{\ww}{{\is{w}}}
\newcommand{\WW}{{\is{W}}}
\newcommand{\uu}{{\is{u}}}

\newcommand{\db}{{\is{d}}}
\newcommand{\ee}{{\is{e}}}

\newcommand{\pint}{\mbox{$\mathbb{N}$}}
\newcommand{\sgn}{\mbox{sgn}}
\newcommand{\arr}{{\rightarrow}}

\newtheorem{lemma}{Lemma}[section]

\newtheorem{prop}[lemma]{Proposition}
\newtheorem{thm}[lemma]{Theorem}
\newtheorem{cor}[lemma]{Corollary}
\theoremstyle{definition}
\newtheorem{Def}[lemma]{Definition}
\newtheorem{exam}[lemma]{Example}
\newtheorem{exams}[lemma]{Examples}

\theoremstyle{remark}

\newtheorem{rem}[lemma]{Remark}
\newtheorem{rems}[lemma]{Remarks}

\newcommand{\vd}{\dot{v}}
\newcommand{\ud}{\dot{u}}
\newenvironment{pfof}[1]{\vspace{1ex}\noindent{\bf Proof of
#1}\hspace{0.5em}}{\hfill\qed\vspace{1ex}}

\newcommand{\vareps}{\varepsilon}
\newcommand{\haf}{\hat{f}}

\title[Equivariant bifurcation]{Equivariant bifurcation, quadratic equivariants, and symmetry breaking for the standard representation of $S_n$}
\author{Yossi Arjevani and Michael Field}
\address{Yossi Arjevani, Center for Data Science, NYU, New York, NY, 10011}
\email{yossi.arjevani@gmail.com}
\address{Michael Field, Department of Mechanical Engineering, UCSB, 
Santa Barbara, CA 93106}
\email{mikefield@gmail.com}

\date{\today}
\begin{document}
\begin{abstract}
Motivated by questions originating from the study of a class of shallow student-teacher neural networks,
methods are developed for the analysis of spurious minima in classes of gradient equivariant dynamics related to neural nets. 
In the symmetric case, methods depend on the generic equivariant bifurcation theory of irreducible representations 
of the symmetric group on $n$ symbols, $S_n$; in particular, the standard representation of $S_n$. 
It is shown that spurious minima do not arise from spontaneous symmetry breaking but rather through a 
complex deformation of the landscape geometry that can be encoded by a generic $S_n$-equivariant bifurcation. 
We describe minimal models for forced symmetry breaking that give a lower bound on the dynamic complexity
involved in the creation of spurious minima when there is no symmetry.  Results on generic bifurcation when there are
quadratic equivariants are also proved; this work extends and clarifies results of
Ihrig \& Golubitsky and Chossat, Lauterback \& Melbourne on the instability of
solutions when there are quadratic equivariants.
\end{abstract}

\maketitle

\section{Introduction}\label{sec: intro}
Using ideas originating in equivariant bifurcation theory, we develop methods that can be used to understand the creation and annihilation of \emph{spurious minima} (non-global local minima) in
shallow neural nets.  Specifically, the results apply to student-teacher networks that inherit symmetry 
from the target model~\cite{ArjevaniField2019c,ArjevaniField2020b,ArjevaniField2021a}. In the first part of the introduction, there is an overview of the
motivation, background and results. In the remainder, we give an outline of contents and description of the mathematical contributions---the focus of the article.

\subsection{Motivation and background}
In general terms, this article developed out of a program to understand why the highly non-convex optimization 
landscapes induced by natural distributions allow gradient based methods, such as stochastic gradient descent (SGD), to find good minima efficiently
(see~\cite{ArjevaniField2021a} for background and sources on neural networks and the student-teacher framework). This article concerns
mathematical aspects of these problems, mainly related to bifurcation theory, and no knowledge of neural networks is required for understanding the results or proofs 
(for the specific optimization problem, see~\cite{ArjevaniField2021a} and the concluding comments section).

Let $S_k$ denote the permutation group on $k$ symbols\footnote{Notations introduced in the introduction are followed throughout the paper.}. 
The foundational theory of generic $S_k$-equivariant steady-state bifurcation on the standard (natural) representation of 
$S_k$ on $H_{k-1} = \{(x_1,\cdots,x_k) \in \real^k \dd \sum_i x_i = 0\}$
was developed by Field \& Richardson about thirty years ago. It 
was shown~\cite{FR1989,FieldRichardson1990,FR1992} that generic branching was always along axes of symmetry (`axial' in the terminology of~\cite{GS}) and a complete
classification of the (signed indexed) branching patterns was obtained~\cite[\S 16]{FR1992}.  A  feature of generic bifurcation 
is that all the non-trivial branches of solutions consist of hyperbolic saddles ($\text{index}\ne 0, k-1$). In particular, if the trivial
branch of sinks loses stability, no branches of sinks or sources will appear post-bifurcation, other than the trivial branch of 
sources generated by the change in stability of the trivial solution.  This applies if bifurcation occurs on a centre manifold locally $S_k$-equivariantly
diffeomorphic to  $H_{k-1}$. If all the transverse directions are (say) attracting, then we see a transition from hyperbolic saddle to hyperbolic sink.
More can be said, but first we 
recall~\cite{FR1992,Field1989} the equations for generic $S_k$-equivariant bifurcation on $H_{k-1}$. 
\begin{eqnarray}
\label{eq: odd}
\xx' & = & \lambda \xx + Q(\xx)\;\; \text{($k$ odd)}\\
\label{eq: even}
\xx' & = & \lambda \xx + Q(\xx) + T(\xx)\;\; \text{($k$ even)}
\end{eqnarray}
Here $Q\ne 0$ is a (any) homogeneous quadratic equivariant gradient vector field, and $T$ is the gradient of a homogeneous quartic $S_k$-invariant which is not identically zero
on the $S_k$-orbit of one special axis of of symmetry (details are given later).  Perhaps surprisingly, pre-bifurcation ($\lambda < 0$) \emph{all indices of branches are at least $[k/2]$}; post bifurcation,
all indices are \emph{at most} $[k/2]$ (equality with $k/2$ occurs iff $k$ is even). A consequence is that the high-dimensional stable manifolds of the saddle branches pre-bifurcation make the attracting trivial 
solution branch increasingly invisible to trajectories initialized far from the origin as $\lambda \arr 0-$. Similar remarks hold post-bifurcation. 
If $k$ is odd, there are $2^{k}-2$ non-trivial branches of solutions; half  of these are backward, and half are forward (two branches are associated to each of the $2^{k-1}-1$ axes of symmetry). 
A slightly more complicated formula, depending on the cubic term, can be given when $k$ is even---see Remarks~\ref{rems: coll}(1). 

Although (\ref{eq: odd},\ref{eq: even}) have been used as a starting point for physical models (for example, the work of Stewart, Elmhirst \& Cohen on speciation~\cite[Chap.~2, \S2.7]{GS}), 
from the point of view of local bifurcation and dynamics, 
the lack of any non-trivial branches of sinks (or sources) limits the use of (\ref{eq: odd},\ref{eq: even}) as general or universal models for bifurcation and dynamics. 
Of course, terms of odd degree may be added, such as $-c\|\xx\|^{2p}\xx$, where $c > 0$, $p \in \pint$, to
create sinks but these are far from the origin of $H_{k-1}$ and not part of the bifurcation at $\lambda = 0$, $\xx = 0$.
If $k$ is small, consideration of 
secondary bifurcation, mode interactions and unfolding theory can be effective tools for the analysis of specific problems (for example,~\cite[Chap.~XV,\S4]{GSS1988}).
For our applications, $k$ will typically not be small.

The approach in this paper to generic steady-state bifurcation on representations of $S_k$, in particular the standard representation of $S_k$, has a different perspective. 
Thus, we regard the generic $S_k$-equivariant steady-state bifurcation, as realized by the equations  (\ref{eq: odd},\ref{eq: even}), as 
encoding the solution to a complex problem related to the creation of
spurious minima in non-convex optimization.  Roughly speaking, as we increase $k$ in these problems, we see the formation of spurious minima. These do \emph{not} arise 
from bifurcation of the global minima. Careful analysis reveals that, in the symmetric case, the spurious minima---at least those seen in the numerics with the appropriate
initialization scheme (cf.~Xavier initialization~\cite{SafranShamir2018},\cite[1.2]{ArjevaniField2021a})---arise through a steady-state bifurcation along 
a copy of the standard representation of $S_k$ (for example, using a centre manifold reduction). Thus the change in stability, in the symmetric case, occurs through the simultaneous collision at the origin of 
$O(2^k)$ hyperbolic saddle points of low index resulting in a branch of spurious minima (directions transverse to the centre manifold are assumed contracting)---in (\ref{eq: odd},\ref{eq: even}) 
this amounts to \emph{decreasing $\lambda$}.
For example, the type II spurious minima described in~\cite{ArjevaniField2021a} appear at $k \approx 5.58$.  Ignoring for a moment the inconvenient detail that $k$ is an integer, representing the number of neurons,
the local mechanism for creation or annihilation of minima via generic bifurcation on the standard representation of $S_k$ should be clear (vary $\lambda$ in (\ref{eq: odd},\ref{eq: even})). 
The argument applies to other irreducible representations of $S_k$ that have quadratic equivariants  (cf.~\cite{IG1984},\cite{CLM1990}; for example, 
external tensor products of standard representations of the symmetric group).
For spurious minima that do not appear in the numerics with Xavier initialization, bifurcation along the exterior square representation of the standard representation of $S_k$
may occur. While this representation does not have quadratic equivariants, the mechanism for creation of spurious minima appears similar to that of type II
minima.  

In practice, rigorous analysis is carried out on fixed point spaces of the action. For a large class of isotropy groups, the associated fixed points spaces
have dimension \emph{independent} of $k$ and the bifurcation equations, restricted to the fixed point space, depend smoothly on $k$, now viewed as a \emph{real} parameter 
(see~\cite{ArjevaniField2021a} where power series in $1/\sqrt{k}$ are obtained for families of critical points and the concluding comments section).

We indicated above that the generic $S_k$-equivariant steady-state bifurcation could be viewed as encoding the solution to the problem of the creation of spurious minima. To gain insight into the general
problem (no symmetry), it is necessary  to describe what happens when we break the symmetry of the model (forced symmetry breaking).
We do this by introducing the notion of a \emph{minimal model} (of forced symmetry breaking). When $k $ is odd, this is an explicit local symmetry breaking perturbation of the equations
(\ref{eq: odd}) to a $C^2$-stable family $\mathcal{F}=\{F_\lambda\dd \lambda \in \real\}$ (stable within the space of asymmetric families) which has \emph{minimal dynamic complexity}. The minimal 
complexity is described in terms of the \emph{minimum} number of saddle-node bifurcations and the \emph{maximum} number of hyperbolic solution curves (defined for $\lambda\in\real$) that the family $\mathcal{F}$ 
must have. For example, if $k = 17$, the minimal model will have exactly 52,666 saddle-node bifurcations and 12,870 hyperbolic solution curves. The minimal model is indicative of the complex landscape geometry that is
involved in the creation of asymmetric spurious minima in non-convex optimization in neural nets (cf.~\cite{ArjevaniField2020b,ArjevaniField2021d}). A similar result holds for $k$ even---on account of the cubic
terms in (\ref{eq: even}), the standard model is defined slightly differently so that no solutions are introduced which are unrelated to the bifurcation. 
In either case, there is a (small) interval of values of $k$ for which there are \emph{no} sinks or sources---a reflection of the previously noted relative 
``invisibility'' of the trivial sink or source near the bifurcation point of the $S_k$-equivariant problem.

\subsection{Outline of paper and main results}
Parts of this article have posed expositional problems on account of missing literature references and foundational definitions. The most important of these issues is the absence in
much of the equivariant bifurcation theory reference literature of the definition of a \emph{solution branch} (for example, \cite{GSS1988,GS}). 
We would argue that this definition should be a key foundational concept in the theory. 
In~\cite{GSS1988,GS}, the default is that of an \emph{axial} solution branch---bifurcation along an axis of symmetry (for example, \cite[\S 2]{GSS1988}).
The existence of axial solution branches (generically always smooth if the underlying family is smooth) uses Vanderbauwhede's version of the equivariant branching lemma~\cite{V} and, mathematically speaking,
the analysis of axial branches (for finite group symmetries) is elementary and depends only on the implicit function theorem (used in the proof of the equivariant branching lemma). However, 
as has been shown many times~\cite{IG1984,FR1989,CLM1990,AGK,FR1992,Mel1994,L2015}, generic branches of solutions in equivariant bifurcation 
theory are typically not axial, even if they are of maximal isotropy, and/or branches of sinks or sources, and/or the
family consists of gradient vector fields.  Related to this problem of definition is the matter of quadratic equivariants. 
Ihrig \& Golubitsky showed that, under certain conditions, steady-state bifurcation 
on an absolutely irreducible representation with non-trivial quadratic equivariants was unstable~\cite[Th.~4.2(B)]{IG1984} (no branches of sinks). Their ``crucial hypothesis'' (H4)~\cite[p.~20]{IG1984} was that branches were axial. 
Later Chossat \emph{et al.}~showed the result of Ihrig \& Golubitsky applied without any restriction on isotropy type~\cite[Theorem 4.2(b)]{CLM1990}. Their proof is elementary except for one detail (see below).
Unfortunately, their result is not mentioned in~\cite[\S 2.3]{GS} and was unknown (or forgotten) by us when we began work on this paper. 

The definition of solution branch first appears in~\cite[\S 2]{FieldRichardson1990} and holds for generic bifurcation---specifically, for an open dense set of 1-parameter families ($C^\infty$-topology). 
The proof of genericity is not hard but depends on non-trivial equivariant transversality arguments~\cite{Field1989}.
The authors of~\cite{CLM1990} were probably unaware of this definition and instead used an approach based on the Curve Selection Lemma~\cite{Milnor1968} which gives the result for
real analytic families but not smooth families: the Curve Selection Lemma holds 
for semianalytic families (most generally, sub-analytic families~\cite{Slo95}) but does not extend to smooth families.  One possible way to extend the result of 
Chossat \emph{et al.} to smooth families is to use stability and determinacy 
results for stable families~\cite{Field1989,FieldM1996,Field2007}, but these use non-trivial (and difficult) results of Bierstone on equivariant jet transversality~\cite{bier2}. A simpler and
more attractive approach is to use the definition of solution branch (see below).

So as to clarify the foundations, Section~\ref{sec: gen} 
includes the key definitions of \emph{solution branch} and (signed, indexed) 
\emph{branching pattern} (Section 2.4), and a statement of the stability theorem 
(Section 2.6), with brief commentary on the proof. 

In Section~\ref{sec: quad}, a simple proof is given of the result of
Chossat \emph{et al.} on quadratic equivariants that only uses the natural notion of a solution branch, rather than arguments invoking the 
Curve Selection Lemma. The main result of the section is expressed in terms of branching patterns 
and  hyperbolic branches of solutions  rather than unstable branches~\cite{IG1984,CLM1990}.  We state the result 
below only for gradient vector fields (the
result extends to compact Lie groups and, subject to a condition, to non-gradient quadratic equivariants).\\

\noindent {\bf Theorem} 
\emph{Let $(V,G)$ be an absolutely irreducible representation of the finite group $G$ and assume there are non-zero quadratic equivariants, all of which are gradient
vector fields. Then for all stable families, every non-trivial branch of solutions is a branch of hyperbolic saddles
with  index lying in $[1, \text{\em dim}(V)-1]$.
Generically, therefore, there are no non-trivial branches of sinks or sources.}\\

Also discussed are recent developments in stratification 
theory giving stability of initial exponents, using the \emph{regular arc-wise analytic} stratification of Parusi\'nski and P\u{a}unescu~\cite{PaPa2017}, 
and perturbation theory estimates that apply if vector 
fields are gradient and analytic. We conclude Section~\ref{sec: quad} with a brief description of open questions about analytic parametrization of solution branches.

In Section~\ref{sec: MMFSB}, we define the notion of a \emph{minimal model of forced symmetry breaking}, emphasizing the case of the standard representation of
$S_k$ on $H_{k-1}$. After reviewing the classification of the signed indexed branching patterns for the standard representation of $S_k$~\cite[\S 16]{FR1992}, we construct minimal
models of forced symmetry breaking for the cases $k$ odd and even. 
For this introduction the emphasis is on the case $k$-odd since the standard model is simple and
given by~\Refb{eq: odd}. If $k$ is even, 
the construction is more complicated on account of the presence of pitchfork bifurcations (along axes of symmetry with isotropy conjugate to $S_{k/2} \times S_{k/2}$); these bifurcations
result from the cubic terms in \Refb{eq: even}. The standard model is now smooth (not analytic) and has no extra solutions forced by the presence of cubic terms in ~\Refb{eq: even}.
In either case, the symmetry breaking is local, supported on
arbitrarily small neighbourhood of the bifurcation point, and the family we construct is stable under $C^3$-small non-equivariant perturbations. 
The model family is also $S_{k-1}$-equivariant---this is important for the non-elementary part of the proof. Before giving the definition, we need some notation.

We adopt the convention that if $f$ is a smooth family with $f_\lambda(\xx) = f(\xx,\lambda) = \lambda \xx + F(\xx)$, $(\xx,\lambda) \in H_{k-1} \times \real$, then $J^p(f)$ is the Taylor polynomial of degree $p$ of $f_0$  
at the origin $\is{0} \in H_{k-1}$ ($F$ may be assumed independent of $\lambda$, see Section~\ref{sec: stabthm}). \\
\noindent {\bf Definition} (Standard representation of $S_k$, $k \ge 3$.) Let $f$ be a stable family with
initial terms given by \Refb{eq: odd} (resp.~\Refb{eq: even}).  
The family $\hat f$ is a \emph{minimal symmetry breaking model} for $f$ if
\begin{enumerate}
\item $J^p(f) = J^p(\hat f)$, where $p=2$ (resp.~$p=3$) if $k$ is odd (resp.~even).
\item The solution set of $\hat f$ consists of
\begin{enumerate}
\item Exactly $\binom{k-1}{[k/2]}$ \emph{crossing curves} (curves of hyperbolic equilibria defined for $\lambda \in \real$).
\item Exactly $2^{k-1} - \binom{k-1}{[k/2]}$ \emph{saddle-node bifurcations}---all other solutions are hyperbolic.
\end{enumerate}
\item $ \hat f$ is a stable family: sufficiently small perturbations of $\hat f$, supported on a compact neighbourhood of 
$(\is{0},0)\in H_{k-1}\times \real$, preserve (1,3) (perturbations are not assumed equivariant).
\end{enumerate}

We provide proofs that the notion of minimal symmetry breaking model is well-defined and, given any open neighbourhood $W$ of $(\is{0},0)\in H_{k-1}\times \real$, can be realized by 
a $C^2$-small perturbation supported in $W$ of the 
model~\Refb{eq: odd} ($k$-odd) or a $C^3$-small perturbation of the standard model ($k$ is even). Although there are many details, most of the proof is elementary with the exception
of the argument showing no new solutions are introduced. This uses results from~\cite{FR1992a}, \cite[\S 4.9]{Field2007} which depend on Bezout's theorem and the pinning of solutions
to the complexification of fixed point spaces.  Full statements of the results appear in Section~\ref{sec: MMFSB}: Theorems~\ref{thm: uodd}, \ref{thm: ueven}.

Although we have not encountered past work on minimal models, it would be surprising if the phenomenon had not been noticed before. 

In the concluding comments, we return to the original motivating problem about the creation and annihilation of spurious minima, indicate how the results of the paper
can be used to understand this phenomenom, and discussed related current and proposed developments.

\section{Generic equivariant bifurcation}\label{sec: gen}
\subsection{Preliminaries and notation}
Let $\pint$ denote the natural numbers---the strictly positive integers---and $\intg$ the set of all integers.
Given $k \in \pint$, define  $\is{k}=\{1,\ldots, k\}$ (so that 
$S_k$ is the symmetric group of permutations of $\is{k}$).
The symbols $\is{k}, \is{m}, \is{n},\is{p}, \is{q}$ are reserved for indexing. For example,
$
\sum_{i=1}^m\sum_{j=1}^n a_{ij} = \sum_{(i,j) \in \is{n}\times \is{m}} a_{ij}$,
otherwise boldface lower case (resp.~upper case) is used to denote vectors (resp.~matrices). 

Some familiarity with the definitions and results of steady-state equivariant
bifurcation theory is assumed; we refer to~\cite{Field2007} for more details. 
The books \cite{GSS1984,GSS1988,CL2000,GS} provide an introduction to aspects of equivariant bifurcation theory and its applications but the 
methods and focus are different from what is required here.
\subsection{Representations} 

Let $V$ be a finite dimensional real vector space with inner product $\langle\,\,,\,\rangle$ and norm $\|\;\|$. 
Let $\On{V}$ denote the orthogonal group of $V$. If $\text{dim}(V) = m$, we often identify $V$ with Euclidean space $\real^m$ and
$\On{V}$ with $\On{m}$ (group of orthogonal $m \times m$ matrices).  

Given a finite\footnote{Most of what we say applies to compact Lie groups but the prerequisites  and technical details are harder. See~\cite[Chap.~10]{Field2007}.}  
group $G$ acting orthogonally on $V$, we refer to $(V,G)$ as an \emph{orthogonal representation} of $G$ on $V$. Usually, we just
say $(V,G)$ is a \emph{representation} of $G$ and assume orthogonality.  The representation is \emph{trivial} if each element of $G$ acts as the identity on $V$, and is
\emph{irreducible} if there are no proper $G$-invariant subspaces of $V$. A linear map
$A: V \arr V$ is a \emph{$G$-map} iff $A(g \vv) = g A(\vv)$ for all $g \in G$, $\vv \in V$. If
$A$ is a $G$-map then $\text{kernel}(A)$ and $\text{image}(A)$ are $G$-invariant linear subspaces of $V$. Consequently, if $A:V \arr V$ is a $G$-map and
$(V,G)$ is irreducible, then $A$ is either the zero map or a linear isomorphism---the orthogonal complement of $\text{image}(A)$ is $G$-invariant and so
if $A \ne 0$, then $A$ must be onto; a similar argument using $\text{kernel}(A)$ shows $A$ is 1:1. 
\begin{rem}\label{rem: gmap}
A $G$-map is \emph{$G$-equivariant} but we prefer the term  $G$-map when dealing with representations and linear maps. 
\rend
\end{rem}
\begin{Def}
An irreducible representation $(V,G)$ is of \emph{real type} or \emph{absolutely irreducible} if the set of $G$-maps consists of all real
multiples of the identity map $I_V$ of $V$.
\end{Def}
\begin{rems}\label{rem: defs}
(1) In representation theory, the term real representation is most commonly used but 
may be confusing here since all vector spaces are real. We follow the conventions in the bifurcation literature
and use only the term absolutely irreducible. \\
(2) To avoid uninteresting special cases, an absolutely irreducible representation is always assumed non-trivial. \rend
\end{rems}
\begin{exam}[Representations of the symmetric group]
Every nontrivial irreducible representation of $S_k$ is absolutely irreducible~\cite{James1978,FultonHarris1991}.  In particular,  the
\emph{standard representation} $\mathfrak{s}_k$ of $S_k$, $k \ge 2$, on $H_{k-1} = \{\xx \in \real^k\dd\sum_{i\in\is{k}}x_i = 0\}\subset \real^{k}$, where $S_k$ acts by
permuting coordinates: $\sigma (x_1,\cdots,x_k) = (x_{\sigma^{-1}(1)},\cdots,x_{\sigma^{-1}(k)})$, $\sigma \in S_k$.  Let $\mathfrak{s}_k$ denote the
isomorphism class of the representation $(H_k,S_k)$ and $\mathfrak{t}$ denote the isomorphism class of the trivial representation $(S_k,\real)$, omitting the subscript
$k$. Thus the isomorphism class of $(\real^k,S_k)$ is $\mathfrak{s}_k+\mathfrak{t}$. \examend
\end{exam}
\subsection{Families of equivariant vector fields}\label{sec: families}
We often omit the prefix `$G$' from $G$-equivariant (or $G$-invariant) maps if no ambiguity results. 

Let $(V,G)$ be absolutely irreducible. A \emph{family} (strictly, $1$-parameter family) of equivariant vector fields on $V$ is a smooth ($C^\infty$)  equivariant
map $f: V \times \real \arr V$, where the action on $V \times \real$ consists of the given action on $G$ and the trivial action on $\real$. For $\lambda \in \real$, define
the equivariant vector field $f_\lambda: V \arr V$ by $f_\lambda(\vv) = f(\vv,\lambda)$, $\vv \in V$.   We denote the $V$-derivative of $f_\lambda$ at $\vv \in V$ by
$Df_{\lambda,\vv}:V\arr V$ and the derivative of $f$ at $(\vv,\lambda)$ by $Df_{(\vv,\lambda)}:V\times\real\arr V$. 
Both $Df_\lambda : V \arr L(V,V)$ and $Df:V\times\real \arr L(V\times \real,V)$ are $G$-equivariant. For example,
$Df_{\lambda,g\vv}g = gDf_{\lambda,\vv}$, $g \in G$, $\vv \in V$ (see Lemma~\ref{lem: equiv}).
\begin{rem}
Maps and families are assumed $C^\infty$. Differentiability requirements can be relaxed though this can be non-trivial~\cite{FieldM1996}. For our
main application to the standard representation of $S_k$, $C^3$ suffices. \rend
\end{rem}

By equivariance, $f_\lambda(\is{0}) = \is{0}$ for all $\lambda \in \real$ (note Remarks~\ref{rem: defs}(3)). Since $(V,G)$ is absolutely irreducible,
$Df_{\lambda,\is{0}} = \sigma(\lambda)I_V$, where $\sigma : \real \arr \real$ is $C^\infty$. The equilibrium $\is{0}$ of $f_\lambda$ will be 
hyperbolic iff
$\sigma(\lambda) \ne 0$. We assume that $\sigma(0) = 0$, implying the possibility of bifurcation at $\lambda = 0$,  and
make the generic assumption on $f$ that
$\sigma'(0) \ne 0$. After a reparametrization, we may assume $\sigma(\lambda) = \lambda$ for $\lambda$ near zero. Since our interest is in 
bifurcation at $\lambda = 0$, it is no loss of generality to assume 
\begin{equation}\label{eq: fam1}
f(\vv,\lambda) = \lambda \vv + F(\vv,\lambda),\;\text{for all } (\vv,\lambda) \in V \times \real,
\end{equation}
where $F$ is equivariant, $C^\infty$, and $DF_{\lambda,\is{0}} = \is{0}$, $\lambda\in\real$. Thus bifurcation of the trivial solution can only occur at $\lambda = 0$. 
Let $\mathcal{V}(V,G) = \mathcal{V}$ denote the space of all families $f$ satisfying~\Refb{eq: fam1}. Families $f, f' \in \mathcal{V}$
are $C^r$-close on a compact $K\subset V \times \real$ if the derivatives of $f$ and $g$ of order at most $r$ are close on $K$. 
If we define the semi-norm $\|\;\|_{K,r}$ on $\mathcal{V}$ by
\[
\|f\|_{K,r} =\max_{0 \le s \le r} \sup_{(\lambda,\vv)\in K}\|D^sf_{(\vv,\lambda)}\|,\; f \in \mathcal{V}.
\]
then the set of semi-norms $\|\;\|_{K,r}$, where $K$ runs over all compact subsets of $V \times \real$, defines the 
the (weak) $C^r$-topology on $\mathcal{V}$. Write $\mathcal{V}^r(V,G)=\mathcal{V}^r$ for 
$\mathcal{V}$ equipped with the $C^r$ topology ($1 \le r \le \infty$). Since the results we need are local, the
Whitney $C^r$-topology is not required.  Later, we use the semi-norm
$\|f\|^0_{K,r} =\max_{0 \le s \le r} \sup_{(\vv,\lambda)\in K}\|D^sf_{\lambda,\vv}\|$ which uses no $\lambda$-derivatives.

\subsection{Branches of solutions}
We need to review the core notions of \emph{solution branch} and \emph{branching pattern}.
A brief overview may be found in~\cite{FieldRichardson1990}; more detail is in~\cite{Field2007} and the original 
papers~\cite{FR1992a,FR1992}.  
\begin{Def}[{cf.~\cite[\S 4.2]{Field2007}}]\label{def: branch}
A \emph{solution branch} of \Refb{eq: fam1} consists of a $C^1$-embedding $\gamma= (\xx,\lambda): [0,\delta] \arr V \times \real$ satisfying
\begin{enumerate}
\item $\gamma(0) = (\is{0},0)$.
\item $f(\gamma(s))=f_{\lambda(s)}(\xx(s)) = \is{0}$, for all $s \in [0,\delta]$.
\end{enumerate}
If we can choose $\delta > 0$ so that
\begin{enumerate}
\item[(a)] $\xx\not\equiv \is{0}$ on $(0,\delta]$, the branch is \emph{non-trivial}.
\item[(b)] $D f_{\lambda(s),\xx(s)}$ is non-singular for $s \in (0,\delta]$, the branch is \emph{non-singular}.
\item[(c)]  $\lambda(s) > 0$ (resp.~$\lambda(s) < 0$) for $s \in (0,\delta]$, the branch is \emph{forward} (resp.~\emph{backward}).
\item[(d)] $\xx(s)$ is a hyperbolic zero for $s \in (0,\delta]$, the branch is \emph{hyperbolic} (necessarily non-singular).
\end{enumerate}
\end{Def}
Recall that the \emph{index} of a hyperbolic equilibrium $\xx$ of $X$  is the number of eigenvalues of $DX_\xx$ with strictly negative real part (counting
multiplicities) and is denoted by $\text{index}(X,\xx)$. 

The family~\Refb{eq: fam1} has two \emph{trivial} solution branches $\tau_\pm$ defined by
\[
\tau_\pm(s) = (\is{0},\pm s), \; (s \in \real),
\]
and $\tau_+$ (resp.~$\tau_-$) is a hyperbolic forward (resp.~backward) branch of index zero (resp.~$\text{dim}(V)$). 

Solution branches $\gamma, \rho$ are \emph{equivalent} if (roughly) the germs of the images of $\gamma$ and $\rho$ at $(\is{0},0)$
are equal. More precisely, if there is a $C^1$ diffeomorphism $\alpha:[0,\vareps_1]\arr[0,\vareps_2]$, mapping $0$ to $0$, such that
$\gamma \alpha = \rho$ on $[0,\vareps_1]$. We denote the equivalence class of $\gamma$ by $[\gamma]$ and let $\Sigma(f)$ (resp.~$\Sigma^\star(f)$) denote the set of all
equivalence classes of solution (resp.~non-trivial solution) branches for $f$. Clearly, $\Sigma(f)$ and $\Sigma^\star(f)$  are  $G$-sets and $[\tau_\pm]\in\Sigma(f)$ are the fixed points of the $G$-action on $\Sigma(f)$.

\begin{lemma}\label{lem: db}
Let $\gamma= (\xx,\lambda):[0,\delta] \arr V \times \real$ be a non-trivial solution branch for $f \in\cal{V}$.
\begin{enumerate}
\item The \emph{direction of branching} $\db(\gamma) = \xx'(0)/\|\xx'(0)\|\in V$ is well-defined, non-zero and independent of the parametrization. 
\item If $\gamma$ is a non-singular branch, $\gamma$ is either forward or backward.
\item If $\gamma$ is hyperbolic, then $\text{\rm index}(f_{\lambda(s)},\xx(s))$ is constant on $(0,\delta]$.
\end{enumerate}
\end{lemma}
\proof See~\cite[\S4.2]{Field2007},\cite{FR1992a} for the elementary proof (for (1), note that if $\xx'(0) = \is{0}$, then $\|\xx(s)\| = o(s)$ and so, using~\Refb{eq: fam1},
$|\lambda(s)| = o(s)$, contradicting the $C^1$-embedding requirement on $\gamma$). \qed

\begin{Def}\label{def: pat}
Let $f \in \mathcal{V}$ and suppose that $\Sigma(f)$ is finite and consists of hyperbolic solution branches. The \emph{signed indexed branching pattern}  of
$f$ is the triple $(\Sigma^\star(f), \sgn,\text{index})$ where
\begin{enumerate}
\item $\sgn: \Sigma^\star(f) \arr \{-1,+1\}$ and $\sgn([\gamma]) = +1$ (resp.~$-1$) if $\gamma$ is a forward (resp.~backward) solution branch ($\sgn$ is the \emph{sign} function).
\item $\text{index}: \Sigma^\star(f) \arr \{0,\cdots,\text{dim}(V)\}$ and $\text{index}([\gamma])$ is the index of $Df_{\gamma(s)}$, $ s \ne 0$.
\end{enumerate}
\end{Def}
\begin{rem}
The sign and index functions are $G$-invariant. 
\rend
\end{rem}
\begin{Def}
If 
$(\Sigma^\star(f_i), \sgn_i,\text{index}_i)$, $i \in \is{2}$, are signed indexed branching patterns, they are \emph{isomorphic} if there is a $G$-equivariant bijection 
$\beta: \Sigma^\star(f_1) \arr \Sigma^\star(f_2)$ such that $\sgn_1 = \sgn_2\circ \beta$ and $\text{index}_1 = \text{index}_2\circ \beta$.
\end{Def}
\begin{rem}\label{rem: core}
Since the general theory develops from Definitions~\ref{def: branch} and \ref{def: pat}, 
it is essential to prove that generic bifurcation can be expressed in terms of solution branches and
branching patterns. In particular, 
solution branches are defined in terms of $C^1$-embeddings (not $C^0$ or $C^\infty$), and a branching pattern
is a \emph{finite} union of solution branches. 
The proof requires ideas from the geometry and stratification of 
semialgebraic sets and equivariant transversality.  
\rend
\end{rem}

\subsection{Stable and weakly stable families}
\begin{Def}
A family $f \in \mathcal{V}$ is \emph{stable} if
\begin{enumerate}
\item $\Sigma(f)$ is finite and consists of hyperbolic solution branches (necessarily, either forward or backward).
\item For some $r \ge 1$, there is a neighbourhood $U$ of $f \in \mathcal{V}^r$ such that if $(\mathbf{f}_t)_{t \in [0,1]}$ is a continuous curve in $U$ 
with $\mathbf{f}_0 = f$, then
\begin{enumerate}
\item There exists $\delta > 0$ such that for all $[\gamma] \in \Sigma(f)$, there is a continuous family $(\gamma_t)_{t \in [0,1]}$ of $C^1$-maps $[0,\delta] \arr V \times \real$ such that
each $\gamma_t$ is a branch of hyperbolic zeros of $\mathbf{f}_t$ and $[\gamma_0] = [\gamma]$.  
\item $\Sigma^\star(f)$ and $\Sigma^\star(\mathbf{f}_t)$ are isomorphic for all $t \in [0,1]$. 
\end{enumerate}
\end{enumerate}
Denote the set of stable families by $\mathcal{S}(V,G) = \mathcal{S}$.
\end{Def}
\begin{rems}
(1) It follows from 2(b) of the definition that $\Sigma(h)$ is isomorphic as a $G$-set to $\Sigma(f)$ for all $h$ in the path connected component of $U$ containing $f$. 
Similarly, using 2(a), the signed index branching patterns for $\Sigma^\star(h)$ and $\Sigma^\star(f)$ are isomorphic. \\
(2) If $f$ is stable and $[\gamma], [\eta] \in \Sigma^\star(f)$, $[\gamma]\ne[\eta]$, then we cannot exclude the possibility that
$\db([\gamma]) = \db([\eta])$ but this does not happen if
the stability is determined by quadratic or cubic terms (for example,
if quadratic terms are of relatively hyperbolic type~\cite[\S 4.6.4]{Field2007}).  \rend
\end{rems}

We also need the concept of weak stability~\cite[\S 4.2.1]{Field2007}.
\begin{Def}
A family $f \in \mathcal{V}$ is \emph{weakly stable} if
\begin{enumerate}
\item $\Sigma(f)$ is finite.
\item For some $r \ge 1$, there is a $U$ of $f \in \mathcal{V}^r$ such that if $(\mathbf{f}_t)_{t \in [0,1]}$ is a continuous curve in $U$ 
with $\mathbf{f}_0 = f$, then
\begin{enumerate}
\item There exists $\delta > 0$ such that for every $[\gamma] \in \Sigma(f)$, there is a continuous family $(\gamma_t)_{t \in [0,1]}$ of $C^1$-maps $[0,\delta] \arr V \times \real$ such that
each $\gamma_t$ is a solution branch of $\mathbf{f}_t$ and $[\gamma_0] = [\gamma]$.  
\item $\Sigma(f)$ and $\Sigma(\mathbf{f}_t)$ are isomorphic as $G$-sets, $t \in [0,1]$. 
\end{enumerate}
\end{enumerate}
Denote the set of weakly stable families by $\mathcal{K}(V,G)$.
\end{Def}

\subsection{The stability theorem}\label{sec: stabthm}
Given $d \in \pint$, let $P_G^{(d)}(V,V)$ denote the space of $G$-equivariant homogeneous polynomials from $V$ to $V$ of degree $d$. For $d > 1$, define
$\widehat{P}_G^d(V,V) = \oplus_{i = 2}^d P_G^{(i)}(V,V)$---equivariant polynomial maps of degree at most $d$ with zero linear part.  
If $f \in \mathcal{V}$, let $J^d(f) = j^d f_0(0) \in \widehat{P}_G^d(V,V)$ denote the $d$-jet of $f_0$ at $0$ (Taylor polynomial of degree $d$ of $f_0$ at $0$). 
\begin{thm}[{\cite{Field1989}}]\label{thm: stable}
(Assumptions and notation as above.) There exist a minimal $\delta=\delta(V,G)  \in \pint$ and an  open and dense semi-algebraic subset
$\mathcal{P}^\delta = \mathcal{P}$ of $\widehat{P}_G^\delta(V,V)$ such that 
if $f \in \mathcal{V}$ and $J^\delta(f) \in \mathcal{P}^\delta$, then $f\in\cal{S}$. In particular, $\cal{S}$ contains a $C^\delta$-open and dense subset of $\mathcal{V}$.\\
Similar results hold for weak stability. 
subset of $\mathcal{V}$; typically with a smaller value $\delta_w$ of $\delta$.  
\end{thm}
\proof The result uses equivariant jet transversality~\cite{bier2}. Details may be found in~\cite[Chap.~7]{Field2007} or \cite{Field1989}; some brief notes are at the end of this section.
In Section~\ref{sec: wkset}, there is an outline proof for weak stability. This result addresses the  points raised in Remark~\ref{rem: core} and does not use equivariant jet transversality. 
\qed
\begin{rems}
(1) The result extends to absolutely irreducible representations of compact Lie groups~\cite[\S 7.6]{Field2007} and irreducible representations of complex type~\cite[Chap.~10]{Field2007}.\\
(2) There is an upper bound for $\delta(V,G)$. If $\{p_1 = \|\;\|^2,\ldots, p_\ell\}$ is a minimal set of homogeneous generators for the
$\real$-algebra  $P_G(V)$ of polynomial invariants on $V$ and  $\{F_1 = I_V,\cdots,F_k\}$ is a minimal set of
homogeneous generators for the $P_G(V)$-module $P_G(V,V)$ of polynomial maps of $V$, then $\delta(V,G) \le \max_i \text{deg} (p_i) + \max_j \text{deg} (F_j)$. 
Often $\delta$ can be chosen much smaller. If $(V,G) =\mathfrak{s}_k$, then
$\max_i \text{deg} (p_i) + \max_j \text{deg} (F_j) = 2k-1$, $k \ge 2$, but we can take $\delta = 2$, if $k$ is odd, and $\delta=3$ if $k$ is even. 
For weak stability, the corresponding minimal $\delta_w$ satisfies $\delta_w \le \max_j \text{deg} (F_j)$. \\
(3) Theorem~\ref{thm: stable} does not imply $f$ is stable only if $J^\delta(f) \in \mathcal{P}$. The resolution of this point is subtle as it depends on the
specific stratification used in equivariant transversality (see Section~\ref{sec: wkset}).\\
(4) Increasing $\delta$ will not change the space of stable maps given by the theorem. See the Section~\ref{sec: det} for this point. \rend
\end{rems}
We have a very useful corollary of the stability theorem.
\begin{cor}
Let $f \in \cal{V}$ and suppose $f(\vv,\lambda) = \lambda \vv + F(\vv,\lambda)$. Define $\hat{f}(\vv,\lambda) = \lambda\vv+F(\vv,0)$.
Then $J^\delta(f) \in \mathcal{P}$ iff $J^\delta(\hat{f}) \in \mathcal{P}$. If either condition holds, both families are stable and have isomorphic signed indexed branching patterns.
\end{cor}
\begin{rem}
The corollary allows us to work with polynomial families and  use methods based on the Curve Selection Lemma~\cite{Milnor1968}.
In this way we obtain \emph{real analytic parametrizations} of solution branches for stable polynomial families $\lambda \xx + P(\xx)$,
$P \in \mathcal{P}$. Results obtained using this approach may extend to stable smooth families and allow for the
sharp analytic estimates; for example on eigenvalues (see Section~\ref{sec: quad}). \rend
\end{rem}
Define the subspace $\mathcal{V}_0(V,G) = \mathcal{V}_0$ of $\mathcal{V}$ by requiring that $f \in \mathcal{V}_0$  iff 
$f(\vv,\lambda) = \lambda\vv + F(\vv)$.  Give $\mathcal{V}_0$ the $C^\infty$-topology defined by the semi-norms $\|\;\|^0_{K,r}$ (Section~\ref{sec: families}).
Let $\mathcal{S}_0(V,G) = \mathcal{S}_0 =\mathcal{S}\cap \cal{V}_0$ denote the set of stable families in $\cal{V}_0$.

\subsection{Determinacy}\label{sec: det}
Following the statement of Theorem~\ref{thm: stable}, equivariant bifurcation problems on $(V,G)$ 
are \emph{$\delta=\delta(V,G)$-determined}. This notion of determinacy is quite different from that used in~\cite{GSS1988}. 

We conclude with brief  
details about the constructions used to prove the stability theorem and determinacy, avoiding the technicalities of equivariant jet transversality.  

Let $p_1,\cdots,p_\ell$ be a minimal set of homogeneous generators for the $\real$-algebra $P_G(V)$. By Schwarz' theorem on smooth invariants~\cite[Chap.~6]{Field2007}, 
if $f \in \cal{V}$,  then
\[
f(\vv,\lambda) = \lambda \vv + \sum_{i\in\is{k}} g_i(p_1(\vv),\cdots,p_\ell(\vv),\lambda)F_i(\vv),
\]
where the $g_i : \real^\ell \times \real \arr \real$ are $C^\infty$ functions and $g_1(\is{0},\lambda) \equiv 0$ since $F_1=I_V$.
Setting $t_j = g_i(\is{0},0)$, $j \ge 2$, $\{t_j \dd j \ge 2\}$ is uniquely determined by our choice of generating set
$\{F_i \dd i \in \is{k}\}$~\cite[\S 6.6.2]{Field2007}. 
\begin{rem}\label{rem: gtrans}
The family $f$ is weakly stable provided that $(t_2,\cdots,t_k)$ avoids
a codimension 1 semi-algebraic subset of $\real^{k-1}$ (the branches may not be hyperbolic but do have a direction of branching and deform
continuously under perturbation of $f$). This result~\cite[Thm.~7.1.1]{Field2007} 
plays an important role in our applications where there are non-trivial quadratic equivariants.  \rend
\end{rem}
Given the coefficient functions $g_i$, we may write $f(\vv,\lambda)$ uniquely as
\[
f(\vv,\lambda) = \lambda \vv + \sum^k_{j=2} t_j F_j(\vv) + \sum_{(i,j) \in \boldsymbol{\ell}\times \mathbf{k}} t_{ij} p_i(\vv)F_j(\vv) + H(\vv,\lambda),
\]
where $t_j, t_{ij}$ are smooth functions of $\lambda$ and $H(\vv,\lambda)$ consists of higher order terms in the invariants. 
The conditions for stability that come from equivariant jet 
transversality depend only on  the  $k-1 + k \ell$ real numbers $t_j(0), t_{ij}(0)$.  Viewed in this way, once we have found
the minimum $\delta(V,G)$ (that depends on which of the $t_j(0), t_{ij}(0)$ do not affect the stability of the family), $\mathcal{P}^d$ is determined for all
$d \ge \delta(V,G)$ with $\mathcal{P}^d$ projecting naturally onto $\mathcal{P}^{d-1}$ for $d > \delta(V,G)$.

\section{Quadratic equivariants}
\label{sec: quad}

\begin{Def}
If $(V,G)$ is an absolutely irreducible representation, then 
$(V,G)$ has \emph{quadratic equivariants} if $\text{dim}(P_G^{(2)}(V,V)) \ge 1$. 
\end{Def}
\begin{exams}
(1) The standard representation $\mathfrak{s}_k$ of $S_k$ on $H_{k-1}$ has quadratic equivariants for $k \ge 3$ and $P^{(2)}_{S_k}(H_{k-1},H_{k-1})$ has basis
$\grad{C| H_{k-1}}$, where $C(\xx) = \frac{1}{3}\sum_{i \in \is{k}} x_i^3$, $\xx \in \real^k$.  \\
(2) Let $\widetilde{M}(k,n)$ (resp.~$M^\star(k,n)$) denote the space of real $k \times n$-matrices such that all rows and columns sum to zero (resp.~the sum of all matrix entries is zero). 
Obviously $\widetilde{M}(k,n)\subsetneq M^\star(k,n)$, if $k, n \ge 2$. 
The external tensor product representation $\mathfrak{s}_k \boxtimes \mathfrak{s}_n$ of $S_k \times S_n\subset S_{k \times n}$ on $H_{k-1} \otimes H_{n-1} \approx \widetilde{M}(k,n)$
is absolutely irreducible and has quadratic invariants iff $k,d \ge 3$. In order to show this, observe that the cubic $S_{k \times d}$-invariant
$C(\WW) = \sum_{(i,j)\in \is{k} \times \is{n}} w_{ij}^3$, $\WW = [w_{ij}]$, does not vanish identically on the subspace $\widetilde{M}(k,n)$ iff $k,d \ge 3$ (for this, it is enough to look at matrices in
$\widetilde{M}(3,3)\subset \widetilde{M}(k,n)$, $k,n \ge 3$).  
Hence $\grad{C|\widetilde{M}(k,n)}$ is a non-zero quadratic equivariant for $\mathfrak{s}_k \boxtimes \mathfrak{s}_n$.
With some additional work, it can be shown that the space of homogeneous cubic invariants on $\widetilde{M}(k,d)$ is $1$-dimensional
for all $k,d \ge 3$ with basis given by $C(\xx) = \frac{1}{3}\sum_{ij} x_{ij}^3$.  
For example, if $k = d = 3$, $\text{dim}(\widetilde{M}(3,3))=4$ and 
\[
C(x_1,x_2,x_3,x_4) = 2\sum_{1 \le i < j < k \le 4}x_ix_jx_k + x_1x_4 (x_1+x_4) + x_2 x_3(x_2+x_3),
\]
where if $X = [x_{ij}] \in \widetilde{M}(3,3)$, $x_1 = x_{11}$, $x_2 = x_{12}$, $x_3 = x_{21}$, $x_4 = x_{22}$,
and the remaining entries are determined by the row and column sum zero condition. The general formula is an easy induction.
\examend
\end{exams}

\subsection{Generic branching when there are quadratic equivariants}
We refer to Section 1.2 for background on the results of Ihrig and Golubitsky~\cite{IG1984}, and Chossat, Lauterbach, \& Melbourne~\cite{CLM1990}, on the instability of
branching when there are quadratic equivariants. In this section we reprove Theorem 4.2(b)~\cite{CLM1990}, using only the definition of solution branch, as well as 
prove a stronger version that uses Theorem~\ref{thm: stable}. 

We show that if the quadratic equivariants satisfy ``\emph{Property $(\mathcal{G})$}'', then the stable signed indexed branching patterns
consist of branches of hyperbolic saddles (no non-trivial branches of sinks or sources). 
$\propG$ always holds if the quadratic equivariants are gradient vector fields; indeed, the ``$\mathcal{G}$'' is short
for ``gradient like''.  

In what follows, $S(V)$ will denote the unit sphere of $V$.

\subsection{$\propG$}
Given $Q\in P_G^{(2)}(V,V)$, $Q \ne 0$, define $\isZ{Q} = \{\uu \in S(V) \dd Q(\uu) = 0\}$. Since $Q \ne 0$, $\isZ{Q}$ is a proper closed subset of $S(V)$.
Let $\cal{Q}_0$ be the set of $Q \in P_G^{(2)}(V,V)\smallsetminus \{0\}$ such that for all $\uu \in \isZ{Q}$, $DQ_\uu$ has an eigenvalue with non-zero real part.
\begin{Def}[{cf.~\cite[Theorem 4.2(B)]{IG1984}}]
If the absolutely irreducible representation $(V,G)$ has quadratic equivariants, then $(V,G)$ satisfies \emph{$\propG$} if $\cal{Q}_0$ 
is an open and dense subset of $P_G^{(2)}(V,V)$.
\end{Def}

\begin{exams}\label{ex: G}
(1) If every quadratic equivariant is a gradient vector field, then $(V,G)$ satisfies $\propG$ 
with $\cal{Q}_0 = P_G^{(2)}(V,V)\smallsetminus \{0\}$~\cite[Remarks 4.3(g)]{IG1984}.
This is obvious since if $Q = \grad{C} \ne 0$, for some $C \in P^{(3)}_G(V)$, then $DQ_\uu$ is a symmetric matrix for all $\uu \in S(V)$
and so, since $Q \ne 0$, $DQ_\uu$ has at least one non-zero real eigenvalue. \\
(2) If the only $Q\in P_G^{(2)}(V,V)$ for which $\is{Z}(Q) \ne \emptyset$ is $Q = 0$, then $(V,G)$ satisfies $\propG$.\\
(3) Absolutely irreducible representations may have quadratic equivariants which are not gradient.
For example, the group $G=\text{Aff}(\mathbb{F}_5)$ of affine linear transformations of the field with $5$-elements is isomorphic to the subgroup of
$S_5$ generated by $t=(12345)$ and $s=(2453)$~\cite[\S 5.4.2]{Field2007}. The group $G$ acts on 
$\real^4 \cong V = \{(z_1,z_2,\bar{z}_2, \bar{z}_1) \dd z_1,z_2 \in \cx\}$
by
\[
t(z_1,z_2) = (\omega z_1,\omega^2 z_2),\quad
s(z_1,z_2)  =  (\bar{z}_2,z_1),
\]
where $\omega = \exp(2\pi\imath/5)$~\cite[\S 5.4.1]{Field2007}. We find that $\text{dim}(P_G^{(2)}(V,V)) =2$ and has $\real$-basis,
$Q_1=(\bar{z}^2_2,z_1^2)$, $Q_2 = (\bar{z}_1z_2, \bar{z}_1\bar{z}_2)$.
It is easy to verify that $\alpha Q_1 + \beta Q_2$ is gradient iff $\beta=2 \alpha$. Note that if
$Q = Q_1-Q_2$, then the cubic invariant $\langle Q(z_1,z_2),(z_1,z_2)\rangle$ is identically zero. 

Computing we find that if $u_1 u_2 \ne 0$, then $DQ_{1,\uu}$ has two non-zero real and a complex conjugate imaginary pair of eigenvalues.
If $Q = \alpha Q_2$, $\alpha \in \real$, $\alpha \ne 0$, then $\isZ{Q} = \{(z_1,z_2) \dd z_1z_2 = 0\}$. 
Direct computation verifies that $DQ_{(z_1,z_2)}$ has
the pair $\pm \alpha \|(z_1,z_2)\|$ of non-zero real eigenvalues. If we set $Q^\pm = Q_1 \pm Q_2$, then $\isZ{Q} \ne \emptyset$ iff $Q \in \real Q^\pm \cup \real Q_2$,
Direct computation verifies that if $Q \in \real Q^\pm$, $Q \ne 0$, then $DQ_\uu $ has eigenvalues with non-zero real part for $\uu \in \isZ{Q}$. If $Q = Q^-$,
then $\uu \in \isZ{Q}$ iff $\uu$ lies on an axis of symmetry (the group orbit of 
$\real(1,0,1,0)$); if $Q = Q^+$, then $\uu \in \isZ{Q}$ iff $\uu$ lies on
the group orbit of $\real(1,0,-1,0)$. 
Hence $(\real^4,\text{Aff}(\mathbb{F}_5))$ satisfies $\propG$ and $\cal{Q}_0 =  P_G^{(2)}(V,V)\smallsetminus \{0\}$. \examend
\end{exams}
\begin{rem}
In Examples~\ref{ex: G}(3), there is a non-zero $Q \in P_G^{(2)}(V,V)$ satisfying $\langle Q(\vv),\vv \rangle = 0$, for all $\vv \in V = \real^4$. 
This non-gradient behaviour suggests there may well exist absolutely irreducible representations $(V,G)$ for 
which $\propG$ fails and the eigenvalues of $DQ_\uu$ along $\real \uu$ are all either zero or pure imaginary. A natural place to look 
is the work by Lauterbach and Matthews on low dimensional families of absolutely irreducible representations with no odd
dimensional fixed point spaces~\cite{L2015}. However, these families do not have quadratic equivariants and the question appears open. \rend
\end{rem}

\subsection{Statement of the main theorem}
\begin{thm}\label{thm: main}
If $(V,G)$ is an absolutely irreducible representation of the finite group $G$ satisfying $\propG$, then
for all $f \in \cal{S}$, $\text{\rm index}:\Sigma^\star(f) \arr [1, \text{dim}(V)-1]$.
In particular, every non-trivial branch is a branch of hyperbolic saddles and so there are no non-trivial branches of sinks or sources.
\end{thm}
\begin{rems}
(1) The result applies to all stable families, not just the open and dense set of stable families given by Theorem~\ref{thm: stable}.\\
(2) If the quadratic invariants vanish identically on a fixed point space $F$, then a backward branch lying in $G(F)$
will not be a branch of maximal index. This result is part (a) of Theorem 4.2 in Chossat \emph{el al.}~\cite{CLM1990}. If $\propG$ holds
(for example, if $Q$ is gradient) then the branch will be a branch of hyperbolic saddles by Theorem~\ref{thm: main}.  \\
(3) Let $f \in \cal{S}$, $[\gamma] \in \Sigma^\star(f)$.   
The direction of branching $\db(\gamma) = \uu \in S(V)$. Setting $Q = J^2(f)$,
$[\gamma]$ will be a branch of hyperbolic saddles if either (a) $Q(\uu) \ne 0$ or (b) $Q(\uu) = \is{0}$ and 
$DQ_\uu$ has an eigenvalue with non-zero real part.
The failure of $\propG$ only concerns solution branches which are tangent to $\real\uu$, $\uu \in \isZ{J^2(f)}$.
\rend
\end{rems}
\begin{Def}
Suppose $(V,G)$ has quadratic equivariants.
Let $f \in \mathcal{S}$ and set $J^2(f) = Q$. Suppose $[\gamma] \in \Sigma^\star(f)$ has direction of branching $\uu\in S(V)$.
If $\uu \notin \is{Z}(Q)$, $[\gamma]$ is a branch of \emph{type S}, otherwise $[\gamma]$ is a branch of \emph{type C}.
\end{Def}

\subsection{Weak stability and equivariant transversality}\label{sec: wkset}
Assume that $(V,G)$ is an absolutely irreducible representation of the finite group $G$ (no assumption yet about quadratic equivariants).
We review the use of equivariant transversality and Whitney regular stratifications in the proof of weak stability (for full details, see~\cite[Chaps.~6,7]{Field2007}).

\subsubsection{Equivariant transversality}
Fix a minimal homogeneous basis $\mathcal{F} = \{F_1 = I_V,F_2,\cdots, F_k\}$ of the $P_G(V)$-module of $G$-equivariant polynomials maps of
$V$. Set $d_i = \text{deg}(F_i)$ and label the polynomials $F_i$ so that $1 = d_1  < d_2\le \cdots \le d_k$. 

Define $\vartheta: V \times \real^k \arr V$ by
$\vartheta(\xx,\is{t}) = \sum_{i\in \is{k}} t_i F_i(\xx)$ and set $\vartheta^{-1}(\is{0}) = \Lambda$:
\[
\Lambda = \{(\xx,\mathbf{t}) \in V \times \real^k \dd \sum_{i\in \is{k}} t_i F_i(\xx) = \is{0}\} \subset V \times \real^k.
\]
Clearly, $\real^k, V \subset \Lambda$ (where $\real^k \defoo \{\is{0}\}\times \real^k$, $V \defoo  V \times \{\is{0}\}$). 
Since $G$ is finite, $\text{dim}(\Lambda) = k$ and $\text{dim}(V) \le k$~\cite[Remark~6.9.3]{Field2007}.

Let $f \in \cal{V}_0$ (arguments are similar if $f\in \cal{V}$). There exist $C^\infty$ invariant functions $g_i:V\arr\real$ such that
\[
f(\xx,\lambda) = \lambda \xx + \sum_{i\in\is{k}}g_i(\xx) F_i(\xx),\; (\xx,\lambda) \in V \times \real,
\]
where $g_1(\is{0}) = 0$ (see~\cite[\S 6.6]{Field2007} for the details which use the Malgrange Division theorem~\cite{Mal1966}). The maps $g_1,\ldots,g_k$ are 
generally not uniquely determined by $f$ but the values $g_i(\is{0})$, $2\le i \le k$, are uniquely determined~\cite[Lemma 6.6.2]{Field2007} and 
depend linearly and continuously on $f$, $C^\infty$-topology (cf.~\cite{Mather1977}).

Define the smooth equivariant embedding $\Gamma_f: V \times \real \arr V \times \real^k$ by
\[
\Gamma_f(\xx,\lambda) = (\xx,(\lambda+ g_1(\xx), g_2(\xx),\cdots,g_k(\xx)),\; (\xx,\lambda) \in V \times \real.
\]
The tangent space to $\Gamma_f(V \times \real)$ at $\Gamma_f(\is{0},0)\in V \times \real^k$ is $V \times \real \is{e} $, where 
$\is{e} = (1,0,\ldots,0)\in \real^k$.

The family $f$ factorizes through $V \times \real^k$ as $f = \vartheta \circ \Gamma_f$, and
$\Gamma_f(\is{0},0) \in \real^{k-1}\defoo\{(\is{0},\mathbf{t}) \in \real^k \subset V \times \real^k \dd t_1 = 0\}$. Define
$\gamma_f:\real \arr \real^k$ by $\gamma_f(\lambda) = (\lambda,g_2(\is{0}),\cdots,p_k(\is{0}))$. Clearly $\gamma_f = \Gamma_f | \{\is{0}\} \times \real$.

We recall some results on Whitney regular stratifications (see~\cite[\S\S 3.9,6.8]{Field2007} for basic definitions 
and results and~\cite{trott2020} for a recent review which includes an extensive bibliography).
Fix a Whitney regular semialgebraic stratification $\mathfrak{S}$ of $\Lambda$---for example, the canonical stratification~\cite{Mather1973}---and define
\[
\cal{K}_{0,\mathfrak{S}} = \{f \in \cal{V}_0 \dd \Gamma_f \trans \mathfrak{S} \;\text{at } (\is{0},0)\}.
\]
It follows by the isotopy theorem for equivariant transversality that $\cal{K}_{0,\mathfrak{S}}$ consists of
weakly regular families~\cite[Thm.~7.7.1]{Field2007}. 

It is easily shown~\cite[Thm.~6.10.1]{Field2007} that $\mathfrak{S}$ induces a stratification $\mathfrak{S}^\star$
of $\real^{k-1}$---the strata of $\mathfrak{S}^\star$ consist of the strata of $\mathfrak{S}$ which are subsets of 
$\real^{k-1}$ intersected, if necessary\footnote{See~\cite[Rem.~7.1.1]{Field2007} if $\exists$
an open $X\subset\real^{k-1}$ with $\Sigma^\star(f) = \emptyset$ when $\gamma_f(0) \in X$. At this time, no examples where this happens are known.},
with $\real^{k-1}$. We have $\Gamma_f \trans \mathfrak{S}$ at $(\is{0},0)$ iff $\gamma_f \trans \mathfrak{S}^\star$ at $\lambda = 0$.
\begin{rem}
The condition for $f$ to lie in $\cal{K}_{0,\mathfrak{S}}$ depends \emph{only} on the values of $g_2,\ldots,g_k$ at $\xx = 0$. The
same arguments apply if $f \in \cal{V}$ and this allows us to assume the coefficients $g_i$ do not depend on $\lambda$. \rend
\end{rem}

\subsubsection{$C^1$ parametrization of solution branches}
We need to examine the stratifications $\mathfrak{S}$ of $\Lambda$ and $\mathfrak{S}^\star$ of $\real^{k-1}$.
If $S \in \mathfrak{S}$ is a connected stratum of dimension $m$, then $\partial S$ will be a union of connected strata of dimension less than or equal to
$m - 1$ (this uses Whitney regularity). Let $\mathfrak{S}_0$ be the union of all connected strata $N$ of $\mathfrak{S}$ which are of dimension
$k$ and for which $\partial N$ has at least one connected stratum $M \in \mathfrak{S}^\star$ of dimension $k-1$ (so defining an open subset
of $\real^{k-1}$).  Let $\mathfrak{S}_0^\star\subset \mathfrak{S}^\star$ denote the set of all connected $k-1$-dimensional strata which are boundary components of some
$N \in \mathfrak{S}_0$. Let $M \in \mathfrak{S}_0^\star$, $N \in \mathfrak{S}_0$ with $M \subset \partial N$. If $f \in \cal{K}_{0,\mathfrak{S}}$ and $\gamma_f(0) \in M$, 
then $\Gamma_f \trans N$ at $(\is{0},0)$ and so, by Whitney regularity, $\Gamma_f$ has a non-trivial transversal intersection with $N$ at $(\is{0}, \gamma_f(0))$
and therefore $\Gamma_f^{-1}(N \cup M)$ contains a 1-dimensional Whitney regular stratified set $C$ with $(\is{0},0)\in \partial C$.
Using Whitney regularity, $C$ is a $C^1$ submanifold of $V\times \real$ with boundary
point $(\is{0},0)$.  Hence $\Gamma_f^{-1}(N \cup M)$ contains a 
non-trivial solution branch in $\Sigma(f)$, with 
$C^1$-parametrization as defined in Definition~\ref{def: branch}.  Alternatively, 
we may invoke Paw\l{}ucki's theorem~\cite{Paw1985} which implies that
$N \cup M$ is a \emph{$C^1$-submanifold of $V \times \real^k$} and so the intersection is a 1-dimensional $C^1$ submanifold 
by the transversality theorem\footnote{Thom's transversality isotopy theorem only gives $C^0$-local trivialization of $N \cup M$ since 
vector fields on $M\cup N$ are \emph{only $C^0$}.}.
More generally, by openness of transversality, we may 
choose a closed neighbourhood $D$ of $(\is{0},0)\in V \times \real$ such that $\Gamma_f|D \trans \mathfrak{S}$ and $(\Gamma_f|D)^{-1}(\Lambda)$ 
gives the branching pattern $\Sigma(f)$. In particular, if $S \in \mathfrak{S}$, then $\Gamma_f(D) \cap S \ne \emptyset$ only if 
$S \in \mathfrak{S}_0 \cup \mathfrak{S}_0^\star$.  

\begin{rems}\label{rem: path}
(1) The argument given above proves that solution branches and the finiteness of the branching pattern are generic in equivariant bifurcation theory. \\
(2) Paw\l{}ucki's theorem applies to Whitney regular stratifications of subanalytic sets---it is not true for general Whitney regular stratifications
with smooth strata~\cite{Paw1985}.  \rend
\end{rems}
\subsection{Analytic parametrization of solution branches}
The $C^1$-parametrization of branches given by equivariant transversality is precisely what is needed  
for the proof of Theorem 4.2(b)~\cite{CLM1990}, see Lemma~\ref{lem: clm} below.  In this section, we give conditions for
analytic parametrization of solution branches for polynomial and analytic families that make use of the Curve Selection Lemma (CSL)~\cite{Milnor1968}.  We only give the details
for polynomial maps (using~\cite{Milnor1968}). The results extend 
easily to real analytic families using the CSL for semianalytic sets~\cite[II, \S3, III, \S8]{Slo95} (see \cite[\S 9]{Kuo} for historical notes on the CSL and its 
significance in singularity theory).

Assume that $n \ge \delta_w(V,G)$ and let $\mathcal{V}^n$ (resp.~$\mathcal{V}_0^n$) be the subset of $\cal{V}$ (resp.~$\cal{V}_0$) consisting of families $f\in\mathcal{V}$ (resp.~$f\in\mathcal{V}_0$) such that 
$f$ is polynomial in $(\xx,\lambda)$ of degree at most $n$. Let $\mathcal{V}^\omega$ and $\mathcal{V}_0^\omega$ be the corresponding spaces of real analytic families.
Details below are only given for $\mathcal{V}_0^n$; results
and methods are the same for $\mathcal{V}^n$,  $\mathcal{V}_0^\omega$ and $\mathcal{V}^\omega$.  We have
$$ \mathcal{V}^n_0 = \{f \in \mathcal{V}_0 \dd f(\xx,\lambda) = \lambda \xx + P(\xx), \;P \in \widehat{P}^n_G(V,V)\}.$$
It follows from Theorem~\ref{thm: stable} that if $n \ge \delta(V,G)$ then $\mathcal{V}_0^n$ contains an open and dense subset
$\mathcal{S}_0^n=\mathcal{V}^n_0\cap \mathcal{S}$ of stable families (standard vector space topology on $\widehat{P}^n_G(V,V) \subset P^n_G(V,V)$. 

If $f \in \mathcal{V}_0^n$, then  $f(\xx,\lambda) = \lambda \xx + \sum_{i\in\is{k}} g_i(\xx)F_i(\xx)$,
where the coefficient functions $g_i$ are polynomial invariants, $g_1(\is{0}) = 0$,
and the values $g_2(\is{0}),\cdots,g_k(\is{0})$ are uniquely determined by $f$. As above, we define the real algebraic proper embedding
$\Gamma_f: V \times \real \arr V \times \real^k$ by
\[
\Gamma_f(\xx,\lambda) = (\xx,(\lambda+ g_1(\xx), g_2(\xx),\cdots,g_k(\xx))),\; (\xx,\lambda) \in V \times \real.
\]

Given a Whitney regular semialgebraic stratification $\mathfrak{S}$ of $\Lambda$, and $n \ge  \delta_w(V,G)$, define  the space $\cal{K}^n_{0,\mathfrak{S}} \subset \mathcal{V}^n_0$
of weakly stable real polynomial families by
\[
\cal{K}^n_{0,\mathfrak{S}} = \{f\in\mathcal{V}^n_0 \dd \Gamma_f \trans \mathfrak{S} \;\text{at } (\is{0},0)\}.
\]
Repeating the arguments given in the previous section, it follows that if $f\in \cal{K}^n_{0,\mathfrak{S}}$, then we can choose a neighbourhood
$D$ of $(\is{0},0) \in V \times \real$ such that $(\Gamma_f|D)^{-1}(\Lambda)$ is a finite union of connected 1-dimensional real algebraic subsets of $V \times \real$ 
with common boundary $(\is{0},0)$. 
Hence, by the CSL, each  non-trivial $[\gamma]=[(\xx,\lambda)] \in \Sigma(f)$ may be parametrized as a real analytic branch
\[
\xx(s) = s^p \vv_p + \sum_{j > p} s^j \vv_j,\quad \lambda(s) = a s^q,\;  s \ge 0,
\]
where $q \ge p\ge 1$, $\vv_p \in V$ is non-zero and we assume is $p$ is minimal. If the branch has isotropy $H$,
then $\vv_j \in V^H$, all $j \ge p$. If $a \ne 0$, we can always reparametrize so that $a = \pm 1$ and so, since $p$ is minimal,
the parametrization of $\xx(s)$ is then unique. Note that $a \ne 0$ iff the branch is either forward or backwards. Hence, for families in $ \mathcal{S}_0^n$,
the parametrization with minimal exponent is unique with $a \in \{\pm 1\}$.
\begin{rems}\label{rem: papa}
(1) In our setting, the \emph{regular arc-wise analytic} stratification $\mathfrak{P}$ of Parusi\'nski and Paunescu~\cite[\S 1.2]{PaPa2017} is a more
natural choice of stratification than the canonical stratification since 
it gives a Whitney regular stratification of
$\Lambda$ into semialgebraic strata which satisfy a strong trivialization property which fails in general for the canonical stratification. 
Thus $\mathfrak{P}$ gives regular arc-wise analytic trivializations
of strata pairs of dimension $k$ and $k-1$ and this implies that initial exponents of real analytic parametrizations are 
locally constant on $\cal{K}^{\delta_w}$ (this follows from Props.~1.6, 7.4 \emph{op.~cit.}).  \\
(2) If $(V,G)$ has no quadratic equivariants,  $q = 2p$ for stable families. 
However, if there are quadratic equivariants and $D^2f_0(\vv_p) = \is{0}$, all that can be claimed 
in general seems to be $q > p$ (for axial branches, obviously  
$p = 1$, $q \in \{ 1,2\}$). See also Section~\ref{sec: analparam}.\rend
\end{rems}
If $p = 1$, then $\gamma$ is a real analytic embedding. If $p > 1$, set $t = s^p$ so that
\[
\xx(t) = t\vv_p + t\left(\sum_{j \ge 1} t^{\frac{j}{p}}\vv_{p + j}\right), \; \lambda(t) = a t^{\frac{p}{q}}, t \ge 0.
\] 
to obtain a $C^{1+\frac{1}{p}}$ fractional power series parametrization of $\gamma$ which is a $C^1$-embedding. Hence 
we obtain a solution curve with $\xx'(0) = \vv_p \ne \is{0}$. Note that if $q > p$, then $\lambda'(0) = 0$ 
and so the branch is tangent to $V \times \{0\}$ at $t = 0$. As indicated in Remarks~\ref{rem: papa}(3), $p$ may not depend continuously on $f$ unless
the stratification satisfies additional conditions going beyond Whitney regularity. If we use the regular arc-wise analytic stratification 
$\mathfrak{P}$ of $\Sigma$ then $p$ is locally constant and it may be shown (using~\cite[\S 7]{PaPa2017}) that 
$\vv_p$, and so the direction of branching $\db(\gamma)$, depend continuously on $f \in  \mathcal{K}_{0,\mathfrak{S}}^n$. 
However, nothing is said about the exponent $q$.

\subsection{Proof of Theorem~\ref{thm: main}: branches of type S}
Assume that $(V,G)$ has quadratic equivariants.
We start with some preliminary results before proving a version of Theorem~\ref{thm: main} that applies to branches $\gamma(s)$  that are not tangent to $V$ at $s = 0$.
First, an elementary  lemma about the derivative of an equivariant map.
\begin{lemma}\label{lem: equiv}
If $f: V \arr V$ is equivariant and $C^1$, then $Df: V \arr L(V,V)$ is $G$-equivariant:
$ g^{-1}Df_{g\xx}g =  Df_\xx,\; g \in G,\, \xx \in V $.
\end{lemma}
\begin{proof} Differentiate $f(g\xx) = g f(\xx)$ using the chain rule to get
$Df_{g\xx} g = g Df_\xx$, for all $\xx \in V$, $g \in G$. 
\end{proof} 

Given $f \in \mathcal{V}_0$,  define $\text{\rm Tr}(f): V \times \real \arr \real$ by
\[
\text{\rm Tr}(f)(\xx,\lambda) = \text{trace}(Df_{\lambda,\xx}), \;\; (\xx,\lambda) \in V \times \real,
\]
where $Df_{\lambda,\xx}$ denotes the derivative of $f_\lambda$ at $\xx$.
\begin{lemma}
$\text{\rm Tr}(f)$ is $G$-invariant.
\end{lemma}
\proof The invariance of $\text{\rm Tr}(f)$ is immediate from Lemma~\ref{lem: equiv}. \qed

\begin{rem}
A similar result holds for the symmetric polynomials in the eigenvalues of $Df_{g\xx}$---the traces of $\wedge^r Df_{\lambda,\xx}$, $2 \le r \le \text{dim}(V)$. \rend
\end{rem}

\begin{lemma}[{\cite[Lemma 4.4]{IG1984}}]\label{lem: trace}
(Notation and assumptions as above.) Suppose $\text{dim}(V) = m$ and $Q \in P^{(2)}_G(V,V)$.
If $F_\lambda(\xx) = \lambda \xx + Q(\xx)$, then
\begin{enumerate}
\item $ \text{\em Tr}(F)(\xx,\lambda) = m \lambda, \; (\xx,\lambda) \in V \times \real$.
\item $ \text{\em Tr}(Q) \equiv 0$.
\end{enumerate}
\end{lemma}
\proof Since $DQ_\xx$ in linear in $\xx$ and there are no non-zero linear invariants, $ \text{Tr}(DF_{\lambda})$ is independent of $\xx$ and (1)
follows since $DF_{\lambda}(\is{0}) =  \lambda I_V$. Hence  $ \text{Tr}(Q) \equiv 0$, proving (2). \qed

\begin{lemma}[{Proof of Theorem~2.2(b)\cite{CLM1990}}]\label{lem: clm}
If $f \in \cal{K}(V,G)$ and $\gamma(s)=(\xx(s),\lambda(s))$ is a type S solution branch of 
$\xx' = f_\lambda(\xx)$, then $\gamma$ is unstable.
\end{lemma}
\begin{proof} Since $f\in \cal{K}(V,G)$, we may require that $\gamma$ is a $C^1$-embedding and 
$\gamma(s)=(s \vv + o(s), \pm s)$, where $\vv/\|\vv\| = \db(\gamma)$.  Set $J^2(f) = Q$.
Without loss of generality, suppose the branch is forward: $\lambda(s) = s$. We have
$f_\lambda(\xx) = \lambda \xx + Q(\xx) + O(|\lambda|\|\xx\|^2) + O(\|\xx\|^3)$ (it is not assumed that $f \in \cal{V}_0$). 
Substituting for $(\xx,\lambda)$, dividing by $s^2$ and setting $s = 0$ gives
$Q(\vv) = -\vv$.  By Euler's theorem, $DQ_{\xx(s)}(\xx(s)) = 2Q(\xx(s))$ and so, substituting for $\xx(s)$, we see
that $DQ_{s\vv}$ has eigenvalue $\mu_r(s)=-2s$ and so $D(s I_V + Q)_{s\vv}$ has the eigenvalue $-s$.  
Hence, by Lemma~\ref{lem: trace}(1), $D(s I_V + Q)_{s\vv}$ has an eigenvalue with strictly positive real part $\alpha(s) = a s$, $a > 0$. 
The terms we have omitted from $Df_{s,\xx(s)}$ are all $o(s)$ and so, by the continuous dependence of eigenvalues on the coefficients
of the characteristic equation, there exists $\vareps > 0$ such that for $s\in (0,\vareps]$, there is an eigenvalue with 
strictly positive real part. Hence the branch is unstable. \end{proof}

\begin{rem}\label{rem: clmnb}
The proof is similar to that in~\cite{CLM1990} except no use is made of the CSL which does not
apply here unless it is assumed that $f$ is polynomial (or real analytic) in $(\xx,\lambda)$~\cite{Slo95} and use is made of Theorem~\ref{thm: stable} (to 
extend the the result to families in $ \cal{K}(V,G) \cap \cal{S}(V,G) =  \cal{S}(V,G)$).
Modulo the use of equivariant transversality in obtaining solution branches, the proof is
simple but gives no quantitative information
on the interval $(0,\vareps]$ for which there are eigenvalues of opposite sign. 
As is shown below, it is possible to obtain estimates on eigenvalues and $\vareps$
using the CSL if we assume families are analytic or polynomial.
\rend
\end{rem}

Suppose that $f \in \mathcal{S}_{0}^\omega$ (similar results hold for families in $\mathcal{S}^\omega$, including polynomial families). 
Let $\gamma = (\xx,\lambda):[0,\delta]\arr V \times \real$
be a non-trivial branch of solutions for $f = 0$. By the CSL, we may write 
\begin{eqnarray}\label{eq: csl1}
\xx(t) & = & \sum_{j=p}^\infty \vv_j t^j \\
\label{eq: csl2}
\lambda(t) & = & \sgn([\gamma]) t^{q},
\end{eqnarray}
where $\vv_j \in V$, $j \ge p$, $\vv_p \ne \is{0}$, $p,q > 0$ and $p$ is minimal. 
The power series for $\xx$ is unique granted the minimality of $p$ and the expression for $\lambda(t)$. 
In what follows, we often assume the branch is forward, so that $\lambda(t)=  t^{q}$
(the arguments we give apply equally to the case $-t^q$). 

\begin{prop}\label{prop: part1}
(Notation and assumptions as above.)
Let $f\in \mathcal{S}_0^\omega$ and $\gamma = (\xx,\lambda)$ be a solution branch with unique analytic parmetrization (\ref{eq: csl1},\ref{eq: csl2}).
Suppose that 
\begin{align}\label{eq: key1}
J^2(f) = Q \in P^{(2)}_G(V,V),& \; \text{and } Q(\vv_p)  \ne 0.
\end{align}
($[\gamma]\in\Sigma^\star(f)$ is of type \emph{S}).
Then $q = p$ and $\gamma$ is a branch of hyperbolic saddles: $\text{\em index}(\gamma) \in [1,\text{\em dim}(V) - 1]$.
If  \Refb{eq: key1} holds for all $[\gamma] \in \Sigma^\star(f)$, then every non-trivial branch $\gamma$ of solutions
of $f$ is a branch of hyperbolic saddles with $\text{\em index}(\gamma) \in [1,\text{\em dim}(V) - 1]$.\\
In particular, if $\mu$ is an eigenvalue of $DQ_{\vv_p}$, then $Df_{\lambda(t),\xx(t)}$ will have an eigenvalue
$\tilde \mu(t)$ where
\begin{enumerate}
\item If $\mu \ne 2$, or $\mu = 2$ and the associated generalized eigenspace is not $\real\vv_p$, then
\[
\tilde \mu(t) = t^p \big[\text{\em sgn}([\gamma])+\mu + O(t^{\frac{1}{\ell}})\big],
\]
If $\mu\ne 2$ has multiplicity $m$, then $Df_{\lambda(t),\xx(t)}$ will have $m$ eigenvalues of this type (counting multiplicities).
\item If $\mu = 2$, with associated eigenvector $\vv_p$, then 
\[
\tilde \mu(t) = t^p \big[-\text{\em sgn}([\gamma]) + O(t^{\frac{1}{\ell}})\big].
\]
\end{enumerate}
In either case, we may take $\ell = 1$ if $f$ is a family of gradient vector fields or if $\mu$ is a simple eigenvalue.\\
The result continues to hold if $f\in \cal{S}^\omega$ or $\cal{S}^n$, $n \ge \delta$.

\end{prop}

\begin{rem}
The main step in the proof is to show that
that $Df_{\lambda(t),\xx(t)}$ has an eigenvalue $\mu(t) = -\lambda(t)(1+  O(t^{\frac{1}{\ell}}))$, and associated  
eigenvector $\ee(t) = \vv_p + O(t^{\frac{1}{\ell}})$,  for some $\ell \in \pint$.  
The estimate holds with $\ell = 1$  if the eigenvalue $-1$ of $I_V + DQ_{\vv_p}$ is simple (using an implicit function theorem argument, this only
requires $f$ $C^\infty$) or if $f$ is an analytic family of gradient vector fields (and so $Df_\lambda$ is symmetric). This is a 
well-known result in perturbation theory~\cite{Re1968}, \cite[Chap.~2]{Ka1976}.
In either case, 
eigenvalues and eigenvectors depend analytically on $t$.

In general, we have to allow
for the eigenvalue $-1$ of $I_V + DQ_{\vv_p}$ to be multiple and/or for $Df_\lambda$ to be asymmetric. Here we 
rely on Puiseaux's theorem to obtain fractional power series expansions for the eigenvalues (and eigenvectors) of $Df_{\lambda(t),(\xx(t)}$,
viewed as a perturbation of $t^p(I_V + DQ_{\vv_p})$. That is, the characteristic equation of $t^{-p}Df_{\lambda(t),(\xx(t)}$ is 
a polynomial with coefficients depending analytically on $t$ and Puiseaux's theorem is used to parametrize the 
eigenvalues (\cite[Chap.~5]{Kn1996} or \cite{Walker}).
We can assume analyticity of coefficients since $f \in \mathcal{S}^\omega$ and the curve selection lemma gives an analytic
parametrization of the solution branch. If we only assume $f \in \mathcal{S}$, then the terms $O(t^{\frac{1}{\ell}})$ in the
estimates are replaced by $o(t)$ (as in the proof of Lemma~\ref{lem: clm}---sometimes this can be improved by approximation 
of $f$ by a Taylor polynomial in $\cal{S}_0^n$). 
We allow for more than one solution branch with the same direction of branching $\vv_p/\|\vv_p\|$: analyticity 
implies these branches will be distinct branches of
equilibria for sufficiently small non-zero values of the parameter. Similarly, if $2$ is not a simple eigenvalue of $DQ_{\vv_p}$, the eigenvector $\vv_p$ might split into
several eigenvectors when we add in the higher order terms. However, these eigenvectors will be close to $\vv_p$ and the associated sum of (generalized) eigenspaces will be close 
to the original generalized eigenspace of $I_V + DQ_{\vv_p}$. The exponent $1/\ell$ may be small.
\rend
\end{rem}

\pfof{Prop.~\ref{prop: part1}.} 
Substituting $\xx(t) = \sum_{i=p}^\infty \vv_i t^i$, $\lambda(t) = t^q$ in $f(\xx,\lambda) = 0$ and equating lowest powers of $t$
we find that $q = p$ and $Q(\vv_p) = -\vv_p$.  By Euler's theorem, $DQ_{\xx(t)}(\xx(t)) = 2Q(\xx(t))$. Substituting for $\xx(t)$ we find that
$DQ_{t^p\vv_p}$ has the eigenvalue $\tilde{\mu}_r(t) = -2t^p = -2\lambda(t)$ and associated eigenvector
$\tilde{\ee}_r = \vv_p$. If we write $f_\lambda(\xx) = \lambda\xx + Q(\xx) + H(\xx)$, where $H(\xx) = O(\|\xx\|^3)$, then
$Df_{\lambda(t),\xx(t)} = \lambda I_V + DQ_{\xx(t)} + t^{2p} A(t)$, where $A(t) $ is an analytic family of linear maps. 
Since $DQ_{\xx(t)} = DQ_{t^p\vv_p} + O(t^{p+1})$, it follows by perturbation theory (see the discussion above), that
$Df_{\lambda(t),\xx(t)}$ has eigenvalue $\mu_r(t) = -t^p + O(t^{p + \frac{1}{\ell}}) = -\lambda(t)(1 + O(t^{\frac{1}{\ell}}))$ and associated eigenvector
$\ee_r(t) =  \vv_p+ O(t^{\frac{1}{\ell}})$ (we may take $\ell = 1$ if  $\tilde{\mu}_r(t)$ is simple or $Q$ is gradient).
Hence for sufficiently small $t > 0$, $Df_{\lambda(t),\xx(t)}$ has an eigenvalue with strictly negative real part.
Setting $F(\xx,\lambda) = \lambda \xx + Q(\xx)$, it follows from  Lemma~\ref{lem: trace} and the 
standard remainder estimate for Taylor's theorem. that
\begin{eqnarray*}
\text{Tr}(f)(\xx(t),\lambda(t))& =& \text{Tr}(F)(\xx(t),\lambda(t)) + O(t^{2p})\\
& = & m \lambda(t) + O(t^{2p}).
\end{eqnarray*}
Since we have shown that $Df_{\lambda(t),\xx(t)}$ has an eigenvalue $-\lambda(t)(1 + O(t^{\frac{1}{\ell}}))$,
it follows that for sufficiently small $t > 0$, $Df_{\lambda(t),\xx(t)}$ has an eigenvalue with
strictly positive real part. The estimates on the remaining eigenvalues of $Df_{\lambda,\xx(t)}$ follow using perturbation theory.

Finally, if $f \in \cal{V}^\omega$, then $f$ is stable if
$J^\delta(f) \in \cal{S}^\delta $, $\delta = \delta(V,G)$, and then $\Sigma^\star(f)$ is isomorphic to $\Sigma^\star(J^\delta(f))$. \qed

\begin{cor}\label{cor: part1}
Theorem~\ref{thm: main} is true if there is an open dense set $\Xi$ of $P^{(2)}_G(V,V)$ such that 
$\is{Z}(Q) = \{0\}$ if $Q \in \Xi$.
\end{cor}
\proof We may assume that if $f \in \cal{S}$, then $J^2(f) \in \Xi$
(this is an open and dense condition). By Proposition~\ref{prop: part1} (or Lemma~\ref{lem: clm}), if $f \in \cal{S}^\delta$, then all solution branches
are branches of hyperbolic saddles with $\text{index}(\gamma) \in [1,\text{dim}(V) - 1]$. 
If $f \in \cal{V}$, then $f$ is stable if 
$J^\delta(f) \in \cal{S}_0^\delta $, and then $\Sigma^\star(f)$ is isomorphic to $\Sigma^\star(J^\delta(f))$. \qed

\begin{exam}
The standard representation $\mathfrak{s}_k$ satisfies the conditions of Corollary~\ref{cor: part1} if $k$ is odd. Of course, the result
is straightforward to prove directly (see~\cite[\S 16]{FR1992}). \examend
\end{exam}
\subsection{Completion of the proof of Theorem~\ref{thm: main}}
It remains to consider branches of type \emph{C} and $\propG$. 
The next lemma is the final step needed for the proof of Theorem~\ref{thm: main}.

\begin{lemma}\label{prop: typeC}
Suppose that \emph{$\propG$} holds. Let $f\in \mathcal{S}$ and set $f^\delta(\xx,\lambda) = \lambda\xx + J^\delta(f) \in \mathcal{S}_0^\delta$.
Let $\gamma = (\xx,\lambda)$ be a solution branch of $f^\delta$ with 
unique analytic parametrization (\ref{eq: csl1},\ref{eq: csl2}).
Assume that $[\gamma]\in\Sigma^\star(f^\delta)$ is of type \emph{C}:
\begin{enumerate}
\item $J^2(f) = Q\in P^{(2)}_G(V,V)$.
\item $Q \ne 0, \; Q(\vv_p)  = 0$.
\end{enumerate}
Then $q > p$ and $\gamma$ is a branch of hyperbolic saddles.  
\end{lemma}
\proof Substituting the series for  $\gamma(t)$ in $f^\delta_\lambda(\xx)$, it follows that $q > p$ since $Q(\vv_p) = 0$.   
By $\propG$  and Lemma~\ref{lem: trace},
$DQ_{\vv_p}$ has at least one eigenvalue with strictly positive real part and one eigenvalue with strictly negative real part.
We have 
\[
Df^\delta_{\lambda(t),\xx(t)} = \lambda I_V + DQ_{\xx(t)} + O(t^{p+1}) = t^p(DQ_{\vv_p} + O(t)),
\]
and so it follows, as in the proof of Lemma~\ref{lem: clm}, that for sufficiently small $t > 0$, $Df^\delta_{\lambda(t),\xx(t)}$ has eigenvalues with strictly positive and negative
real parts. \qed

\begin{rems}
(1) Little use is made of the analytic parametrization. In particular, nothing is said about $q$ except that $q > p$. \\
(2) Although the proof of Proposition~\ref{prop: typeC} is easier than the proof of Proposition~\ref{prop: part1}, it is
unsatisfying as we do not address the ``radial'' eigenvalue associated to the direction $\vv_p$. 
If we \emph{assume} $Q | V^{G_{\vv_p}} \equiv 0$, then it is straightforward to show that $q = 2p$ and the radial eigenvalue is $-2\lambda(t) + O(t^{2p+1})$
(this is related to~\cite[Thm.~4.2(a)]{CLM1990}).  However, if say $Q | V^{G_{\vv_p}} \not\equiv 0$, $q = p+1$, and $p > 1$, then there is the possibility that 
$Q(\vv_{p+1})$ is a non-zero multiple of $\vv_p$ and in this case the radial eigenvalue will not be determined by the cubic terms though it is still dominated by the non-radial
eigenvalues of $DQ_{\vv_p}$. 
Of course, all this is easy for axial branches (for example, the branches with isotropy conjugate to $S_\ell\times S_\ell$ for $\mathfrak{s}_{2\ell}$).  \rend
\end{rems}
\pfof{Theorem~\ref{thm: main}.} If  $f\in \mathcal{S}$, then $f^\delta$ and $f$ have isomorphic signed indexed branching patterns by Theorem~\ref{thm: stable}. 
By Proposition~\ref{prop: part1} (or Remark~\ref{rem: clmnb}) and Lemma~\ref{prop: typeC}, all non-trivial branches in $\Sigma^\star(f^\delta)$ are branches of hyperbolic saddles. \qed
\subsection{Notes on the analytic parametrization for type \emph{C} branches}\label{sec: analparam}

Suppose $f \in \mathcal{S}_0^\omega$ and $\gamma = (\xx,\lambda):[0,\delta]\arr V \times \real$
is a non-trivial branch of solutions of $f = 0$. Write
\begin{equation}\label{eq: series}
\xx(t)  =  \sum_{i=p}^\infty \vv_i t^i,\; \lambda(t)  =  \pm t^{q} 
\end{equation}
where $\vv_i \in V$, $i \ge p$, $\vv_p \ne \is{0}$, $p,q > 0$ and $p$ is minimal. Assume the branch is forward, so that $\lambda(t)=  t^{q}$.
The power series for $\xx$ is unique granted the minimality of $p$ and the expression for $\lambda(t)$. Suppose that $Q(\vv_p) = 0$ so that
$\gamma$ is a type \emph{C} branch and $q > p$.  Since $Q$ is a homogeneous quadratic polynomial, there is a unique symmetric bilinear form
$A: V^2 \arr V$  satisfying $Q(\xx) = A(\xx,\xx)$, $\xx \in V$, defined by  $A(\xx,\yy) = \frac{1}{2}[Q(\xx+\yy) -Q(\xx)-Q(\yy)]$, $\xx,\yy \in V$. 

Substituting $\xx(t)$, $\lambda(t)$ in $\lambda \xx + Q(\xx) = F(\xx,\lambda)$ we find that if $p < q < 2p$ and $p > 1$,  then
\begin{equation}\label{eq: pq}
F(\xx(t),\lambda(t))=t^{p+q} \vv_p + \sum^{p-1}_{\ell=1} t^{2p+\ell} K_\ell(\vv_p,\cdots, \vv_{p+\ell}) + O(t^{3p}),
\end{equation}
where $K_\ell(\vv_p,\cdots, \vv_{\ell-p}) = \sum_{i+j=\ell}A(\vv_i,\vv_j)$ (sum over $i,j \ge p$), and
\begin{eqnarray}\label{eq: q1a}
K_\ell(\vv_p,\cdots, \vv_{p+\ell})&=& 0, \; \ell < q-p,\\
\label{eq: q2a}
K_{q-p}(\vv_p,\cdots, \vv_q)& =& -\vv_p,  \\
\label{eq: q3a}
K_{\ell}(\vv_p,\cdots, \vv_{p+\ell}) & = & 0, \; q-p < \ell < p .
\end{eqnarray}
We give one computation to illustrate some of the issues. Suppose $1 < p < q = p+1 < 2p$,  $\text{dim}(V^{G_{\vv_p}}) > 1$ (the branch is not axial), and the branch is forward.
Set $G(t) = t^{-p} DF_{\lambda(t),\xx(t)}$. Substituting the series for $\xx(t)$ given by \Refb{eq: series} in $F$ and equating lowest order coefficients we find that
\begin{equation}\label{eq: abc}
2A(\vv_{p+1},\vv_p) + \vv_p = 0
\end{equation}
Since $\vv_p \ne \is{0}$, $A(\vv_{p+1},\vv_p) \in V^{G_{\vv_p}}$ is non-zero. Computing $G(t)$, we find
\[
G(t) = tI_V + DQ_{\vv_p} + t DQ_{\vv_{p+1}} + O(t^2).
\]
Hence, by \Refb{eq: abc},
\begin{eqnarray*}
G(t)(\vv_p) & = & t\vv_p + 2t A(\vv_{p+1},\vv_p) + O(t^2) = O(t^2)\\
G(t)(\vv_{p+1}) & = & -\vv_p + t(\vv_{p+1} + Q(\vv_{p+1})) + O(t^2).
\end{eqnarray*}
Hence $\lambda = 0$ is a non-simple eigenvalue of $G(0)$ and
without further information on $Q$ it is possible that $G(0)|V^{G_{\vv_p}}$ has pure imaginary eigenvalues.
If $p=1$, a cubic term $C(\xx)$ will contribute an $O(t)$-term to $G(t)(\vv_{p+1})$ and $G(t)(\vv_p)$ now has a term $t C(\vv_p)$. 
Even if $\propG$ holds, it seems difficult to determine an analytic form for the radial direction as the low order contribution given by $DQ_{\vv_p}(\vv_{p+1})$ may
dominate those coming from higher order terms. Many questions remain.

\section{Minimal models of forced symmetry breaking of generic bifurcation on $\mathfrak{s}_k$}\label{sec: MMFSB}
Suppose given a generic steady-state bifurcation defined on an absolutely irreducible representation $(V,G)$. For example,
the pitchfork bifurcation on $(\real,\intg_2)$ or 
the $S_k$-equivariant bifurcation on the standard irreducible representation $\mathfrak{s}_k$. In order to understand symmetry breaking
perturbations, it is natural to ask if there is a way to embed
the bifurcation in a non-equivariant multiparameter family of vector fields which typically exhibit only generic bifurcation (that is, saddle-node
bifurcations).  More formally, given a smooth 1-parameter family $f_\lambda$ of $G$-equivariant vector fields
defined on an open neighbourhood $U$ of $(0,0)\in V \times \real$, an \emph{unfolding} of the family consists of a 
smooth map $K:(U \times \real) \times \real^m \arr V; ((\xx,\lambda),\boldsymbol{\eta})\mapsto K_{\boldsymbol{\eta}}(\xx,\lambda)$
such that $K_{\is{0}}(\xx,\lambda) = f(\xx,\lambda)$, $(\xx,\lambda) \in U$. Roughly speaking, the unfolding is \emph{universal} if for every 
sufficiently small smooth perturbation $\hat{f}$ of $f$, there exists
$\boldsymbol{\eta} \in \real^m$ (close to $\is{0}$) such that $K_{\boldsymbol{\eta}}$ is equivalent in some sense to 
$\hat{f}$ (see~\cite[Chap.~II, p.~51]{GSS1988} for `strong equivalence' and more details on the singularity approach to unfoldings which 
we do not follow here). The aim is to be able to describe  
all small perturbations of $f$ in terms of the extended family $K$. 
Unfortunately, it is not 
realistic to look for universal unfoldings of generic $S_k$-equivariant bifurcation if $k \gg  3$ as the number of parameters $m$ required 
is likely to grow rapidly with $k$ (the case $k$ even is especially awkward).

Our approach will be to show that that there are ``minimal models'' for symmetry-breaking perturbations of generic bifurcation on 
$\mathfrak{s}_k$ and to emphasize deformation to a minimal model rather than an explicit construction of a universal unfolding. Although
we restrict here to $\mathfrak{s}_k$, the methods we describe can likely be extended to $\mathfrak{s}_k \boxtimes \mathfrak{s}_d$, $k,d \ge 3$,
and possibly to families of equivariant gradient vector fields on general absolutely irreducible representations admitting quadratic equivariants.

The qualifier \emph{minimal} is intended to suggest a minimal
level of complexity in the dynamics of the symmetry breaking perturbation given by a  minimal model. 
Our motivation lies in describing mechanisms for creating or destroying local minima in gradient systems 
\emph{without} creating new local minima in the process. The question arises from problems in
non-convex optimization in neural nets and the occurrence of spurious minima. The
spurious minima do not come from bifurcation of the global minima (or spontaneous symmetry breaking) but are created locally through changes in the geometry of the optimization 
landscape. 

\subsection{Generic bifurcation for  $\mathfrak{s}_k$}\label{sec: gensk}
Generic steady-state equivariant bifurcation of the trivial solution for families of $S_k$-equivariant vector fields on $\mathfrak{s}_{k}$
is well understood~\cite[\S 16]{FR1992}, \cite{FR1989}. If $k$ is odd, equivariant bifurcation is $2$-determined,
and for stable families all branches are of type \emph{S}. If $k$ is even, equivariant bifurcation is $3$-determined and
for stable families all branches
are of type \emph{S} except branches with isotropy conjugate to $S_\ell \times S_\ell$; these are of type \emph{C}.
All branches of solutions arising from generic bifurcation of the trivial solution are axial~\cite{FR1989}  and
the set $\{\Sigma^\star(f) \dd f \in \mathcal{S}(H_{k-1},S_k)\}$ is known~\cite{FR1992} (see below).

We recall definitions and results from~\cite{FR1992,Field2007} needed later.

The space $P_{S_k}^{(2)}(H_{k-1},H_{k-1})$ is 1-dimensional with basis given by the gradient
of the homogeneous cubic $S_k$-invariant $C(\xx)= \frac{1}{3}\sum_{i\in \is{k}}x_i^3|H_{k-1}$.
Set $Q = \grad{C}$.  We have
\[
Q(\xx) = (x_1^2 - \frac{1}{k}\sum_{i\in\is{k}}x_i^2,\cdots ,x_k^2 - \frac{1}{k}\sum_{i\in\is{k}}x_i^2),\; \xx=(x_1,\cdots,x_k) \in H_{k-1}.
\]
The \emph{phase vector field} $\cal{P}_Q$ of $Q$ is defined on the unit sphere $S^{k-2} \subset H_{k-1}$ by
\[
\cal{P}_Q(\uu) = Q(\uu) - \langle Q(\uu),\uu\rangle \uu,\; \uu \in S^{k-2},
\]
and $\cal{P}_Q = \grad{C|S^{k-2}}$.
Denote the zero set of $\cal{P}_Q$ by $\mathbf{Z}(\cal{P}_Q)$ and note that
$\uu \in \mathbf{Z}(\cal{P}_Q)$ iff $-\uu\in \mathbf{Z}(\cal{P}_Q)$ iff $Q(\uu) \in \real \uu$.

Given $1 \le p < k$, set $q = k-p$ and define $\boldsymbol{\varepsilon}_p\in S^{k-2}$
by\footnote{This is slightly different from~\cite[\S 16]{FR1992} where
$\boldsymbol{\varepsilon}_p$ defined to be $(\frac{1}{p}^p,-\frac{1}{q}^q)$.}
\begin{equation}\label{eq: vareps}
\boldsymbol{\varepsilon}_p = \frac{1}{\sqrt{pqk}}(q,\cdots,q, -p,\cdots,-p)\defoo \frac{1}{\sqrt{pqk}}(q^p,-p^q),
\end{equation}
where $q^p$ means $q$ is repeated $p$ times.  Note that
$\boldsymbol{\varepsilon}_{q}= -\sigma \boldsymbol{\varepsilon}_p$, where $\sigma\in S_k$ is given by $\sigma(i) = q+i,\mod k$, and
$-\boldsymbol{\varepsilon}_{p}\in S_k \boldsymbol{\varepsilon}_p$ iff $k$ is even and $p = q = k/2$.

Write $k = 2\ell+1$ ($k$ odd) or $k = 2\ell$ ($k$ even). For $p \in \boldsymbol{\ell}$, set $L_p = \real \boldsymbol{\varepsilon}_p$. The line $L_p$ is an \emph{axis of symmetry} for $\mathfrak{s}_k$
and the isotropy of non-zero points on $L_p$ is $S_p \times S_q$. In every case, except when $k$ is even and $p = \ell$,
$S_p \times S_q$ is a \emph{maximal} proper subgroup of $S_k$.  The set of axes of symmetry with isotropy conjugate to
$S_p \times S_q$ is $A_p = \{ g L_p \dd g \in S_p \times S_q\}$ and $\cup_{p \le \ell} A_p$ is the complete set of $2^{k-1}-1$ axes of symmetry
of $\mathfrak{s}_k$.

Simple computations~\cite[\S 16]{FR1992} verify $\boldsymbol{\varepsilon}_p \in \mathbf{Z}(\cal{P}_Q)$, $p \in \is{k-1}$, and 
\begin{eqnarray} \label{eq: pvf1}
Q(\boldsymbol{\varepsilon}_p)&=&\sqrt{\frac{1}{pqk}}(q-p) \boldsymbol{\varepsilon}_p, \; p \in \is{k-1} \\
\label{eq: pvf2}
\bigcup_{p=1}^{k-1} S_k \boldsymbol{\varepsilon}_p&=&\mathbf{Z}(\cal{P}_Q).
\end{eqnarray}
Moreover, $\mathbf{Z}(\cal{P}_Q)$ consists of hyperbolic zeros~\cite[\S 4]{FR1992a}
and for $p \in \is{k-1}$
\begin{eqnarray}\label{eq: Pvf1}
\quad\text{index}(\mathcal{P}_Q; \boldsymbol{\varepsilon}_p)& =& k-p-1,\\
\label{eq: Pvf2}
\quad\text{index}(\mathcal{P}_{\pm Q}; \mp \boldsymbol{\varepsilon}_p)  & =& p-1,\\
\label{eq: Pvf3}
\quad\text{index}(\mathcal{P}_Q; \boldsymbol{\varepsilon}_p) + \text{index}(\mathcal{P}_Q; -\boldsymbol{\varepsilon}_p)&= & k -2=\text{dim}(S^{k-2}).
\end{eqnarray}
The statements for $p =1$ (resp.~$k-1$) follow since points in $S_k \boldsymbol{\varepsilon}_1$ (resp.~$S_k \boldsymbol{\varepsilon}_{k-1}$) give the
absolute maximum (resp.~minimum) value of $C: S^{k-2} \arr \real$. The remaining indices can easily be found using an inductive
argument or just computed directly. We use the results on $\mathcal{P}_Q$ to give a complete description of the signed indexed branching patterns of
stable families.

Replacing $\xx$ by $-\xx/a$, $a \ne 0$, $\xx' = \lambda \xx + a Q(\xx)$, transforms to
\begin{equation}\label{eq: stform}
\xx' = \lambda \xx -  Q(\xx)
\end{equation}
Analysis of~\Refb{eq: stform} gives all the hyperbolic branches except for those with isotropy conjugate to $S_\ell \times S_\ell$, when
$k = 2\ell$.

Representative solution branches $(\xx_p^\pm,\lambda):[0,\infty)\arr H_{k-1}$ of~\Refb{eq: stform} are given for $1 \le p < k/2$ by
\begin{enumerate}
\item[(B)] $\xx_p^-(s) = -s\frac{\sqrt{pqk}}{q-p}\boldsymbol{\varepsilon}_p,\, \lambda(s) = -s$, $s \in [0,\infty)$, is a
backward branch of hyperbolic saddles of index $k-p-1$.
\item[(F)] $\xx_p^+(s) = s\frac{\sqrt{pqk}}{q-p}\boldsymbol{\varepsilon}_p,\, \lambda(s) = s$, $s \in [0,\infty)$, is a forward branch of hyperbolic saddles of index $p$.
\end{enumerate}
It follows from (B,F) that for $1 \le p < k/2$, we have
\begin{equation}\label{eq: reverse}
\text{index}([\xx_p^-]) + \text{index}([\xx_p^+]) = k-1.
\end{equation}

Excluding the case $k = 2\ell$, $p = \ell$, the set of forward and backward solution branches of~\Refb{eq: stform}
is obtained by taking the $S_k$-orbits of each of the representative solution branches. Observe that the radial eigenvalue along the branch is
always $-\lambda$---since~\Refb{eq: stform} is quadratic---and the transverse eigenvalues are given by the eigenvalues of
$\mathcal{P}_{\mp Q}$ at $\boldsymbol{\varepsilon}_p$, multiplied by $R = \|\xx(s)\|\sim s$.
If we introduce higher order terms in~\Refb{eq: stform}, then the solutions
and eigenvalues are perturbed by terms of order
$O(s^2)$ and the signed indexed branching patterns are unchanged.

\begin{lemma}\label{lem: boxest}
If $\rho > 0$ and $\delta \ge \big(\frac{k}{2}\sqrt{k+1}\big) \rho$,
then every forward solution branch
$\gamma(x) = (\xx(s),\lambda(s))$ of~\Refb{eq: stform} satisfies
\[
\gamma([0,\rho]) \subset D_\delta(\is{0}) \times [0,\rho] \subset H_{k-1} \times \real
\]
provided that we exclude branches along axes in $S_kL_{k/2}$ if $k$ is even.\\
A similar statement holds for backward solution branches. 
\end{lemma}
\proof The expressions for solution branches $(\xx(s),\lambda(s))$ given by (B,F) above imply
that the slope of the line $(|s|,\|\xx(s)\|)\subset \real^2$ is less than
$\frac{k}{2}\sqrt{k+1}$---this holds for $k$ odd or even, provided branches along axes in $S_kL_\ell$,
are excluded when $k = 2\ell$. \qed

It remains to look at solution branches when $k = 2\ell$ and $p = q = \ell$. Consider the cubic system
\begin{equation}\label{eq: stform2}
\xx' = \lambda \xx -  Q(\xx) + S(\xx),
\end{equation}
where $S \in P^{(3)}_G(H_{k-1},H_{k-1})$. For $k\ge 4$, the space $P^{(3)}_G(H_{k-1},H_{k-1})$ has basis
$R(\xx) = \|\xx\|^2\xx$ and $\grad{T}$, where $T(\xx) = \frac{1}{4}\sum_{i\in\is{k}}x_i^4 | H_{k-1}$.
Since $S: L_\ell\arr L_\ell$, $S(\boldsymbol{\varepsilon}_\ell) = c \boldsymbol{\varepsilon}_\ell$ and the generic case is when
$c \ne 0$. If $c < 0$ (resp.~$c > 0$) we have a supercritical (resp.~subcritical) pitchfork bifurcation along $L_\ell$.
In what follows we assume $c < 0$, the analysis for $c> 0$ is similar.

Computing, we find the branches along $L_\ell$ are given by $(\pm\xx_\ell(s),\lambda(s))$, where
\[
\xx_\ell(s) = \frac{1}{\sqrt{-c}}s\boldsymbol{\vareps}_\ell,\;  \lambda(s) = s^2,\; s\in [0,\infty)
\]
The radial eigenvalue $\mu_r(s)$ at $\pm\xx_\ell(s)$ is therefore $-2\lambda(s)=-2s^2$ and so $\mu_r(s) = O(s^2) = O(R^2)$, where
$R = \|\xx(s)\|$.  On the other hand, the eigenvalues at $\xx_\ell(s)$ in directions transverse to $L_\ell$ are
given by the eigenvalues of the Hessian of $\mathcal{P}_{-Q}$ at $\boldsymbol{\varepsilon}_\ell$ scaled by $R$.
Hence these eigenvalues are $O(R)$ and dominate any transverse eigenvalues coming from the cubic term $S$. Since
$\text{index}(\mathcal{P}_{\pm Q};\boldsymbol{\varepsilon}_\ell)$ is $\ell-1$, it follows that
$\text{index}([\pm\xx_\ell]) = \ell$. For $k \ge 4$, $\ell \in [1,k-2]$, and so
$(\pm\xx_\ell(s),\lambda(s))$ is a branch of hyperbolic saddles.
\begin{rems}\label{rems: coll}
(1) If $k=2\ell$, then the maximal index $\ell$ of the forward branches occurs for branches of isotropy type
$(S_\ell \times S_\ell)$ when $c < 0$ (supercritical branching in $L_\ell$). In particular, for $k \ge 4$, when the trivial solution loses stability,
not only are there no branches of sinks but the maximal index of
the new forward solution branches is  $\ell  = k/2$. The minimal index of the backward branches is also $\ell$.
Continuing to assume $c < 0$, generic equivariant bifurcation on $\mathfrak{s}_k$ results from the simultaneous collision of
$2^{k-1}-1 -\binom{2\ell-1}{\ell}$ branches of saddles of relatively high index $\ge \ell$, followed by the emergence of
$2^{k-1}-1 +\binom{2\ell-1}{\ell}$ branches of saddles of relatively low index $\le \ell$. 
Similar results hold for $k$ odd: there are now $2^{k-1}-1$ forward (resp.~backward) non-trivial branches with minimal (resp.~maximal) index $[k/2]$.
Summarizing, for all $k \ge 3$, generic equivariant bifurcation on $\mathfrak{s}_k$, changing the
trivial solution from a sink to a source, results from a collision of non-trivial saddle branches of relatively high index $\ge [k/2]$, followed by
the emergence of non-trivial saddle branches of relatively low index $\le [k/2]$.  \\
(2) Lemma~\ref{lem: boxest} fails for branches of isotropy type
$(S_\ell \times S_\ell)$ unless $\rho > 1/\sqrt{|c|}$.  
\rend
\end{rems}

\subsection{Minimal models for $\mathfrak{s}_k$.}

Suppose that $k=2\ell +1$, $\ell \in \pint$, and set $f(\xx,\lambda) = \lambda \xx - Q(\xx)$, where
$Q = \grad{C|H_{k-1}}$, $C = \frac{1}{3}\sum_{i \in \is{k}}x_i^3$. 

\begin{figure}[h]
\centering
\includegraphics[width=\textwidth]{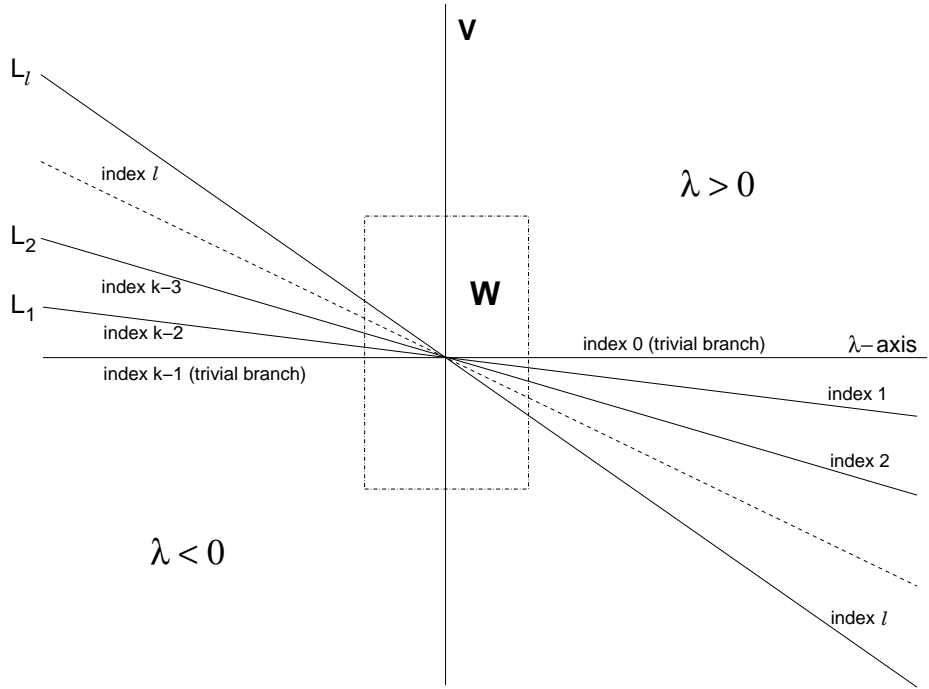}
\caption{Indices for branches, $k$ odd}
\label{fig: brandata}
\end{figure}

Let $\rho > 0$. Choose $\delta_i = K_i\rho$, $i \in \is{2}$, where $K_1 > 0$, $K_2 = K_2(k) > 0$, and
let $W$  be the compact neighbourhood of $(\is{0},0)\in (L_1 \oplus L_1^\perp) \times \real=H_{k-1}\times \real$ defined by 
\begin{equation}\label{eq: Wnbd}
W=W_\rho = ([-\delta_1,\delta_1] \times \overline{D}_{\delta_2}(\is{0})) \times [-\rho,\rho]
\end{equation}
Outside of $W$, $f$ has only hyperbolic critical points. Indeed, 
the only non-hyperbolic zero of $f$ on $V \times \real$ occurs when $\lambda = 0$ and $\xx = \is{0}$. 
Our interest is in choosing $K_i$ so that we can perturb the family $f$
so as to obtain a smooth, but not $S_k$-equivariant, family $\hat f$, equal to $f$ on $V \times \real \smallsetminus W$, such that
\begin{enumerate}
\item $\hat f$ is stable under all sufficiently small $C^2$ perturbations.
\item $\hat f$ has the maximum number of \emph{crossing curves} $\hat \gamma = (\xx,\lambda): \real \arr V \times \real$, where
$\lambda(t)$ takes all values in $\real$. For every
crossing curve, it is required that $\hat \gamma(t)$ will be a hyperbolic zero of $\hat f_{\lambda(t)}$, all $t \in \real$. 
\item $\hat f$ has the minimum number of saddle node bifurcations (necessarily in  $W$). All other zeros of $\hat{f}$ are hyperbolic.
\item As $\rho \arr 0$, $K_1,K_2 \arr 0$ (convergence of $K_2$ is not uniform in $k$).
\end{enumerate}
In section~\ref{sec: gensk}, we gave the index and branching data for solution curves of $f$ (statements (B,C)). 
We display these in Figure~\ref{fig: brandata}. 

Taking account of the indices of branches, it is clear that crossing curves must be of index $\ell$.  All
solution branches with index not equal to $\ell$ must have a saddle node bifurcation. Since the number of
branches along $S_k L_\ell$ is $\binom{k}{\ell}$, the number of crossing curves is at most $\binom{k}{\ell}$.
If there are $m$ crossing curves then the number of saddle-node bifurcations is at least $(2^{k} -2 m)/2$,
since there are $2^k$ solution branches of $f$. If $\lambda < 0$, branches differing by $1$ in index can join through a saddle
node bifurcation in the region $W$; similarly for $\lambda > 0$. A straightforward count verifies that
$m \le \binom{2\ell}{\ell}$. Hence the number of crossing curves is at most $\binom{2\ell}{\ell}$ and the number of saddle-node
bifurcations is at least $2^{2\ell} - \binom{2\ell}{\ell}$. 

\begin{figure}[h]
\centering
\includegraphics[width=\textwidth]{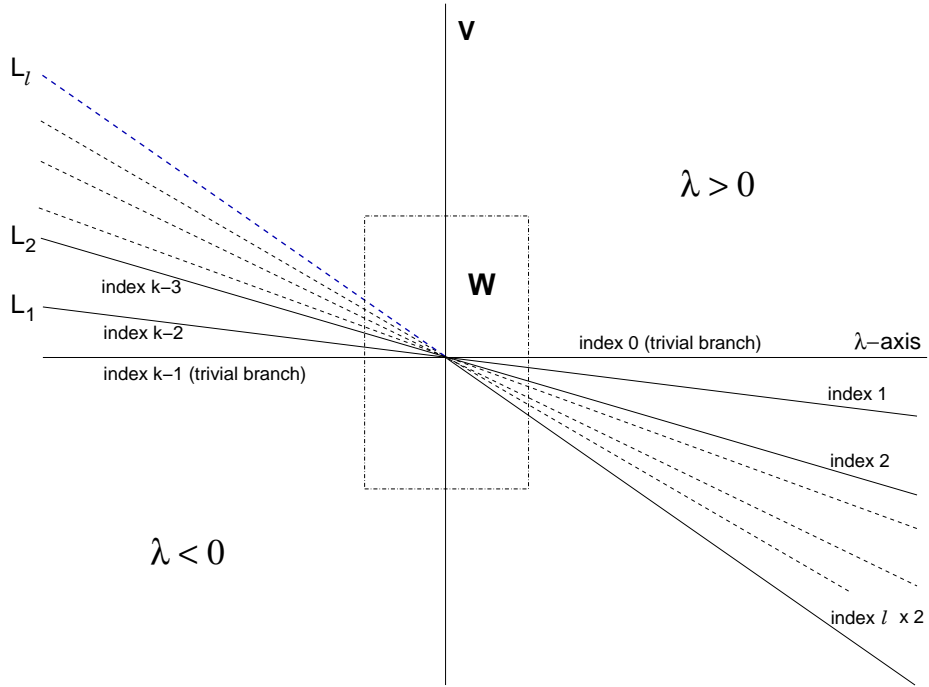}
\caption{Indices for branches, $k$ even. Branches along $S_k L_\ell$ occur only for $\lambda > 0$.
A crossing curve results from an index $\ell$ branch along $L_{\ell-1}$, $\lambda < 0$, joining with an index $\ell$ branch 
along $L_\ell$, $\lambda > 0$.}
\label{fig: brandata2}
\end{figure}

If $k = 2\ell$ is even, the analysis is slightly different as generically there are pitchfork bifurcations along the axes $S_k L_\ell$ and we need to
take account of cubic equivariants. If we take the family $f(\xx,\lambda) = \lambda \xx -Q(\xx) -c \|\xx\|^2\xx$, $c > 0$, then the
pitchfork bifurcation is supercritical and generates a curve of solution branches of index $\ell$ along $L_\ell$, $\lambda > 0$. 
We refer to Figure~\ref{fig: brandata2} for a schematic.  
Along similar lines to the above, the number of crossing curves is at most
$\binom{2\ell-1}{\ell}$ and the number of saddle-node
bifurcations is at least $2^{2\ell-1} - \binom{2\ell-1}{\ell}$.

\begin{Def}
(Assumptions and notation as above.) The family $\hat f$ is a \emph{minimal symmetry breaking model} for $f$ if
\begin{enumerate}
\item $f = \hat f$ on $V \times \real \smallsetminus W$.
\item $\hat f$ has exactly $\binom{k-1}{[k/2]}$ crossing curves.
\item $\hat f$ has exactly $2^{k-1} - \binom{k-1}{[k/2]}$ saddle-node bifurcations. 
\end{enumerate}
\end{Def}
\subsection{Minimal symmetry breaking model: $k$ odd}
\begin{thm}\label{thm: uodd}
Assume $k = 2\ell +1$ ($k$ is odd) and let $f_\lambda(\xx)=\lambda\xx - Q(\xx)$,  where
$Q$ is the standard quadratic equivariant on $(S_k,H_{k-1})$.  Let $\eta_0 > 0$ and define 
$\rho = 4\sqrt{\eta_0}$, $\delta_1 = \rho/2$, 
$\delta_2 = \sqrt{\frac{k}{3}}\rho$. Following~\Refb{eq: Wnbd}, define 
\[
W = W_\rho=\big([-\delta_1,\delta_1] \times \overline{D}_\delta(\is{0})\big) \times [-\rho,\rho]\subset (L_1 \oplus L_1^\perp) \times \real.
\]
There exist $C > 0$, depending only on $\eta_0$ and $k$, and $\eta_1 \in (0,\eta_0]$, such that for every $\eta \in (0,\eta_1]$, there is a smooth $S_{k-1}$-equivariant family
$\haf^\eta_{\lambda}$, satisfying
\begin{enumerate}
\item $\haf^\eta_{\lambda}(\xx) = f_\lambda(\xx)$ if 
$(\xx,\lambda) \notin W$.
\item $\|\haf^\eta- f\|_{W,2} < C\eta$.
\item The only bifurcations of the family $\haf^\eta_\lambda$ are saddle-node bifurcations and $\haf^\eta_\lambda$ 
is stable under $C^2$-small perturbations supported on a compact neighbourhood of $(\is{0},0)\in H_{k-1} \times \real$ (no assumption of equivariance).
\item The family $\haf^\eta_\lambda$ has exactly $2^{2\ell}-\binom{2\ell}{\ell}$ saddle node bifurcations and 
\emph{all} of the branches of solutions with index $\ne \ell$ will end or start with a saddle-node bifurcation that occurs in $W$.
\item There are exactly $\binom{2\ell}{\ell}$ crossing curves $(\xx,\lambda):\real \arr H_{k-1}\times \real$ of solutions to $\haf^\eta_\lambda = 0$; each of these curves 
consists of hyperbolic equilibria of index $\ell$. 
\end{enumerate}
\end{thm}

\begin{rems}
(1) Suppose $a \in \real$ is non-zero. 
Recall~\Refb{eq: stform} that the change of coordinates $\bar\xx = -a^{-1}\xx$, transforms $\bar \xx' = \lambda \bar \xx + aQ(\bar \xx)$ to $\xx' = \lambda \xx -Q(\xx)$. Hence Theorem~\ref{thm: uodd} 
gives minimal symmetry breaking models for all generic families $f_\lambda(\xx) = \lambda\xx + aQ(\xx)$, $a \ne 0$.\\
(2) The stability under $C^2$-small perturbations (rather than $C^1$) is required on account of the saddle-node bifurcations which are not necessarily preserved under $C^1$-small
perturbations. \\
(3) The interest of the result lies in small values of $\eta$ and $\rho, \delta_1,\delta_2$. In the proof, 
$\haf_\lambda^\eta = \haf_\lambda^{\eta_1}$ if
$\eta \ge\eta_1$ and $\haf_\lambda^\eta$ is smooth in $\eta$ on $(0,\eta_1]$. 
It is straightforward to modify the construction to obtain 
smooth dependence for $\eta\in (0,\infty)$.  \rend

\end{rems}

\subsubsection{Sketch of the proof} 
We break symmetry from $S_k$ to $S_{k-1}$ using a perturbation parallel to
$L_1$. Initially, we assume the perturbation is constant. 
Next follow a number 
of lemmas that describe the effect of the perturbation on the dynamics restricted to
a sequence of flow-invariant 2-planes. Most of these results will hold for $k$ odd or even. The final step is to localize the perturbation
to have support in $W$ and show that the localization  process does not introduce (or destroy)
solutions, or change stabilities, within $W$. With the possible exception of one detail, 
used for localization, the proof of minimal symmetry breaking model is elementary. 

\subsection{Minimal symmetry breaking model: preliminary results, $k$ odd or even}\label{sec: odd}

Set $V = H_{k-1} \subset \real^k$. For $p \in \is{k-1}$, set   $q = k-p$. If $k$ is odd (resp.~even), define $\mathfrak{L} = \ell +1$ (resp.$\mathfrak{L} = \ell$). 
Recall from Section~\ref{sec: gensk} that for $p \in \boldsymbol{\ell}$, $L_p$ is the axis of symmetry through $\boldsymbol{\vareps}_p$ and that
\[
L_p = \{(qx^p,-px^q) \in V \subset \real^k \dd x \in \real \} = V^{S_p \times S_q}
\]
For $p \le \mathfrak{L}$, define the $2$-plane $E_p\subset V $ by
\[
E_p = \{(x,y^{p-1},z^q) \dd x + (p-1)y + qz = 0\} = V^{S_1 \times S_{p-1} \times S_q}.
\]
In what follows, $S_{k-1} \defoo S_1 \times S_{k-1} \subset S_k$. The isotypic decomposition of the representation $(S_{k-1},V)$ is $\mathfrak{s}_{k-1} + \mathfrak{t}$,
where the trivial factor is $(L_1,S_{k-1})$.
\begin{lemma}\label{lem: symE1}
(Notation and assumptions as above.) 
\begin{enumerate}
\item For $2 \le p \le \ell$, $E_p$ contains exactly three axes of symmetry
$L_1, L_p$ and $L^\star_{p-1}$ where
\[
L^\star_{p-1} =\{(z, y^{p-1}, z^q) \dd (p-1)y + (q+1)z = 0\} = \sigma L_{p-1}
\]
and $\sigma = (1p)\in S_k$.
\item If $k$ is odd, $E_{\ell+1}$ contains exactly three axes of symmetry
$L_1, L_\ell$ and $L^\star_{\ell}$ where
\[
L^\star_{\ell} =\{(z, y^{\ell}, z^\ell) \dd (p-1)y + (q+1)z = 0\} = \sigma L_{\ell}
\]
and $\sigma \in S_{k-1}$. 
\item For the $S_{k-1}$-action  on $V$ and $p \le \mathfrak{L}$,
\[
E_p \smallsetminus L_1 = \{\xx \in V \dd (S_{k-1})_\xx = S_{p-1} \times S_q\}.
\]
\item For all $k \ge 4$ we have
\begin{eqnarray*}
\theta_{L_1,L_p}&=&\cos^{-1}\left(\sqrt{\frac{q}{(k-1)p}}\right)\in [\frac{\pi}{3},\frac{\pi}{2}) \\ 
\theta_{L^\star_{p-1},L_p}&=&\cos^{-1}\left(\sqrt{\frac{q(p-1)}{(q+1)p}}\right)\in (0,\cos^{-1}(1/3)]\\
\theta_{L^\star_{p-1},L_1}&=&\cos^{-1}\left(-\sqrt{\frac{p-1}{(k-1)(q+1)}}\right) \in \big(\frac{\pi}{2},\frac{2\pi}{3}\big].
\end{eqnarray*}
\item If $k = 3$, $p$ must be $\mathfrak{L} = 2$, and $E_2 = \real^2$ and $L_2, L_1, L_1^\star$ are the three axes of symmetry for the standard $S_3$-action on $\real^2$.
\item For $p \ne p'$, $2 \le p,p' \le \mathfrak{L}$, $E_p \cap E_{p'} = L_1$.
\end{enumerate}
\end{lemma}
\begin{proof} All statements are easy to verify and we omit the details. \end{proof}
\begin{rem}
Taking the $S_{k-1}$-action on $V$, statement (3) implies that non-zero points on the axes $L_p$, $L^\star_{p-1}$ have the same isotropy. Hence, there is the possibility of
$S_{k-1}$-equivariant deformations of the original $S_k$-equivariant family on $V$ that allow us to connect zeros on the axes $L_p$, $L^\star_{p-1}$ via a saddle-node bifurcation.
This observation lies at the core of our construction. The $S_{k-1}$-symmetry organizes the details.
\rend
\end{rem}
For $p \le \mathfrak{L}$, define the linear map $U_p:\real^2 \arr E_p$ by
\begin{eqnarray*}
U_p(u,0) & = & \frac{1}{\sqrt{k(k-1)}}\left((k-1)u,-u^{p-1},-u^q\right), \; u \in \real\\
U_p(0,v) & = &\frac{1}{\sqrt{(k-1)(p-1)q}}\left(0,qv^{p-1},-(p-1)v^q\right), \; v \in \real
\end{eqnarray*}
Observe that $\|U_p(1,0)\| = \|U_p(0,1)\| = 1$ and $U_p(u,0)\perp U_p(0,v)$ for all $(u,v)\in\real^2$. Hence $U_p$ maps $\real^2$ isometrically onto $E_p$.
Let $\{\ee_1 = (1,0),\, \ee_2 = (0,1)\}$ be the standard Euclidean basis of $\real^2$.
\begin{lemma}\label{lem: ax2}
(Assumptions and notation as above.)
\begin{enumerate}
\item $U(\ee_1) = \boldsymbol{\vareps}_1$ and $U_p(\real \ee_1)= L_1$.
\item $U_p^{-1}(L_p) = \{(u,m_p u) \dd u \in \real\}$, where $m_p = \sqrt{\frac{k(p-1)}{q}}$ and \\ $\langle U_p^{-1}(\boldsymbol{\vareps}_p),\ee_2  \rangle > 0$.
\item $U_p^{-1}(L^\star_{p-1}) = \{(-u,m_{p-1}^\star u) \dd u \in \real\}$, where $m_{p-1}^\star = \sqrt{\frac{qk}{p-1}}$, \\
$\langle U_p^{-1}(\boldsymbol{\vareps}^\star_{p-1}),\ee_2  \rangle > 0$, and $\boldsymbol{\vareps}_{p-1}^\star = (1p)\boldsymbol{\vareps}_{p-1}$.
\end{enumerate}
In particular, if we take the standard orientation of $\real^2$ and the orientation on $E_p$ defined by
$\boldsymbol{\vareps}_1, \boldsymbol{\vareps}_p$, $U_p$ preserves orientation.
\end{lemma}
\proof (1) is obvious by the definition of $U_p(u,0)$. For (2), note that since $p \ge 2$, $U_p(u,m_p u) \in L_p$ iff the first two components of $U_p(u,m_p u)$ are equal.
This leads to the condition
\[
\frac{k-1}{\sqrt{k(k-1)}}u = -\frac{u}{\sqrt{k(k-1)}} + \frac{qm_p u}{\sqrt{(k-1)(p-1)q}}, \; u \in \real.
\]
Dividing through by $u$ and simplifying, we obtain $m_p = +\sqrt{\frac{k(p-1)}{q}}$.

The first component of $U_p(u,m_pu)$ is strictly positive iff $u > 0$. Hence there exists $u > 0$ such that $\boldsymbol{\vareps}_p = U_p(u,m_pu)$ and so
$\langle U_p^{-1}(\boldsymbol{\vareps}_p),\ee_2  \rangle > 0$.  The proof of (3) is similar. \qed

\begin{figure}[h]
\centering
\includegraphics[width=\textwidth]{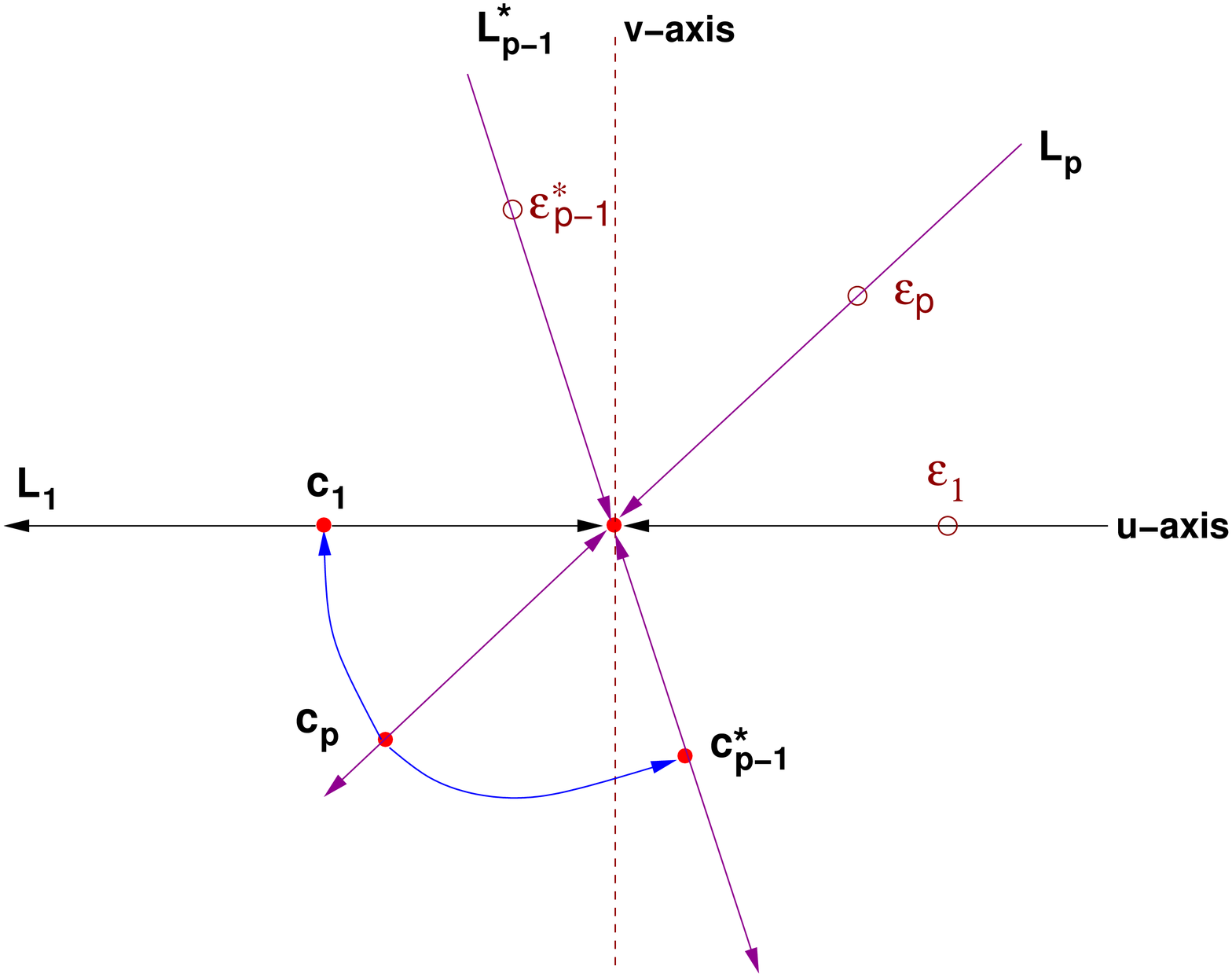}
\caption{Zeros and dynamics of $\xx' = \lambda \xx - Q(\xx)$ on $E_p$, $2 \le p \le \ell$, $\lambda < 0$.}
\label{fig: lines3}
\end{figure}

\begin{lemma}\label{lem: symE2}
(Notation and assumptions as above.) Suppose $\lambda < 0$ and denote the
zeros of $\xx' = \lambda \xx - Q(\xx)$ lying on $L_p$, $L^\star_{p-1}$ by $c_p$, $c_{p-1}^\star$ respectively.
For $2 \le p \le \ell$ there is are connections from $c_p$ to $c_1$ and $c_{p-1}^\star$ (see Figure~\ref{fig: lines3}). A similar result holds
when $\lambda > 0$ with connections now from $c_1$ and $c_{p-1}^\star$ to $c_p$.
\end{lemma}
\proof The fixed point space $E_p$ is invariant by the flow of $f_\lambda(\xx) = \lambda \xx  - Q(\xx)$. All the zeros of $f_\lambda$ occur on axes of symmetry,
since the coefficient of $Q$ is non-zero~\cite{FR1992}. The index of $c_1$ is $k-2$ and so the index of $c_1$ for $f_\lambda|E_p$ is $1$.  Observe that if the index of $c_p$ for $f_\lambda|E_p$
is $1$, there would have to be an additional zero for the phase vector field  in the interior of the region of $S^1\subset S(V)$ determined by the wedge defined by $c_1, c_p$
contradicting the result that all zeros of the phase vector field lie on axes of symmetry. Hence the index of $c_p$ for $f_\lambda|E_p$ is $0$ and from this
it follows easily the there is a connection from $c_p$ to $c_1$. Similar arguments show there must be a connection from  $c_p$ to $c_{p-1}^\star$.
The argument when $\lambda > 0$ is similar with the zeros $c_1\in L_1, c_p\in L_p, c_{p-1}^\star\in L_{p-1}^\star$ now lying on the other side of the origin. \qed

We need to compute the pull-back by $U_p$ of the family $(\lambda I_V -Q)| E_p$.  For this it suffices to compute the pull back $Q_p$ of $Q$ to $\real^2$ since $\lambda I_V$ pulls back to
$\lambda I_{\real^2}$.
\begin{lemma}\label{lem: sn0}
(Assumptions and notation as above.)
For $2 \le p \le \mathcal{L}$, $Q_p = (Z_1,Z_2)$ where
\begin{eqnarray}\label{eq: z1}
Z_1(u,v) & = & \frac{1}{\sqrt{k(k-1)}}((k-2)u^2-v^2) \\
\label{eq: z2}
Z_2(u,v) & = & -\frac{2}{\sqrt{k(k-1)}} uv + \frac{q-p+1}{\sqrt{q(k-1)(p-1)}}v^2
\end{eqnarray}
\end{lemma}
\proof Since $Q$ is a gradient vector field, and $U_p$ is an isometry,
it suffices to find the gradient of the pull back $C_p(u,v) = c(U_p(u,v))$ of the cubic $c: V \arr \real$ defining $Q$. Now 
$c(\xx) = \frac{1}{3}(\sum_{i\in\is{k}} x_i^3)$, $\xx \in V$, and so, taking $\xx = U_p(u,v)$, $(u,v) \in \real^2$, and an easy computation gives 
{\small
\[
C_p(u,v) = \frac{1}{3}\left(A^3(k-1)^3u^3 + (p-1)(-Au+Bqv)^3 - q(Au + B(p-1)v)^3\right), 
\]}\normalsize
where $A=1/\sqrt{k(k-1)}$ and  $B = 1/\sqrt{(k-1)(p-1)q}$.
After some elementary algebra, we find that find the components $Z_1,Z_2$ of $Q_p = \grad{C_p}$ are given by~(\ref{eq: z1},\ref{eq: z2}). \qed

\begin{prop}\label{prop: sn0}
In $(u,v)$-coordinates, the dynamics of $\xx' = \lambda \xx - Q(\xx)$ on $E_p$ is given by
\begin{eqnarray}\label{eq: Muu}
\ud & = & \lambda u  - \frac{1}{\sqrt{k(k-1)}}((k-2)u^2-v^2)\\
\label{eq: Mvu}
\vd & = & \lambda v + \frac{2}{\sqrt{k(k-1)}} uv - \frac{q-p+1}{\sqrt{q(k-1)(p-1)}}v^2
\end{eqnarray}
We have solution curves $c_1=(u_1,v_1), c_p=(u_p,v_p), c_{p-1}^\star=(u_{p-1}^\star,v_{p-1}^\star)$ for (\ref{eq: Muu},\ref{eq: Mvu}) given for $\lambda \in \real$ by
\begin{eqnarray}
(u_1, v_1)(\lambda) & = & \lambda \frac{\sqrt{k(k-1)}}{k-2}\ee_1\\
(u_p,v_p)(\lambda) &= &\lambda\sqrt{\frac{k}{k-1}} \frac{q}{q-p}\left(1 , \sqrt{\frac{k(p-1)}{q}}\right) \\
(u_{p-1}^\star,v_{p-1}^\star)(\lambda) & = & \lambda\sqrt{\frac{k}{k-1}}\frac{p-1}{q-p+2}\left(-1,\sqrt{\frac{kq}{p-1}}\right)
\end{eqnarray}
\end{prop}
\proof A straightforward computation using Lemma~\ref{lem: sn0}. \qed
\subsubsection{Forced symmetry breaking to $S_{k-1}$}
\begin{prop}\label{prop: sn1}
Let $\eta > 0$ and consider the perturbed equations
\begin{eqnarray}\label{eq: Mu}
\ud & = & \lambda u  - \frac{1}{\sqrt{k(k-1)}}((k-2)u^2-v^2) - \eta\\
\label{eq: Mv}
\vd & = & \lambda v + \frac{2}{\sqrt{k(k-1)}} uv - \frac{q-p+1}{\sqrt{q(k-1)(p-1)}}v^2
\end{eqnarray}
For $1 \le p \le \ell$, define
\begin{equation}
\gamma_{k,p} = \begin{cases}
& 2\sqrt{\eta}\frac{\sqrt{k-2}}{\sqrt[4]{k(k-1)}},\; p = 1\\
&2\sqrt{\eta}\frac{\sqrt{k-2}}{\sqrt[4]{k(k-1)}}\frac{q-p+1}{(k-1)}\sqrt{1 - \frac{4q(p-1)}{(q-p+1)^2k(k-2)}}, \; p > 1
\end{cases}
\end{equation}
\begin{enumerate}
\item[(a)] Suppose that $2 \le p < k/2$. The system
(\ref{eq: Mu},\ref{eq: Mv}) has saddle-node bifurcations at $\lambda = \pm \gamma_{k,p}$. Specifically,
at $\lambda = -\gamma_{k,p}$, the (perturbed) branches $c_p(\lambda,\eta), c^\star_{p-1}(\lambda,\eta)\subset E_p$ collide in a saddle-node bifurcation
as $\lambda \nearrow -\gamma_{k,p}$, and at $\lambda = +\gamma_{k,p}$, the branches $c_p(\lambda,\eta), c^\star_{p-1}(\lambda,\eta)$ are created
in a saddle-node bifurcation as $\lambda \nearrow \gamma_{k,p}$. Set $c_p(\pm\gamma_{k,p},\eta) = \pm\is{b}_p\in E_p$
\item [(b)]The system (\ref{eq: Mu},\ref{eq: Mv}) has saddle-node bifurcations along $L_1$ at $\lambda = \pm \gamma_{k,1}$.
Specifically,
at $\lambda = -\gamma_{k,1}$, the (perturbed) branch $c_1(\lambda,\eta)$ collides with the (perturbed) trivial solution branch in a saddle-node bifurcation
as $\lambda \nearrow -\gamma_{k,1}$ and at $\lambda = \gamma_{k,1}$, the branch $c_1(\lambda,\eta)$ and the perturbed trivial solution branch
are created in a saddle-node bifurcation as $\lambda \nearrow \gamma_{k,1}$. Set $\pm\is{b}_1 = c_1(\pm\gamma_{k,1},\eta)\in L_1$.
\item [(c)] For all $k \ge 3$, and $p \in [1,k/2)$, $|\gamma_{k,p}| \le 2\sqrt{\eta}$. For fixed $k$, $|\gamma_{k,p}|$ is a strictly monotone decreasing function of $p \in [1,k/2)$ and
\[
|\gamma_{k,p}| \in \left[2\sqrt{\eta}\frac{\sqrt{3}}{\sqrt{k}\sqrt[4]{k(k-1)}},2\sqrt{\eta}\frac{\sqrt{k-2}}{\sqrt[4]{k(k-1)}}\right]
\]
\item [(d)]$\|\is{b}_1\| \le \sqrt{\eta}$ and for all $p \in [1,k/2)$, 
\[
\|\is{b}_p\| < 2\sqrt{\frac{\eta k}{3}}.
\]
\end{enumerate}
\end{prop}
\begin{rems}
(1) If $k$ is even, then $\gamma_{k,\ell} = 0$. Cubic terms are needed to resolve this case. \\
(2) (a) only applies if $k > 4$; (b) applies for all $k \ge 3$. \\
(3) (c) implies that as $\lambda \nearrow 0$, there is a sequence of saddle-node bifurcations. The first  on $L_1$; the 
second is an $S_{k-1}$-orbit of the saddle-node bifurcation on $E_2$. The sequence ends with the $S_{k-1}$-orbit of the saddle-node bifurcation on $E_\ell$ (resp.~$E_{\ell-1}$)
if $k$ is odd (resp.~even). The order of the sequence is reversed when $\lambda$ increases through zero. \rend
\end{rems}

Before giving the proof of Proposition~\ref{prop: sn1}, we give a lemma that helps simplify and organize the computations.

\begin{lemma}\label{lem: trc}
(Assumptions of Prop.~\ref{prop: sn1}.)
Under the linear coordinate change $u = \frac{\sqrt{k(k-1)}}{k-2} \bar{u}$, $v =  \sqrt{\frac{k(k-1)}{k-2}} \bar{v}$, (\ref{eq: Mu},\ref{eq: Mv}) transforms to
\begin{eqnarray}\label{eq: MuS}
\dot{\bar{u}} & = & \lambda \bar{u}  - \bar{u}^2 + \bar{v}^2 - \bar{\eta}\\
\label{eq: MvS}
\dot{\bar{v}}& = & \lambda \bar{v} + \frac{2}{(k-2)} \bar{u}\bar{v} - C_p \bar{v}^2,
\end{eqnarray}
where
\begin{enumerate}
\item $\bar{\eta} = \frac{k-2}{\sqrt{k(k-1)}}\eta$ and $C_p = \frac{(q-p+1)}{\sqrt{q(p-1)}}\sqrt{\frac{k}{k-2}}$.
\item For fixed $k$, and $p \in [2,\ell]$, $C_p$ is strictly monotone decreasing in $p$ and
\begin{align*}
C_2& = \frac{k-3}{k-2}\sqrt{k}\\
C_\ell& = \begin{cases}
\frac{4\sqrt{k}}{(k+1)(k-1)(k-3)},\; \text{$k$ is odd}\\
\frac{2}{(k-2)},\; \text{$k$ even}.
\end{cases}
\end{align*}
\end{enumerate}
\end{lemma}
\proof A straightforward computation and we omit details. \qed

\pfof{Prop.~\ref{prop: sn1}.} Fix $p \in [2,\ell]$. Following Lemma~\ref{lem: trc}, transform to the equations (\ref{eq: MuS},\ref{eq: MvS}). We look
for solutions not lying on $L_1$. That is, solutions with $\bar{v} \ne 0$. It follows from  \Refb{eq: MvS} that
\[
\bar{u} = \frac{k-2}{2}(C_p\bar{v} - \lambda).
\]
Substituting for $\bar{u}$ in \Refb{eq: MuS}, we find that $\bar{v}$ satisfies the equation
\[
\bar{v}^2 ((k-2)^2 C_p^2- 4) -2\bar{v}\lambda C_p(k-2)(k-1) + \lambda^2k(k-2) + 4\bar{\eta} = 0
\]
This equation has a double root iff
\[
\lambda^2 C_p^2 (k-2)^2(k-1)^2 = \lambda^2 ((k-2)^2 C_p^2- 4) k(k-2)  +4((k-2)^2 C_p^2- 4)\bar{\eta}
\]
Solving for $\lambda$, and using the expressions for $C_p$ given in Lemma~\ref{lem: trc}, we find that
\[
\lambda = \pm 2\sqrt{\eta}\frac{\sqrt{k-2}}{\sqrt[4]{k(k-1)}}\frac{q-p+1}{(k-1)}\sqrt{1 - \frac{4q(p-1)}{(q-p+1)^2k(k-2)}}, \; p  > 1.
\]
It is straightforward to verify that these values of $\lambda$ define saddle-node bifurcation points $\pm \gamma_{k,p}$ for the perturbed branches associated to
$c_p, c_{p-1}^\star$, $\lambda < 0$, and the corresponding pair of branches for $\lambda > 0$.  The case $p = 1$ is easy to prove
directly---take $\bar{v} = 0$ in \Refb{eq: MuS}---but the expression for $\gamma_{k,1}$ follows by taking $p = 1$ in the formula for $\gamma_{k,p}$. 

The remaining statements of the proposition follow by straightforward, though lengthy, computation. For the estimate of $\|\is{b}_p\|$, we compute the $v$-coordinate of
$\is{b}_p = (u_p,v_p)$, $p > 1$, and prove that $|v_p| \le \sqrt{\frac{\eta k}{3}}$, all $p \in [1,k/2)$. Finally, we show that for $k/2 > p > 1$, $|u_p/v_p|$ is uniformly bounded by $1$. \qed
\subsubsection{The space $F_\ell=E_{\ell+1}$, $k$ odd.}
\begin{figure}[h]
\centering
\includegraphics[width=\textwidth]{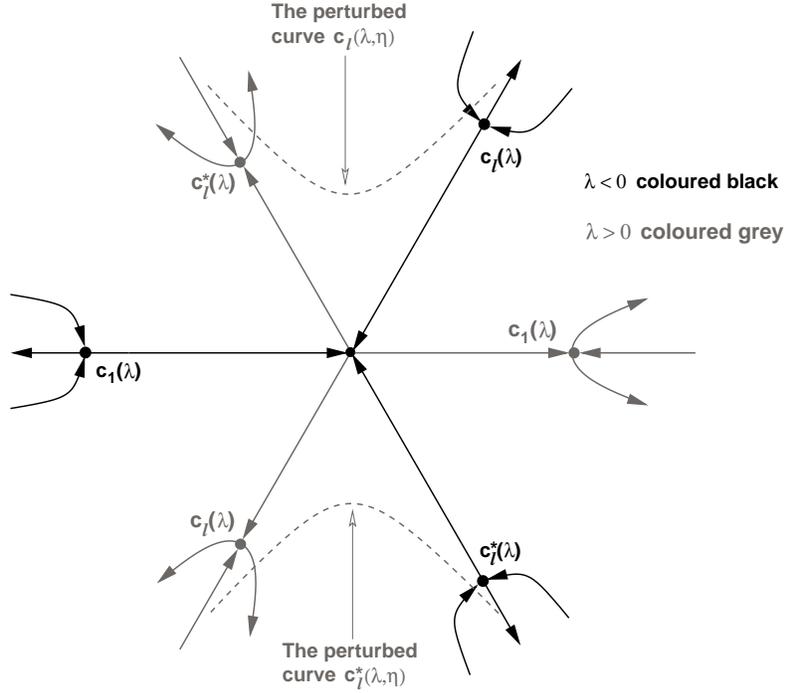}
\caption{Zeros and dynamics of $\xx' = \lambda \xx - Q(\xx)$ on $F_\ell$, $k = 2\ell + 1$. Dynamics for $\lambda < 0$ is shown in black, that for
$\lambda > 0$ in grey. The dashed line indicate the perturbed curves $c_\ell(\lambda,\eta)$ and $c^\star_\ell(\lambda,\eta)$. }
\label{fig: F2}
\end{figure}

We assume $k = 2\ell+1$ is odd and set $E_{\ell + 1} = F_\ell$ so that
\[
F_\ell = \{(x,y^\ell,z^\ell) \dd x + \ell y + \ell z = 0\}.
\]
Setting $U_{\ell+1} = T_\ell$, the isometry $T_\ell : \real^2 \arr F_\ell$ is given by
\begin{eqnarray*}
T_\ell (u,0)&=&\frac{1}{\sqrt{k(k-1)}} \big(2 \ell u,-u^\ell,-u^\ell\big),\; u \in \real \\
T_\ell (0,v)&=&\frac{1}{\ell\sqrt{2\ell}} \big(0,\ell v^\ell,-\ell v^\ell\big),\; v \in \real
\end{eqnarray*}
Recall from Lemma~\ref{lem: ax2} that
\begin{enumerate}
\item $T(\boldsymbol{\vareps}_1) = \ee_1$ and $T_\ell(\real\ee_1) = L_1$.
\item $T^{-1}_\ell(L_\ell^\star) = \{(u,-\sqrt{k} u) \dd u \in \real\}$.
\item $T^{-1}_\ell(L_\ell) = \{(u,\sqrt{k} u) \dd u \in \real\}$.
\end{enumerate}
\begin{prop}\label{prop: Fell}
In $(u,v)$-coordinates, dynamics of $\xx' = \lambda \xx - Q(\xx)$ restricted to $F_\ell$ is given by
\begin{eqnarray}\label{eq: Muul}
\ud & = & \lambda u  - \frac{1}{\sqrt{k(k-1)}}((2\ell - 1)u^2-v^2)\\
\label{eq: Mvul}
\vd & = & \lambda v + \frac{2}{\sqrt{k(k-1)}}uv 
\end{eqnarray}
Denote the
zeros of $\xx' = \lambda \xx - Q(\xx)$ lying on $L_1$, $L^\star_{\ell}$ and $L_{\ell}$ by $c_1(\lambda)$, $c_\ell^\star(\lambda)$ and $c_\ell(\lambda)$ respectively.
We have
\begin{enumerate}
\item $c_1(\lambda) =\lambda\left(\frac{\sqrt{k(k-1)}}{k-2},0\right) $.
\item $c_\ell^\star(\lambda) = \lambda\left(-\frac{\sqrt{k(k-1)}}{2},\frac{k\sqrt{k-1}}{2}\right) $.
\item $c_\ell(\lambda) = \lambda\left(-\frac{\sqrt{k(k-1)}}{2},-\frac{k\sqrt{k-1}}{2}\right) $.
\end{enumerate}
If $\lambda \ne 0$, then all zeros are of index $1$ within $F_\ell$ and
there are no connections between $c_1(\lambda)$, $c_\ell^\star(\lambda)$ and $c_\ell(\lambda)$ (see Figure~\ref{fig: F2}).
\end{prop}
\proof The proof uses the explicit expressions for the zeros together with (\ref{eq: Pvf1}---\ref{eq: Pvf3}) giving the
the index of zeros of the phase vector field and so stabilities in directions transverse to radial direction. \qed

\begin{rem}
When $k=3$, the dynamics shown in Figure~\ref{fig: F2} is that of the generic $S_3 = \mathbf{D}_3$ bifurcation on
$\real^2$. \rend
\end{rem}
\begin{prop}\label{prop: snlu}
Let $\eta > 0$ and consider the perturbed equations on $F_\ell$
\begin{eqnarray}\label{eq: Mul}
\ud & = & \lambda u  - \frac{1}{\sqrt{k(k-1)}}((k-2)u^2-v^2) - \eta\\
\label{eq: Mvl}
\vd & = & \lambda v + \frac{2}{\sqrt{k(k-1)}} uv 
\end{eqnarray}
In terms of the parameter $\eta$, we have a curve $c_\ell(\lambda,\eta)$ of zeros such that for $\lambda < 0$, $c_\ell(\lambda,\eta)$ is close to $c_\ell(\lambda)$ and
for $\lambda > 0$, $c_\ell(\lambda,\eta)$ is close to $c_\ell^\star(\lambda)$. The closest approach of  $c_\ell(\lambda,\eta)$ to $L_1$ occurs when
$\lambda = 0$, and then $c_\ell(0,\eta) = (0, \sqrt{\eta}\sqrt[4]{k(k-1)})$. A similar result holds for $c_\ell^\star(\lambda,\eta)$ with
$c_\ell^\star(0,\eta) = c_\ell(0,\eta)$.
\end{prop}
\proof Straightforward computation of the perturbed zeros not lying on the axis $L_1$.  \qed

\subsubsection{The $S_{k-1}$-equivariant family $f_\lambda^\eta(\xx)=\lambda \xx - Q(\xx) -\eta\boldsymbol{\vareps}_1$.}
If we regard $H_{k-1}$ as an $S_{k-1} (= S_1 \times S_{k-1})$-representation then the isotypic decomposition is
$H_{k-1}=L_1 \oplus L_1^\perp$, where $(L_1,S_{k-1})$ is the trivial representation and $(L^\perp_1,S_{k-1})$ is the standard representation
of $S_{k-1}$. Identify $L_1 = \real \boldsymbol{\vareps}_1$ with $\real$ and $L_1^\perp$ with $\real^{k-2}$. Let $\pi_1:\real^{k-1}\arr \real$ and $\pi_2:\real^{k-1}\arr\real^{k-2}$
denote the orthogonal
projections onto $\real$ and $\real^{k-2}$ respectively. Denote coordinates on $\real \times \real^{k-2}$ by $(x,\yy)$.

For $\xx_1,\xx_2 \in H_{k-1}$, define the $H_{k-1}$-valued symmetric bilinear form $B: H_{k-1}^2 \arr H_{k-1}$ by
\[
B(\xx_1,\xx_2) = \frac{1}{2}\left[ Q(\xx_1+\xx_2) - Q(\xx_1)-Q(\xx_2)\right].
\]
For all $\xx \in H_{k-1}$, $Q(\xx) = B(\xx,\xx)$ and for all $\xx_1,\xx_2 \in H_{k-1}$, $g \in S_k$, $B(g \xx_1,g \xx_2) = B(\xx_1,\xx_2)$.

Define $A \in L_{S_{k-1}}(\real^{k-2},\real^{k-2})$, $Q_1\in P^{(2)}_{S_{k-1}}(\real^{k-2},\real)$, and $Q_2 \in P_{S_{k-1}}^{(2)}(\real^{k-2},\real^{k-2})$ by 
\begin{eqnarray*}
A(\yy)& =& 2\pi_2 B(\boldsymbol{\vareps}_1,\yy) = \alpha \yy,\; \yy \in \real^{k-2} \\
Q_1(\yy)& =& \pi_1 Q(\yy), \; \yy \in \real^{k-2} \\
Q_2(\yy)& =& \pi_2 Q(\yy), \; \yy \in \real^{k-2}
\end{eqnarray*}
where $\alpha \in \real$ is uniquely determined since $(\real^{k-2},S_{k-1})$ is absolutely irreducible and so $A$ is a real multiple of $I_{\real^{k-2}}$, 
\begin{lemma}
(Notation and assumptions as above.) In $(x,\yy)$ coordinates, the system $\xx' = \lambda \xx - Q(\xx) -\eta\boldsymbol{\vareps}_1$ may be written as
\begin{eqnarray}\label{eq: sk-11}
x' & = & \lambda x - x^2 - Q_1(\yy) -\eta \\
\label{eq: sk-12}
\yy' & = & (\lambda + x\alpha)\yy - Q_2(\yy),
\end{eqnarray}
where $\alpha > 0$ and \Refb{eq: sk-12} is $S_{k-1}$-equivariant.
\end{lemma}
\proof If $\xx = (x,\yy)$, then $Q(\xx) = B(x\boldsymbol{\vareps}_1 + \yy,x\boldsymbol{\vareps}_1 + \yy)$ and the result follows easily by writing
$B(x\boldsymbol{\vareps}_1 + \yy,x\boldsymbol{\vareps}_1 + \yy)$ in terms of $Q(x\boldsymbol{\vareps}_1)=x^2\boldsymbol{\vareps}_1$, $Q(\yy)$ and 
$2xB(\boldsymbol{\vareps}_1,\yy)=xA(\yy) = x\alpha \yy$.  
To show $\alpha > 0$, either compute directly or use \Refb{eq: Mv} of Proposition~\ref{prop: sn1}. \qed
\begin{cor}\label{cor: sols}
For $\eta \ge 0$, the only zeros of the $S_{k-1}$-equivariant family $f^\eta_\lambda(\xx) =  \lambda \xx - Q(\xx) -\eta\boldsymbol{\vareps}_1$
are those given by Propositions~\ref{prop: sn1}, \ref{prop: snlu}. All zeros are hyperbolic except the saddle-node bifurcation points
listed in Proposition~\ref{prop: sn1}. 
\end{cor}
\begin{proof} It follows from~\cite{FR1989,FR1992} that for each non-zero value of $\lambda + x\alpha$, \Refb{eq: sk-12} 
has exactly $2^{k-2}-1$ non-trivial hyperbolic (within $\real^{k-2}$) solutions each of which lies on an axis of symmetry for the $S_{k-1}$-action and so on the union of the $S_{k-1}$-orbits of 
$E_p\cap (\{x\} \times \real^{k-2}\times \{\lambda\})$, $2 \le p \le \ell+1$.    For these solutions
to extend to solutions of (\ref{eq: sk-11}, \ref{eq: sk-12}), additional conditions may have to be satisfied (depending on $x, \eta , \lambda$) but
no new solutions can be created and so there are no solutions outside $S_{k-1}\big(\cup_{2 \le p \le \ell+1} E_p\big)$. 
This leaves the question of what happens if $\lambda = -x \alpha$. Since $Q_2$ is a quadratic $S_{k-1}$-equivariant and $k-1$ is \emph{even},
$Q_2(\yy)=0$ has solutions if $k > 4$. Substituting in \Refb{eq: sk-12}, we see that if $\eta \ge 0$ and $Q_1(\yy) = 0$, there are no solutions of
$f^\eta_\lambda$ unless $\lambda = x = 0$ (crossing solution). In other words, if $\lambda \ne 0$, $\lambda,x$ have to be of opposite sign if $Q_2(\yy) = \is{0}$ 
which they are not, see Figure~\ref{fig: F2}. 
\end{proof}
\begin{rems}
(1) If $\eta < 0$, then the argument at the end of proof of Corollary~\ref{cor: sols} fails. In this case we expect to find pitchfork bifurcations. This happens
already for the case $k = 3$ and additional symmetry breaking is then required to obtain a minimal model. \\
(2) The proof of Corollary~\ref{cor: sols} implicitly relies on the Bezout's theorem in that the number of solutions of the homogeneous equation
is determined by looking for solutions of the homogeneous equation $\lambda \xx -Q(\xx) = 0$. Introduction of
the term $-\eta$ can destroy solutions through saddle-node bifurcations but the solutions exist over the complexes and are
``pinned'' to the corresponding complexified fixed point space. See \cite[\S 4.9]{Field2007} for more details. \rend
\end{rems}

Finally, some elementary symmetry and combinatorics needed for the proof of Theorem~\ref{thm: uodd}.

\begin{lemma}\label{lem: arith}
(Notation and assumptions as above.) Regard $S_{k-1}$ as the subgroup $S_1\times S_{k-1}$ of $S_k$ and assume $\ell + 1\ge p \ge 2$.
\begin{enumerate}
\item For all $\sigma \in S_{k-1}$, $\sigma|L_1 = I_{L_1}$ and so $(L_1,S_{k-1})$ is the trivial representation of $S_{k-1}$. In particular, 
for all $\sigma,\tau\in S_{k-1}$, $L_1 \subset \sigma E_p \cap \tau E_p$.
\item If $P_1 = \sigma E_p$, $P_2 = \nu E_p$, then $P_1 = P_2$ iff $\sigma\nu^{-1} \in S_{p-1} \times S_q\subset S_{k-1}$.
\item There are $\binom{k-1}{p-1}$ distinct planes in the $S_{k-1}$-orbit of $E_p$.
\item $\sum_{j=0}^\ell (-1)^{j} \binom{2\ell+1}{j} = (-1)^\ell\binom{2\ell}{\ell}$, $\sum_{j=0}^\ell (-1)^{j} \binom{2\ell}{j} = (-1)^\ell\binom{2\ell-1}{\ell}$.
\end{enumerate}
\end{lemma}
\proof (1) is immediate since if $\xx \in L_1$, the $S_k$ isotropy group of $\xx$ contains $S_{k-1}$. For (2) it suffices to recall that $E_p\smallsetminus L_1 = V^{S_{p-1} \times S_q}$.
Hence the set of distinct planes in the $S_{k-1}$-orbit of $E_p$ has cardinality $\binom{k-1}{p-1}$, proving (3).
Finally (4) results from the binomial identity $\binom{m}{n} = \sum_{j=0}^n (-1)^{n-j} \binom{m+1}{j}$ (which follows
easily by induction using the identity $\binom{m}{n} =\binom{m-1}{n} + \binom{m-1}{n-1}$). \qed

\subsection{\pfof{Theorem~\ref{thm: uodd}}}\mbox{ }\\ We shall assume that $k = 2\ell +1 \ge 3$ 
(most of the arguments below are valid for $p < \ell$ if $k$ is even). Fix $\eta > 0$ and consider the $S_{k-1}$-equivariant family
\begin{equation}\label{eq: famsk-1}
f^\eta_\lambda(\xx) = \lambda(\xx) = \lambda \xx -Q(\xx) - \eta \boldsymbol{\vareps}_1
\end{equation}
The first step is to show that~\Refb{eq: famsk-1} satisfies (4,5) of Theorem~\ref{thm: uodd}. For $p \in [0,\ell+1]$, set
$\chi(p) = (-1)^p \big[\sum_{j=0}^p (-1)^j \binom{2\ell+1}{j}\big]$. With the notation of Proposition~\ref{prop: sn1},
$f^\eta_\lambda$ has exactly $2^{k-2}$ hyperbolic zeros if $\lambda < -\gamma_{1,k}$. As we increase $\lambda$ there is, by Prop.~\ref{prop: sn1}(2), a saddle-node bifurcation at 
$(\is{b}_1,-\gamma_{1,p})$ in $L_1$ resulting from
the collision of the trivial solution branch and the branch $c_1(\lambda, \eta)\subset L_1$ at $\lambda = -\gamma_{1,k}$. If $\ell = 1$ ($k=3$), we have captured all
the single ($\chi(0)$) saddle-node bifurcation that occurs for $\lambda \le 0$ and $\chi(1) = 2$ branches remain of index 1. If $\ell \ge 2$, we continue to increase $\lambda$, and the remaining
$\chi(1)$ branches of index $k-2$ will collide with $\chi(1)$ branches of index $k-3$ in $\chi(1)$-saddle-node bifurcations all occurring at
$\lambda =-\gamma_{2,k} > -\gamma_{1,k} $. More precisely, by Prop.~\ref{prop: sn1}(1) the curves $c_1^\star(\lambda,\eta),c_2(\lambda,\eta) \subset E_2$ collide in a saddle-node
bifurcation at $(\is{b}_2,-\gamma_{2,k})$;  the other saddle-node bifurcations lie on the $S_{k-1}$-orbit of $(\is{b}_2,-\gamma_{2,k})$. There will be
$\chi(2)$ remaining branches  of index $k-3$. Proceeding inductively,
at the $p$th stage, assuming $p \le \ell$, there will be $\chi(p)$-branches of
index $k-p$ colliding with $\chi(p)$ branches of index $k-p-1$ in $\chi(p)$-saddle-node bifurcations all occurring at
$\lambda = -\gamma_{p,k}$. The set of saddle-node bifurcations is given by Prop.~\ref{prop: sn1}(1) and is
the $S_{k-1}$-orbit of $(\is{b}_p,-\gamma_{p,k})$. The process terminates when $p = \ell$. We are then left with $\chi(\ell+1)$ hyperbolic branches 
of index $\ell$. These branches 
connect with hyperbolic branches of index $\ell$, defined for all $\lambda \ge 0$, by Prop.~\ref{prop: snlu}. Specifically, 
the branches $c_\ell(\lambda,\eta), c^\star_\ell(\lambda,\eta) \subset F_\ell$ are connected at $\lambda = 0$ and then we use $S_{k-1}$-equivariance to obtain the
remaining $\chi(\ell)-1$ crossing branches. To complete the process, we now reverse the preceeding steps as we increase $\lambda$  through the sequence
of bifurcation points $\gamma_{\ell,k} < \ldots < \gamma_{1,k}$. Application of Lemma~\ref{lem: arith}(4) then gives statements (3,4,5) of Theorem~\ref{thm: uodd}
(with $\eta_1 = \eta_0=+\infty$).

It remains to prove that for all $\rho > 0$, a family $\haf^\eta_\lambda$ can be constructed, using a perturbation supported on a neighbourhood $W_\rho$ of
$(\is{0},0) \in V \times \real$, so as to satisfy all the statements of Theorem~\ref{thm: uodd}. 

Fix $\eta_0 > 0$ and set $\rho = 4\sqrt{\eta_0}$, $\delta_1 = \rho/2$, $\delta_2 = \sqrt{\frac{k\eta_0}{3}}$ and 
$W = W_\rho$ as in the statement of the theorem. It follows from Prop.~\ref{prop: sn1}(3,4) that for all $0 \le \eta \le \eta_0$, the bifurcation points 
of $f_\lambda^\eta$ are contained in $W_{\rho/2}$. That is, for all $p \in [1,\ell]$,
$S_{k-1}(\pm \is{b}_p,\pm \gamma_{p,k}) \subset W_{\rho/2}$.

Choose a $C^\infty$ function $\varphi: \real\arr\real$ such that
\begin{eqnarray}\label{eq: x1}
\varphi(t) &=&
\begin{cases}
1,\; \text{if } t \le 1\\
 0,\;  \text{if } t \ge 2
\end{cases}\\
\label{eq: x2}
\varphi(t)& \in & (0,1),\; \text{if } t \in (1,2) \\
\label{eq: x3}
\varphi'(t) & \le & 0, \; \forall t \in \real.
\end{eqnarray}
Define the smooth $S_{k-1}$-equivariant vector field  $\hat{\eta}$ on $(L_1 \oplus L_1^\perp) \times \real$ by
\begin{equation*}
\hat{\eta}((x,\yy),\lambda) = \varphi\big(\frac{2\lambda}{\rho}\big)\varphi\big(\frac{2x}{\delta_1}\big)\varphi\big(\frac{2\|\yy\|}{\delta_2}\big)\eta \boldsymbol{\vareps}_1,\; (x,\yy)\in L_1 \oplus L_1^\perp, \lambda \in \real
\end{equation*}
Observe that $\hat{\eta}|W_{\rho/2} = \eta \boldsymbol{\vareps}_1$ and $\hat{\eta}| (V\times\real \smallsetminus W_\rho) \equiv 0$.

If we replace $\eta  \boldsymbol{\vareps}_1$ by $\eta_1(\lambda) = \varphi\big(\frac{2\lambda}{\rho})\eta  \boldsymbol{\vareps}_1$ it is easy to see that $\eta_1$ is supported in
$H_{k-1} \times [-\rho,\rho]$ and that conditions (2---4) of Theorem~\ref{thm: uodd} hold with $\eta\in (0,\eta_0]$.  Turning to the vector field $\hat{\eta}$, the argument used for
Corollary~\ref{cor: sols} shows that no new zeros are introduced---first add the $\varphi\big(\frac{2x}{\delta_1})$ multiple, and use $S_{k-1}$-equivariance. 
Then add $\varphi\big(\frac{2\|\yy\|}{\delta_2}$ and 
note that for all $\eta \le \eta_0$, no new zeros are created in $W_\rho\smallsetminus W_{\rho/2}$ (using Propositions~\ref{prop: sn1}, \ref{prop: snlu}). 
However, there is the possibility that stabilities of solutions
could be changed in $W_\rho\smallsetminus W_{\rho/2}$, $\lambda \in [-\rho,\rho]$, on account of the $x$ and $\yy$ derivatives of $\varphi$ that occur. However, since these 
derivatives of $\hat{\eta}$
are supported on a compact set, disjoint from $(\is{0},0)\in H_{k-1}\times \real$, and all multiplied by $\eta$, we can choose $C > 0$ (which will depend on $\sqrt{\eta_0}$) and 
$\eta_1 \in (0,\eta_0]$ so that (1---5) of Theorem~\ref{thm: uodd} are satisfied. \qed

\subsection{Theorem~\ref{thm: uodd} and the Poincar\'e-Hopf theorem}
Let $\mathfrak{ind}_\zz(X)$ denote the Poincar\'e-Hopf index of a hyperbolic zero $\zz$ of the vector field $X$ ($\mathfrak{ind}_\zz(X)=+1$ (resp.~$-1$) if the the index
of $X$ at $\zz$ is even (resp.~odd), see~\cite[\S 6]{Milnor1965}).  Assume $k = 2\ell+1$ and let $\is{Z}(f_\lambda)$ denote the zero set of $f_\lambda = \lambda\xx-Q(\xx)$.
Since all the zeros of $f_\lambda=\lambda \xx - Q(\xx)$ are non-singular for $\lambda \ne 0$, $-I_{k-1}$ is isotopic to $I_{k-1}$,  and we may assume all zeros lie inside the 
sphere $S^{k-2}$ of radius $1$ for $|\lambda|$ sufficiently small, it follows that
$\sum_{\zz \in Z(f_\lambda)} \mathfrak{ind}_\zz(f_\lambda)$ is constant on $\real$. Either straightforward direct computation, or Theorem~\ref{thm: uodd}(5), shows that
$\sum_{\zz \in Z(f_\lambda)} \mathfrak{ind}_\zz(f_\lambda) = (-1)^{\ell+1} \binom{2\ell}{\ell}$ and so
any perturbation of $f_\lambda$ to a $C^2$-stable family must have at least $\binom{2\ell}{\ell}$ solutions 
for each $\lambda \in \real$. 

\subsection{Minimal symmetry breaking model: $k$ even}
Assume $k=2\ell$ is even. Since $Q | L_\ell \equiv 0$, we need to take account of higher order terms.
Recall~\cite[\S\S 16, 17]{FR1992} that for $k \ge 4$ a basis for $P^{(3)}_{S_k}(H_{k-1},H_{k-1})$ is 
given by $\{T_1, T_2\}$ where
\begin{eqnarray*}
T_1(\xx) & = & \|\xx\|^2\xx,\\
T_2(\xx) & = & (x_i^3 - \frac{1}{k}\sum_{j\in \is{k}} x_j^3),\; \xx = (x_1,\ldots,x_k) \in H_{k-1}.
\end{eqnarray*}
\begin{lemma}
(Notation and assumptions as above.) For $p \in [1,\ell]$,
\begin{eqnarray*}
T_1(\boldsymbol{\vareps}_p) & = & \boldsymbol{\vareps}_p \\
T_2(\boldsymbol{\vareps}_p) & = & \alpha_p \boldsymbol{\vareps}_p,
\end{eqnarray*}
where $\alpha_p = \frac{1}{k}\big(\frac{p}{q}+\frac{q}{p} - 1\big) \in [\frac{1}{k}, 1]$. $\alpha_p$ is strictly monotone decreasing on $[1,\ell]$ with
minimum value of $\alpha_\ell=\frac{1}{k}$ and maximum 
value of $\alpha_1=1 - \frac{1}{k} + O(\frac{1}{k^2})$.
\end{lemma}
\begin{proof} The statement for $T_1$ is trivial. The expression for $T_2$ is a straightforward computation using~\Refb{eq: vareps} and the definition of $T_2$. \end{proof}

Every $T \in P^{(3)}_{S_k}(H_{k-1},H_{k-1})$ may be written uniquely as $T = \alpha_1 T_1 + \alpha_2 T_2$, $\boldsymbol{\alpha} = (\alpha_1,\alpha_2)\in\real^2$. 
For $p \in [1,\ell]$, set $\beta_p=\beta_p(T) = \alpha_1 + \alpha_2 \alpha_p$. Define the open and dense subset $\mathcal{T}_3$ of $P^3_{S_k}(H_{k-1},H_{k-1})$ to consist of all 
$T$ for which $\beta_\ell(T) \ne 0$. Since $T(\boldsymbol{\vareps}_\ell) = \beta_\ell \boldsymbol{\vareps}_\ell$, 
\[
T \in \mathcal{T}_3 \;\;\text{iff  } T(\boldsymbol{\vareps}_\ell) \ne \is{0}.
\]
\begin{prop}\label{prop: cubic}
Let $T \in \mathcal{T}_3$ and
\begin{equation}\label{eq: cub}
F_\lambda(\xx) = \lambda \xx - Q(\xx) + T(\xx)
\end{equation}
\begin{enumerate}
\item If $\beta_\ell < 0$ (resp.~$\beta_\ell > 0$), then~\Refb{eq: cub} has a supercritical (resp.~subcritical) pitchfork bifurcation along $L_\ell$. The branches
are given for $t \in [0,\infty)$ by
\begin{eqnarray*}
\lambda(t) & = & -\text{\rm sgn}(\beta_\ell)t^2\\
\xx^\pm_s(t) & =& \pm \frac{t}{\sqrt{\text{\rm sgn}(\beta_\ell)\beta_\ell}} \boldsymbol{\vareps}_\ell 
\end{eqnarray*}
\item If $p \in [1,\ell-1]$ and $\beta_p = 0$, the branch along $L_p$ is the same as that for $\xx' = \lambda\xx-Q(\xx)$ (Section~\ref{sec: gensk}, (B,F))
and there are no other non-trivial zeros of \Refb{eq: cub} on $L_p$.
\item If $p \in [1,\ell-1]$ and $\beta_p < 0$, then the forward branch of $\xx' = \lambda\xx-Q(\xx)$ along $L_p$ perturbs to $(\xx^+_p(t),t) = (t(\frac{\sqrt{pqk}}{q-p}+O(t))\boldsymbol{\vareps}_p,t)$
and is defined for all $t \ge 0$. There is also a branch $(\zz^+_p(t),t)$  along $L_p$,  for which
$\|\zz^+_p(t)\|- \|\xx^+_p(t)\| \ge \frac{p-q}{\beta_p \sqrt{pqk}}$, all $t \ge 0$.

The backward branch of $\xx' = \lambda\xx-Q(\xx)$ along $L_p$ is perturbed to $\xx^-_p(t) = -t(\frac{q-p}{\sqrt{pqk}}+O(t))\boldsymbol{\vareps}_p$ and is defined for 
$t \in [0,-\frac{(q-p)^2}{4pqk\beta_p}]$. At $t = -\frac{(q-p)^2}{4pqk\beta_p}$, the branch collides with the branch $\tilde{\zz}^-_p(t)$ along $L_p$ in a saddle node bifurcation
and neither branch is defined for $t > -\frac{(q-p)^2}{4pqk\beta_p}$.  
We have $\|\zz^+_p(0)\| = \frac{q-p}{2|\beta_p|\sqrt{pqk}}$, 
and for all $t < -\frac{(q-p)^2}{4pqk\beta_p}$, $\|\tilde\zz^-_p(t)\|-\|\tilde\xx^-_p(t)\| > 0$.
Similar statements hold when $\beta_p > 0$.
\end{enumerate}
\end{prop}
\begin{proof} We omit the straightforward computation. \end{proof}
\begin{cor}\label{cor: ests}
(Notation and assumptions as above.)
Suppose $T=- T_1$ and set $\lambda_0 = \frac{4}{k(k^2-4)}, R_0 = \sqrt{\lambda_0}$.  The only branches of solutions to \Refb{eq: cub} meeting 
$D_{R_0}(\is{0}) \times (-\lambda_0,\lambda_0) \subset H_{k-1} \times \real$ are the $S_k$-orbits of the 
supercritical branches $\xx^\pm_s$  and perturbed branches $\xx^{\pm}_p$, $p\in[1,\ell-1]$,
given by  Proposition~\ref{prop: cubic}. In particular, the branches $(\xx^{\pm}_p(t),\pm t)$, $t \in [0,\lambda_0)$, are contained in $D_{R_0}(\is{0}) \times (-\lambda_0,\lambda_0)$
for all $p \in [1,\ell-1]$. The same result holds with  $T = T_1$ (subcritical pitchfork).  \\
Choosing $\lambda_0 > 0$ smaller if necessary, we may require that the indices of the branches $\xx_p^\pm(t)$, 
$p < \ell$, and $\xx_s^\pm(t)$  are constant on $(0,\lambda_0]$. 
\end{cor}
\proof The proof follows from Proposition~\ref{prop: cubic} or by direct computation of the branches. \qed

\begin{rem}
Unlike what happens when $k$ is odd, there is no \emph{natural} family $f: H_{k-1} \times \real\arr H_{k-1}$ for $S_k$-equivariant bifurcation 
if $k$ is even since the addition of higher order terms typically results in the appearance of additional solution branches 
which can and do merge with the branches of interest along $L_p$, $p < \ell$.  However, as indicated by the proposition, the signed indexed branching
pattern is uniquely determined  by the sign of $\beta_\ell$. In particular,
we may replace $T \in \mathcal{T}_3$ by $\sgn(\beta_\ell)T_1$ without changing the signed indexed branching pattern. What we shall do is
modify $\lambda\xx-Q(\xx) \pm \|\xx\|^2\xx$ to define an $S_k$-equivariant family that models the bifurcation and has \emph{only} the branches along $S_k L_p$, $p < \ell$,
given by the previous model for $k$ odd, and only super- or subcritical branching along $S_k L_\ell$.   
\rend
\end{rem}

Let $\is{Z} = \is{Z}(\mathcal{P}_Q)\subset S^{k-2}$ 
be the zero set of $\mathcal{P}_Q$. 
Let $\rho$ denote the standard $\On{k-1}$-invariant metric on $S^{k-2}$ and set
$\kappa = \min_{\uu, \vv \in \is{Z}, \uu\ne\vv} \rho(\uu,\vv)$. For $\tau > 0$,  define the $S_\ell \times S_\ell$-invariant
closed neighbourhood $B_\tau$ of $\boldsymbol{\vareps}_\ell$ by
\[
B_\tau = \{ \uu \in S^{k-2} \dd \rho(\uu_\ell,\boldsymbol{\vareps}_\ell) \le \tau\}. 
\]
Choose $\tau \ll \kappa$ , for example $\tau = \kappa/100$, so that the $S_{k-2}$-orbit of $B_\tau$ is a set of disjoint disc neighbourhoods of 
the zeros in $S_k \boldsymbol{\vareps}_\ell \subset \is{Z}(\mathcal{P}_Q)$ and 
$(\cup_{g \in S_k}B_\tau) \cap  \is{Z}(\mathcal{P}_Q) = S_k \boldsymbol{\vareps}_\ell$.
Choose a smooth $S_\ell\times S_\ell$-invariant function $\psi: B_\tau\arr \real$ satisfying
\begin{eqnarray*}
\psi(\uu) & = & 1,\; \rho(\uu, \boldsymbol{\vareps}_\ell) \le \tau/2\\
& \in & (0,1),\; \rho(\uu, \boldsymbol{\vareps}_\ell) \in (\tau/2, 3\tau/4)\\
& = & 0 ,\;\rho(\uu, \boldsymbol{\vareps}_\ell) \in [3\tau/4.\tau]
\end{eqnarray*}

Extend $\psi$ $S_k$-equivariantly to $S_k B_\delta$ and then to $S^{k-2}$ by taking $\psi\equiv 0$ on $S^{k-2}\smallsetminus S_k B_\tau$.
Thus $\psi: S^{k-1}\arr [0,1] \subset \real$ will be a $C^\infty$ $S_k$-equivariant map equal to $1$ on $S_kB_{\tau/2}$ and equal to zero outside $S_k B_{3\tau/4}$.

Choose $\varphi \in C^\infty(\real)$ satisfying (\ref{eq: x1}---\ref{eq: x3}). Define the $C^\infty$ $S_k$-equivariant radial vector field $S$ on $H_{k-1}\times \real$ by
\[
S(\xx,\lambda) =  [\varphi(2R/R_0)\varphi(2\lambda/\lambda_0) + (1-\varphi(2R/R_0)\varphi(2\lambda/\lambda_0))\psi(\uu)]T_1(\xx),
\]
where $R_0,\lambda_0> 0$ are given by Corollary~\ref{cor: ests} and $\xx = R\uu$ ($R = \|\xx\|$, and $\uu = \xx/\|\xx\|$, $\xx \ne \is{0}$).
Define the family
$F_\lambda^\pm$ on $H_{k-1}$ by
\begin{equation}\label{eq: mod}
F^\pm_\lambda(\xx) = \lambda \xx - Q(\xx) \pm S(\xx,\lambda).
\end{equation}
\begin{prop}\label{prop: model}
The family $F^\pm_\lambda$ has exactly $2^k-2$ non-trivial solution branches. All of these branches are axial and   
\begin{enumerate}
\item If $p \in [1,\ell-1]$, then  the forward (resp.~backward) branch 
$(\xx_p^+(t),t)$ (resp.~$(\xx_p^-(t),-t)$) along $L_p$ is defined for all $t \ge 0$ and 
consists of hyperbolic equilibria of index $p$ (respectively. $k-p-1$).
The eigenvalues of  $DF^\pm_{\pm t,\xx^\pm_p(t)}$ corresponding to eigendirections transverse to $L_p$ are given by the
eigenvalues of the Hessian of $\mathcal{P}_Q$ at $\pm\boldsymbol{\vareps}_p$.   \\
These results are  independent of the choice of $F_\lambda^+$ or $F_\lambda^-$.
\item If $p = \ell$, then $F_\lambda^-$ (resp.~$F_\lambda^+$) has a supercritical (resp.~subcritical) pitchfork bifurcation 
along $L_\ell$ and the branches are given by $(\xx_s^\pm(t),\lambda(t)) = (\pm t \boldsymbol{\vareps}_\ell,t^2)$ (resp.~$(\xx_s^\pm(t),\lambda(t)) = (\pm t \boldsymbol{\vareps}_\ell,-t^2)$),
$t \ge 0$. These branches are hyperbolic of index $\ell$. 
\end{enumerate}
\end{prop}
\begin{proof} Transform \Refb{eq: mod} to spherical coordinates $(R,\uu)$ to obtain 
\begin{eqnarray} \label{eq: req}
R' & = & \lambda R -R^2\langle Q(\uu),\uu\rangle \pm R^3 H(R,\uu,\lambda)\\
\label{eq: ueq}
\uu' & = & R\mathcal{P}_Q(\uu),
\end{eqnarray}
where the $C^\infty$ scalar function $H$ takes values in $[0,1]$ and is equal to zero iff (a) $R \ge R_0$ or $|\lambda| \ge \lambda_0$,  
and (b) $\uu \notin S_k B_{3\tau/4}$. If $(\hat R, \hat \uu)$ is a zero of (\ref{eq: req},\ref{eq: ueq}) with $\hat R \ne 0$,
then $\mathcal{P}_Q(\hat \uu) = 0$. Hence $\hat \uu$ (and $\hat \xx = \hat R \hat \uu$) must lie on an axis of symmetry~\cite{FR1992}.
Substituting in \Refb{eq: req}, and cancelling an $\hat R$ factor, $\hat R$ satisfies
\begin{equation}\label{eq: rad}
\lambda  -\hat R\langle Q(\hat \uu),\uu\rangle \pm \hat R^2 H(\hat R,\hat \uu,\lambda) = 0
\end{equation}
If $\hat \uu \in S_kL_\ell$, then $Q(\hat \uu) = 0$ and so, since $H$ is equal to one on a neighbourhood of $S_kL_\ell \times \real$ in
$H_{k-1}\times \real$, $\lambda \pm \hat R^2 = 0$, proving statement (2) of the proposition. 

It remains to complete the proof of (1). 
The solutions of $F^\pm(\xx,\lambda) = 0$ on $\{(\xx,\lambda) \in H_{k-1} \times \real \dd S(\xx,\lambda) = \is{0}\}$ 
are given by the solutions of $f(\xx,\lambda)=\lambda\xx -Q(\xx) = 0$---that is, the solution branches
of $f = 0$ along axes in $\cup_{p \in [1,\ell-1]} S_k L_p$.  It follows from Corollary~\ref{cor: ests} and the definition of $S$ that
the only zeros of \Refb{eq: req} with $\uu = \hat \uu$ and $\xx,\lambda$ lying in the support of $S$ are those described by
Corollary~\ref{cor: ests}. 
\end{proof}
\begin{rem} 
Every solution of the family \Refb{eq: mod} lies on a solution branch starting at
the bifurcation point $(\is{0},0) \in H_{k-1}\times \real$ and defined for all $t \ge 0$. The index and direction of branching
are constant on each branch and there are no spurious solutions resulting from the presence of higher order polynomial terms. 
\rend
\end{rem}

Just as for the case when $k$ is odd, we consider symmetry breaking perturbations of $F^\pm$ that are of the form 
$F^{\eta,\pm} = F^\pm  - \eta \boldsymbol{\vareps}_1$, where $ \eta > 0$. The main new feature 
is the effect of the perturbation on the supercritical branches that occur along  axes in $S_k L_\ell$.

\begin{thm}\label{thm: ueven}
Assume $k = 2\ell$ is even and let $F^\pm(\xx,\lambda)$ be the model \Refb{eq: mod}
as described in Proposition~\ref{prop: model}.  Given a compact neighbourhood $W$ of $(\is{0},0) \in H_{k-1} \times \real$,
there exist $C, \eta_1 > 0$ (depending on $W$ and $k$) such that for $\eta \in (0,\eta_1]$ there is a smooth $S_{k-1}$-equivariant family $F^{\pm,\eta}$ satisfying
\begin{enumerate}
\item $F^{\pm,\eta}_\lambda(\xx) = F^{\pm,\eta}_\lambda(\xx)$ if $(\xx,\lambda) \notin W$.
\item $\|F^{\pm,\eta}-F^\pm\|_{W,3} \le C \eta$.
\item The only bifurcations of the family $F^{\pm,\eta}_\lambda$ are saddle-node bifurcations and $F^{\pm,\eta}_\lambda$ is stable under
$C^2$-small perturbations supported on a compact neighbourhood of  $(\is{0},0) \in H_{k-1} \times \real$.
\item The family $F^{\pm,\eta}_\lambda$ has exactly $2^{2\ell -1} -\binom{2\ell -1}{\ell}$ saddle node bifurcations. Specifically,
all of the branches of solutions with $\text{\em index} \ne \ell$ will end or start with a saddle-node bifurcation (connecting possibly to
a branch of {\em index} $\ell$).
\item There are exactly $\binom{2\ell-1}{\ell}$ crossing curves $(\xx,\lambda):\real \arr H_{k-1} \times \real$
of solutions to $F^{\pm,\eta}=0$; each of these curves consists of hyperbolic saddles of {\em index} $\ell$.
\end{enumerate}
\end{thm}

\begin{exam}
In Figure~\ref{fig: scheven}, we illustrate Theorem~\ref{thm: ueven} in the case $k=6$ and the supercritical bifurcation is forward.
Here exactly half the of the $20$ branches generated by the supercritical pitchfork bifurcations along $S_6 L_3$ connect to $10$ index $3$ branches to give 
$10$ crossing curves, the remaining $10$ branches join $10$ index $2$ branches through saddle-node bifurcations. In more detail, assume $\eta > 0$ is sufficiently small. Starting with $\lambda < 0$, 
the index $5$ trivial solution branch will connect, via a saddle-node bifurcation, with the index $4$ branch which is the perturbation of the branch along $L_1$. As $\lambda< 0$ is increased, the
remaining $5$ index $4$ branches will connect, via saddle-node bifurcations, with  $5$ index $3$ branches leaving $10$ index $3$ branches. It is at this point that 
the process differs from what happens if $k$ is odd. The $10$ index $3$ branches will join with $10$ of the $20$ branches generated by the pitchfork bifurcation along the 
$10$ axes in $S_6 L_3$. The remaining $10$ index $3$ branches  connect with $10$ index $2$ branches via the usual saddle-node bifurcation and the process continues until the index 
$0$ trivial solution branch is generated in a saddle-node bifurcation with an index $1$ branch. A count verifies there are $10$ crossing curves (index $3$) and $22$ saddle-node
bifurcations.
\begin{figure}[th]
\centering
\includegraphics[width=\textwidth]{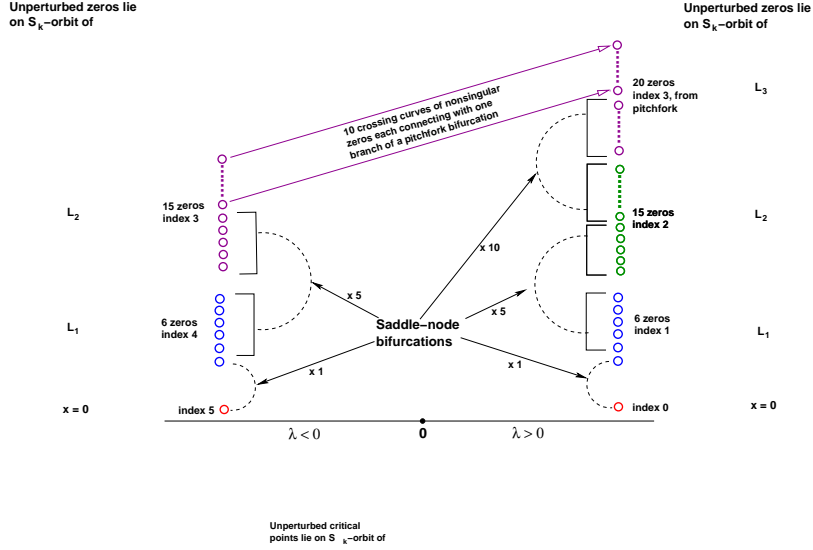}
\caption{Schematic illustrating minimal symmetry breaking model of the $S_6$-equivariant family to an $S_5$-equivariant family with only saddle-node bifurcations. There are 
$22$ saddle-node bifurcations and $10$
crossing curves of index $3$ nonsingular zeros traversing the origin.}
\label{fig: scheven}
\end{figure}
\examend
\end{exam}

In large part the proof of Theorem~\ref{thm: ueven} follows that of Theorem~\ref{thm: uodd} and only 
two additional results are needed related to the presence of pitchfork bifurcation along axes lying in the $S_k$-orbit of $\boldsymbol{\vareps}_\ell$.
For this we need to quantify dynamics on $E_\ell = \{(x, y^{\ell-1}, z^\ell) \dd x + (\ell-1) y + \ell z = 0\}$.
Recall from Section~\ref{sec: odd} that the map $U_\ell: \real^2 \arr E_\ell$ defined by
\begin{eqnarray*}
U_\ell(u,0)&=&\frac{1}{\sqrt{k(k-1)}}\big((k-1)u,-u^{\ell-1},-u^\ell\big)\\
U_\ell(0,v) & = & \frac{2}{\sqrt{k(k-1)(k-2)}}\big(0, \ell v^{\ell-1},-(\ell-1)v^\ell\big)
\end{eqnarray*}
is an isometry and $E_\ell $ contains the axes of symmetry
\begin{eqnarray*}
L_1 & = & \real \boldsymbol{\vareps}_1  
 =   \{U_\ell(u,0) \dd u \in \real\}\\
L_{\ell-1}^\star & = & (1\ell)\real \boldsymbol{\vareps}_{\ell-1}
 =  \{U_\ell(u,-k\sqrt{\frac{1}{k-2}}\,u) \dd u \in \real\}\\
L_\ell & = & \real \boldsymbol{\vareps}_\ell 
 = \{U_\ell(u,\sqrt{(k-2)}\,u) \dd u \in \real\}
\end{eqnarray*}

\begin{lemma}
Let $f^-_\lambda(\xx) = \lambda\xx -Q(\xx) - \|\xx\|^2\xx$.
The zeros of $f^-_\lambda|E_\ell$ are $c_1(\lambda) \in L_1$, $c_{\ell-1}^\star(\lambda) \in L^\star_{\ell-1}$ and, for $\lambda > 0$,  $c^{\pm}_\ell(\lambda) \in L_\ell$, where
\begin{enumerate}
\item $c_1(\lambda) = \lambda \frac{\sqrt{k(k-1)}}{k-2}U_\ell(1,0) + O(\lambda^2)$.
\item $c_{\ell-1}^\star(\lambda) = \frac{\lambda}{4} \sqrt{\frac{k}{k-1}} U_\ell\big(-(k-2),k\sqrt{k-2}\big) + O(\lambda^2)$
\item $c_{\ell}^\pm(\lambda) = \pm\sqrt{\frac{\lambda}{(k-1)}}U_\ell(1,\sqrt{k-2}) = \pm \lambda^\frac{1}{2} \left(\big(\frac{1}{\sqrt{k}}\big)^\ell,\big(\frac{-1}{\sqrt{k}}\big)^\ell\right)$.
\end{enumerate}
A similar result holds for $f^+_\lambda$.
\end{lemma}
\proof Statements (1,2) follow from Proposition~\ref{prop: sn0} (the $O(\lambda^2)$ terms come from $- \|\xx\|^2\xx$); (3) is a standard computation. \qed

\begin{figure}[h]
\centering
\includegraphics[width=\textwidth]{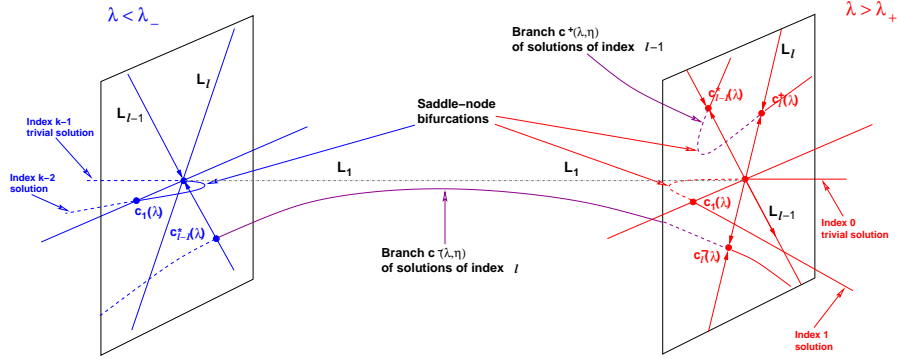}
\caption{Zeros, dynamics and bifurcation on $E_\ell$, $k = 2\ell$. Dynamics for $\lambda < 0$ is shown in blue, that for
$\lambda > 0$ in red, and new branches/connections in purple. Note that $\lambda_-$ (resp.~$\lambda_+$) denotes the
$\eta$-dependent value of $\lambda$ at which the branch of sinks (resp.~sources) meets the index $k-2$ (resp.~$1$) branch in a
saddle-node bifurcation.}
\label{fig: F22}
\end{figure}

\begin{prop}\label{prop: snlu2}
Let $\eta > 0$. The perturbed equations on $E_\ell$ are
{\small
\begin{eqnarray}\label{eq: Mul2}
\hspace*{0.4in} \ud & = & \lambda u  - \frac{1}{\sqrt{k(k-1)}}((k-2)u^2-v^2) -(u^2+v^2) u - \eta\\
\label{eq: Mvl2}
\hspace*{0.4in} \vd & = & \lambda v + \frac{2}{\sqrt{k(k-1)}} uv  - \frac{2}{\sqrt{k(k-1)(k-2)}} v^2 - (u^2+v^2)v,
\end{eqnarray}} \normalsize
\hspace*{-0.05in}In terms of the parameter $\eta$, we have a curve $c^-(\lambda,\eta)$ of non-singular
zeros of index $\ell$ such that for $\lambda \ll 0$, $c^-(\lambda,\eta)$ is close to $c^\star_{\ell-1}(\lambda)$ and
for $\lambda \gg 0$, $c^-(\lambda,\eta)$ is close to $c_\ell^-(\lambda)$. There
is also a branch $c^+(\lambda,\eta)\subset E_\ell \times \real^+$ with a single saddle-node bifurcation near $\lambda = 0$.
The index along the branch changes from $\ell$ to $\ell-1$, with the index $\ell$ component
approximating $c_\ell^+(\lambda)$ and the index $\ell-1$-component approximating
$c^\star_{\ell-1}(\lambda)$ (see Figure~\ref{fig: F22}).
The $S_{k-1}$-equivariance implies analogous results for all the
curves lying in the $S_{k-1}$-orbit of $E_\ell$.
\end{prop}
\proof Similar to that of Proposition~\ref{prop: snlu} and we omit the details. \qed

\pfof{Theorem~\ref{thm: ueven}.} The proof is broadly similar to that of Theorem~\ref{thm: uodd}.
First, the saddle-node bifurcations of
the perturbed branches of index not equal to $\ell, \ell \pm 1$ are handled along the same lines as in the proof of Theorem~\ref{thm: uodd}.
The crossing branches, and saddle-node bifurcation between index $\ell$ and index $\ell-1$ branches, use Proposition~\ref{prop: snlu2}.
Finally, for localization, constants are chosen, as in the proof of Theorem~\ref{thm: uodd}, so that all the 
perturbation and bifurcation occurs with the assigned neighbourhood $W$. Typically,
this will require $\eta_1>0$ to be small. In order to show that no new zeros are introduced, we use the known result that
$Q$ is of \emph{relatively hyperbolic type}\cite[\S\S 4,10]{FR1992a}, \cite[\S 16.2.5]{FR1992}. This implies that, for sufficiently small $\eta>0$, all the zeros within $W$ are pinned
to fixed point spaces (even if no longer real) and by Bezout's theorem no new zeros are introduced (same argument as in the proof of Theorem~\ref{thm: uodd}).
\qed

\section{Concluding comments}\label{sec: comments} 
The focus in this article has been on the creation of local minima and forced symmetry breaking for the standard 
representation of $S_k$. Motivation for this work came from an analysis of symmetry properties of a student-teacher shallow neural net used for
theoretical investigations in machine learning (we refer to~\cite{ArjevaniField2021a} for background and references). 
In the simplest case, when the number of neurons $k$ equals the number of inputs $d$,  the weight space for the student-teacher network is the space $M(k,k)$ of $k \times k$-matrices.
If we fix a  target weight $\VV \in  M(k,k)$, then the \emph{loss}  is defined by 
\[
\mathcal{L}(\WW) = \frac{1}{2}\mathbb{E}_{\xx\sim \mathcal{N}(0,I_k)}\left(\sum_{i\in\is{k}}\sigma(\ww^i\xx) -\sum_{i\in\is{k}}\sigma(\vv^i\xx)\right)^2, \;\WW\in M(k,k),
\]
where $\sigma$ is the ReLU activation function defined by $\sigma(t) = \max\{0,t\}$, $t \in \real$, and $\ww^i$ denotes 
the $i$th row of $\WW\in M(k,k)$~\cite[\S 4]{ArjevaniField2021a}. Note that the distribution $\mathcal{N}(0,I_k)$ is orthogonally invariant and equivalent to Lebesgue---the
key properties used. There is a natural action of $\Gamma \defoo S^r_k \times S^c_k$ on $M(k,k)$ defined by permuting rows ($S^r_k$-factor) and columns ($S^c_k$-factor). 
It is trivial that $\mathcal{L}$ is always $ S^r_k$-invariant. If we take  $\VV=I_k\in M(k,k)$, then  $\mathcal{L}$ is $\Gamma$-invariant~\cite{ArjevaniField2019c,ArjevaniField2021a}
and the global minimum zero of $\mathcal{L}(\WW)$ is attained iff $W \in \Gamma \VV$~\cite[Prop.~4.14]{ArjevaniField2021a}. The isotropy group $\Gamma_\VV$ is equal to
$\Delta S_k$ (diagonal subgroup of $\Gamma$). Henceforth, we write  
$S_p$ rather than $\Delta S_p$, $p \in \is{k}$---only actions of subgroups of $\Delta S_k \approx S_k$ are considered.

In the introduction, we remarked that under gradient descent (or SGD), there is a transition from saddle to local minimum at
$k \approx 5.58$~\cite{ArjevaniField2021a,ArjevaniField2021d}. These minima are strictly positive and are usually referred to as \emph{spurious minima}. They 
were first seen numerically in this problem for $k \in [6,20]$~~\cite{SafranShamir2018}.  The associated critical points
of the loss function have isotropy conjugate to $S_{k-1} = S_{k-1} \times \{e\}$ and, following~\cite{ArjevaniField2021a},
we refer to these critical points (or the minima) as being of \emph{type II}.  We indicate next how the results of the article help to understand this transition. 

A full analysis of the stability of type II critical points requires the 
isotypic decomposition of $S_{k-1}$ acting on $M(k,k)$ (the decomposition is independent of $k \ge 5$ and $\mathfrak{s}_{k-1}$ has
multiplicity $5$ in the $S_{k-1}$-representation $M(k,k)$~\cite[Thm.~4]{ArjevaniField2020b}).
The fixed point space $F_{k-1,1} \subset M(k,k)$ of $S_{k-1}\times \{e\}\subset S_k$ is of dimension $5$ (independent of $k \ge 3$) and contains 
one critical point $\mathfrak{c}_k$ of type II. In the
standard way, we reduce the analysis to $F_{k-1,1}$. For gradient descent dynamics on $F_{k-1,1}$,  $\mathfrak{c}_k$ is a sink for dynamics 
restricted to $F_{k-1,1}$, $k \ge 3$---and is easy to
find numerically using gradient descent on $F_{k-1,1}$. For dynamics on $M(k,k)$,
$\mathfrak{c}_k$ is a saddle point for $k \le 5$, and a local minimum  if $k \ge 6$ (detectable by gradient descent or SGD not initialized on $F_{k-1,1}$~\cite{SafranShamir2018}). 

The change in stability of $\mathfrak{c}_k$ on $M(k,k)$ can be shown by spectral analysis of the Hessian 
which verifies a change in sign of eigenvalues (using the $k \ge 5$-independent $S_{k-1}$ isotypic decomposition) associated to a bifurcation tangent to a copy of 
$\mathfrak{s}_{k-1}$ at $k \approx 5.58 $~\cite{ArjevaniField2020b,ArjevaniField2021d}. 
The representations $\mathfrak{s}_4$  and $\mathfrak{s}_5$ both have two conjugacy classes of axes of symmetry. For generic bifurcation on $\mathfrak{s}_{4}$, we expect (generically)
$4 + \delta 6$ branches of hyperbolic saddles of low index to collide with the source at the 
bifurcation point where $\delta = 1$ (resp.~$0$) if there are (resp.~are not) pitchfork branches $k \le 5$. 
Similarly, at $k = 6$, we expect $15$ branches of hyperbolic saddles of high index to collide with the sink at the bifurcation point, $k < 6$. The hyperbolic saddles should all lie on
axes of symmetry.  In particular, if $k=5$, we expect the saddle points to have isotropy conjugate to either $S_3 = S_3 \times \{e\} \subset S_4$ or $S_2 \times S_2 \subset S_4$.
If $k =6$, the  saddle points have isotropy conjugate to either $S_4  = S_4 \times \{e\} \subset S_5$ or $S_3 \times S_2 \subset S_5$. The associated fixed point spaces $F_{k-2,1,1}$,
$F_{k-3,2,1}$ have dimensions $10$ and $11$ respectively and $F_{k-2,1,1}\cap F_{k-3,2,1} = F_{k-1,1}$, for all $k \ge 5$. 
The natural conjecture is that, within the fixed point spaces, 
bifurcations along all axes of symmetry occur at the same $k$ value $\approx 5.58$ found by the Hessian analysis and that this a general phenomenon 
for the creation of spurious minima.  
For type II critical points, numerical checking of the conjecture shows that
bifurcation along axes of symmetry does occur at $k\approx 5.58$. 

Although the bifurcation does not occur at an integer value of $k$, analysis of the Hessian and dynamics on fixed point spaces, strongly indicate that at $k = 5$ (resp.~$k = 6$),
$\mathcal{L}$ is close to a generic bifurcation at $\mathfrak{c}_5$ (resp.~ $\mathfrak{c}_6$) along a centre manifold tangent to $\mathfrak{s}_5$ (resp.~$\mathfrak{s}_6$). If this is
so, there are strong implications about the existence of saddle points near $\mathfrak{c}_5, \mathfrak{c}_6$. Moreover the minimal unfolding results given in 
Section~\ref{sec: MMFSB} have implications for forced symmetry breaking---here, of the target $\VV$ implicit in the definition of the loss function $\mathcal{L}$.  For example,
the possibility of a small range of $k$ values for the perturbed system for which there are no sinks or sources. In general, 
it is difficult to `find' critical points numerically which are saddles in a fixed point space (gradient descent is not helpful) and this is a significant issue when $k$ is large. 
We address these issues, and mechanisms for the annihilation of spurious minima, elsewhere~\cite{ArjevaniField2020c}.

The creation of spurious minima is not artifact of the existence of non-smooth points of the loss function. For example, spurious minima can occur when 
polynomial activation is used
in optimization problems of symmetric tensor decomposition~\cite{Arjevanietal2021}.

For large values of $k$, the minimal model we construct implies a high-dimensional local landscape deformation for the creation of spurious minima without introducing
additional spurious minima with, for example, lower symmetry. Other constructions of relatively simple minimal models are surely possible---for example,
changing the sign of $\eta$ in the perturbation $\eta\boldsymbol{\vareps}_1$ used in our construction. 
This already results in interesting geometry and pitchfork bifurcations in case $k=3$.  Although the generic bifurcation on the standard representation is 
often viewed as being `transcritical', the reality is that even if $k$ is odd, pitchfork bifurcations play a significant role in understanding
dynamics. This is not so surprising, as bifurcations along axes when $k$ is odd are never a simple exchange of stability---whatever the dynamics
appear to be by restricting attention to the fixed point space.  Rather they are a stability inversion---a consequence of the presence of quadratic equivariants and
analysis of the phase vector field (cf.~\Refb{eq: reverse}). We believe the interest of quadratic equivariants lies in this point (rather than just the existence of
unstable solution branches).

For reasons of exposition, we have restricted attention to the representation $\mathfrak{s}_k$. However, our methods likely extend without difficulty to the external products
$\mathfrak{s}_k \boxtimes \mathfrak{s}_n$, $k,n \ge 3$. Indeed, the space of quadratic gradient equivariants for these representations is $1$-dimensional and, 
although we have not checked all the details,
every homogeneous quadratic equivariant is likely gradient, as is the case for $\mathfrak{s}_k$. 
These irreducible representations may well occur in the mechanisms leading to the annihilation of spurious minima. 

Certain spurious minima are associated with bifurcation tangent to the exterior square representation of $S_k$. These minima do not decay to 
zero as $k \arr \infty$~\cite[\S 8]{ArjevaniField2021a} and are not seen
in~\cite{SafranShamir2018} where Xavier initialization is used.
Since this phenomenon suggests possible connections between decay rates and the formation of spurious minima, further exploration and analysis is merited.

Finally, the phenomena we have described for type II minima with isotropy $S_{k-1}$ (that is, $\Delta S_{k-1}\subset S_k^r \times S_k^c$) also occur for
families of spurious minima with isotropy $S_{k-p} \times S_p$, where $p \ll k$ (these minima are referred to as type M in~\cite{ArjevaniField2021a}). 
Modulo terms of order $O(k^{-\frac{1}{2}})$, these critical points exhibit the same Hessian spectrum~\cite[Thm.~1]{ArjevaniField2021d} suggesting 
the possibility of underlying self-similar structure in the landscape geometry of $\mathcal{L}$.
\section{Acknowledgments}
Special thanks to Ian Melbourne, for pointing out his result with Pascal Chossat and Reiner Lauterbach on the 
instability of branching~\cite[Thm.~4(b)]{CLM1990}; to Adam Parusi\'nski and Laurentiu Paunescu 
for their help with the \emph{regular arc-wise analytic stratification};  and to David Trotman for his help with Paw\l{}ucki's theorem. Last, but not least, MF would like to
express his gratitude to Tzee-Char Kuo for his wisdom and many
rewarding mathematical discussions over the years on singularity and stratification theory, for recently telling us about the arc-wise analytic 
stratification of Parusi\'nski \& Paunescu, and for his variation on Steenrod's comment about the importance of having the correct 
definition: \emph{The definition is wrong, hence the whole theory is wrong}~\cite[\S 8 \& \S 2]{Kuo}.


\begin{thebibliography}{99}
\bibitem{Arjevanietal2021} Y Arjevani, J Bruna, M Field, J Kileel, M Trager, \& F. Williams. `Symmetry breaking in symmetric tensor decomposition',
preprint 2021, arXiv:2103.06234.
\bibitem{ArjevaniField2019c}  Y Arjevani and M Field. 'Spurious Local Minima of Shallow ReLU Networks Conform with the Symmetry
  of the Target Model', arXiv:1912.11939.
\bibitem{ArjevaniField2020b} Y Arjevani and M Field. `Analytic Characterization of the Hessian\\
        in Shallow ReLU Models: A Tale of Symmetry', {\em Proc. NeurIPS 2020,  Vancouver, Canada, 2021}, arXiv:2008.01805.
\bibitem{ArjevaniField2021a} Y Arjevani and M Field. `Symmetry \& critical points for a model shallow neural network', to appear in \emph{Physica D}, special issue on Machine Learning and Dynamical Systems, arXiv:2003.10576.
\bibitem{ArjevaniField2021d} Y Arjevani and M Field. `An analytic study of families of spurious minima in two-layer ReLU neural networks', preprint 2021.
\bibitem{ArjevaniField2020c} Y Arjevani and M Field. `Bifurcation, spurious minima and over-specification', in preparation. 
\bibitem{AGK} D G Aronson, M G Golubitsky \& M Krupa. `Coupled arrays of
Josephson junctions and bifurcation of maps $S_n$ symmetry', \emph{Nonlinearity} {\bf 4} (1991), 861.

\bibitem{bier2} E Bierstone. `Generic equivariant maps', \emph{Real and Complex Singularities} (Proc. Ninth Nordic Summer School/NAVF Sympos. Math.,
Oslo, 1976, Sijthoff and Noordhoff, Alphen aan den Rijn, 1977), 127--161.
\bibitem{CLM1990} P Chossat, R Lauterbach \& I Melbourne. `Steady-State Bifurcation with $\On{3}$ Symmetry',
\emph{Arch. Rational Mech.~Anal} {\bf 113} (1990), 313--376.
\bibitem{CL2000} P Chossat and R Lauterbach, \emph{Mathods in Equivariant Bifurcations and Dynamical Systems} (World Scientific, Advanced series in nonlinear dynamicsi {\bf 15}, 2000).
\bibitem{Field1989} M J Field. `Equivariant Bifurcation Theory and Symmetry Breaking',
\emph{J. Dynamics and Diff. Eqns.} {\bf 1}(4) (1989), 369--421.
\bibitem{FieldM1996} M J Field. \emph{Symmetry breaking for compact Lie groups} (Mem. Amer. Math. Soc., {\bf 574}, 1996).
\bibitem{Field2007} M J Field. \emph{Dynamics and Symmetry} (Imperial College Press Advanced Texts in Mathematics --- Vol. 3, 2007.)
\bibitem{FR1989} M J Field and R W Richardson. `Symmetry breaking and the Maximal Isotropy Subgroup Conjecture for Reflection Groups',
\emph{Arch. Rational Mech.~Anal} {\bf 105}(1) (1989), 61--94.
\bibitem{FieldRichardson1990}  M J Field and R W Richardson. `Symmetry breaking in equivariant bifurcation
problems', \emph{Bull. Am. Math. Soc.} {\bf 22} (1990), 79--84.
\bibitem{FR1992a} M J Field and R W Richardson. `Symmetry breaking and Branching Patterns in Equivariant Bifurcation Theory I', \emph{Arch. Rational Mech.~Anal} {\bf 118} (1992), 297--348.
\bibitem{FR1992} M J Field and R W Richardson. `Symmetry breaking and Branching Patterns in Equivariant Bifurcation Theory II', \emph{Arch. Rational Mech.~Anal} {\bf 120} (1992), 147--190.
\bibitem{FultonHarris1991} W Fulton and J Harris. \emph{Representation Theory} (Graduate Texts in Mathematics {\bf} 129, Springer-Verlag, 1991).
\bibitem{GSS1984}     M Golubitsky and D G  Schaeffer.
\emph{Singularities and Groups in Bifurcation  Theory,  Vol.  I} (Appl. Math. Sci. Ser. 51, Springer-Verlag, New York, 1988.)
\bibitem{GSS1988}     M Golubitsky, I N  Stewart, \& D G  Schaeffer.
\emph{Singularities and Groups in Bifurcation  Theory,  Vol.  II} (Appl. Math. Sci. Ser. 69, Springer-Verlag, New York, 1988.)
\bibitem{GS} M Golubitsky and I Stewart. \emph{The symmetry perspective: from equilibrium to chaos in phase space and physical space} (Birkh\"auser Verlag {\bf 200}, Basel, 2002).
\bibitem{IG1984} E Ihrig and M Golubitsky. `Pattern selection with $O(3)$ symmetry', \emph{Physica D} {\bf 12} (1984), 1--33.
\bibitem{James1978} G D James. \emph{The Representation Theory of the Symmetric Groups} (Spring Lecture Notes {\bf 682}, Springer-Verlag, 1978).
\bibitem{Ka1976} T Kato. \emph{Perturbation theory for linear operators} (Grundlehren {\bf 132}, 1976, Springer-Verlag, Berlin, New York).
\bibitem{Kn1996} K Knopp. \emph{Theory of functions, Parts I and II} (Dover, 1996).
\bibitem{Kuo} T C Kuo.  `An old man's mathematical Stories' \emph{Proc. JARCS 2017, Australian-Japanese Real and Complex Singularities Workshop}.
\bibitem{L2015} R. Lauterbach, `Equivariant Bifurcation and Absolute Irreducibility in $\real^8$: A Contribution to Ize Conjecture and Related Bifurcations', \emph{J.  Dynam. and Diff. Eqns}
{\bf 27} (2015),  841-–861.
\bibitem{Slo95}  S Lojasiewicz. `On semi-analytic and subanalytic geometry', \emph{Banach Center Publications} {\bf 34} (1995), 89--104.
\bibitem{Mal1966} B Malgrange. \emph{Ideals of differentiable functions} (OUP, London, 1966).
\bibitem{Mather1973} J N Mather. `Stratifications and mappings', \emph{Proceedings
of the dynamical systems conference, Salvador, Brazil} (ed M Peixoto) (Academic Press, New York, 1973), 195--223.
\bibitem{Mather1977} J N Mather. `Differentiable invariants', \emph{Topology} {\bf 16} (1977), 145--155.
\bibitem{Mel1994} I Melbourne. `Maximal isotropy subgroups for absolutely irreducible representations of compact Lie groups', \emph{Nonlinearity} {\bf 7}(5) (1994), 1385--1394.
\bibitem{Milnor1965} J Milnor. \emph{Topology from the differentiable viewpoint} (Princeton University Press, 1997).
\bibitem{Milnor1968} J Milnor. \emph{Singular points of complex hypersurfaces} (Annals of Math.~Studies~61, Princeton University Press, 1968).
\bibitem{PaPa2017} A Parusi\'nski and L P\u{a}unescu. 
`Arc-wise analytic stratification, Whitney fibring conjecture and Zariski equisingularity', \emph{Adv.~in Math.}  {\bf 309} (2017), 254--305.
\bibitem{Paw1985} W Paw\l{}lucki. `Quasi-regular Boundary and Stokes formula for a sub-analytic leaf', (in
\emph{Seminar on Deformations, \L{}odz-Warsaw 1981--83}, Springer Lect.~Notes in Math. {\bf 1165} (1985), 235--252).
\bibitem{Re1968} F Rellich. \emph{Perturbation Theory of Eigenvalue Problems} (1953 New York University Lect. Notes reprinted by Gordon and Breach, 1968).
\bibitem{SafranShamir2018} I Safran and O Shamir. `Spurious Local Minima are Common in Two-Layer ReLU Neural Networks', 
\emph{Proc. of the 35th Int. Conf. on Machine Learning} {\bf 80} (2018), 4433--4441 (for data sets, see
https://github.com/ItaySafran/OneLayerGDconvergence).
\bibitem{trott2020} D Trotman. \emph{Stratification theory} (Cisneros-Milina Jos\'e Luis, D\~{u}ng Tr\'ang L\^{e}, Seade Jos\'e (Eds.).
Handbook of Geometry and Topology of Singularities I, 2020.
hal-03186972).
\bibitem{V} A Vanderbauwhede, `”Local bifurcation and symmetry', \emph{Research Notes in Mathematics} {\bf 75}, Pitman (Boston), 1982.
\bibitem{Walker} R J Walker. \emph{Algebraic Curves}, Springer, 1978.
\end{thebibliography}
\end{document}